%% file: root.tex
\documentclass[runningheads]{llncs}
\usepackage[T1]{fontenc}
\usepackage{graphicx}
\usepackage{url}
\usepackage{cite}
\usepackage{amsmath} 

\usepackage{multirow}
\usepackage{comment}
\usepackage{rotating}
\usepackage{subfig}

\usepackage{docmute}
\usepackage{pdflscape}
\newsavebox\thesmashminipage
\newenvironment{smashminipage}
  {\begin{lrbox}{\thesmashminipage}\begin{minipage}[t]{1.5\textwidth}}
  {\end{minipage}\end{lrbox}\smash{\usebox{\thesmashminipage}}\clearpage}

\begin{document}
\title{Dense Road Surface Grip Map Prediction from Multimodal Image Data}
%
%
\author{Jyri Maanp{\"a}{\"a}\thanks{The authors Jyri Maanp{\"a}{\"a} and Julius Pesonen shared an equal contribution to this work as first authors.}\inst{,1,2}\orcidID{0000-0001-6772-9611} \and
Julius Pesonen\inst{\star,1}\orcidID{0009-0000-9175-7129} \and
Heikki Hyyti\inst{1}\orcidID{0000-0003-4664-6221} \and
Iaroslav Melekhov\inst{2}\orcidID{0000-0003-3819-5280} \and
Juho Kannala\inst{2}\orcidID{0000-0001-5088-4041} \and
Petri Manninen\inst{1}\orcidID{0000-0003-1289-2811} \and
Antero Kukko\inst{1}\orcidID{0000-0002-3841-6533}\and
Juha Hyypp{\"a}\inst{1}\orcidID{0000-0001-5360-4017}}
\authorrunning{J. Maanp{\"a}{\"a} et al.}
%
\institute{Finnish Geospatial Research Institute FGI, National Land Survey of Finland, 02150 Espoo, Finland \\ \email{jyri.maanpaa@nls.fi, julius.pesonen@nls.fi} \and
Department of Computer Science, Aalto University, 02150 Espoo, Finland}
\maketitle              
\begin{abstract}
Slippery road weather conditions are prevalent in many regions and cause a regular risk for traffic. Still, there has been less research on how autonomous vehicles could detect slippery driving conditions on the road to drive safely. In this work, we propose a method to predict a dense grip map from the area in front of the car, based on postprocessed multimodal sensor data. We trained a convolutional neural network to predict pixelwise grip values from fused RGB camera, thermal camera, and LiDAR reflectance images, based on weakly supervised ground truth from an optical road weather sensor.

The experiments show that it is possible to predict dense grip values with good accuracy from the used data modalities as the produced grip map follows both ground truth measurements and local weather conditions, such as snowy areas on the road. The model using only the RGB camera or LiDAR reflectance modality provided good baseline results for grip prediction accuracy while using models fusing the RGB camera, thermal camera, and LiDAR modalities improved the grip predictions significantly.

\keywords{Grip prediction  \and Autonomous driving \and Convolutional neural networks.}
\end{abstract}
\section{Introduction}

Harsh winter conditions pose unique challenges to autonomous driving. According to the Road Weather Management Program by the U.S. Department of Transportation, 24~\% of weather-related vehicle crashes in the U.S. occur on snowy, slushy, or icy pavement and 15~\% happen during snowfall or sleet each year~\cite{rwmbyus}. Besides low visibility, significant challenges posed by winter conditions are changes in road surface slipperiness. Snowy and icy road surfaces in particular can cause the friction between the wheels of the vehicle and the road to be much smaller than on dry or wet roads. Thus, autonomous driving systems have to be capable of distinguishing such scenarios for which specialized sensing solutions are required. 

Several approaches exist for estimating the grip on the road. However, the greatest shortcoming of most of these methods in the sense of autonomous driving has been the lack of ability to sense the road ahead of the vehicle, thus only allowing the vehicle to react in situations where the slippery conditions have already affected the driving. To enable sensing of the road further ahead, cameras or other forward-facing, longer-range sensors must be deployed. 

In addition, the grip can often vary between different sections of the road depending on local snow, ice, and water layer thicknesses. For example, snowy roads with frequent traffic usually have clear tire tracks, which human drivers follow to avoid the snowy areas on the road. Human drivers can also distinguish sudden icy or wet areas on the road allowing them to either avoid these areas or decrease driving speeds accordingly. 
Therefore producing dense grip predictions would enable autonomous vehicles to react to the sensed conditions in a manner that human drivers can achieve.

\begin{figure*}[t]
    \centering
    \includegraphics[width=0.9\textwidth]{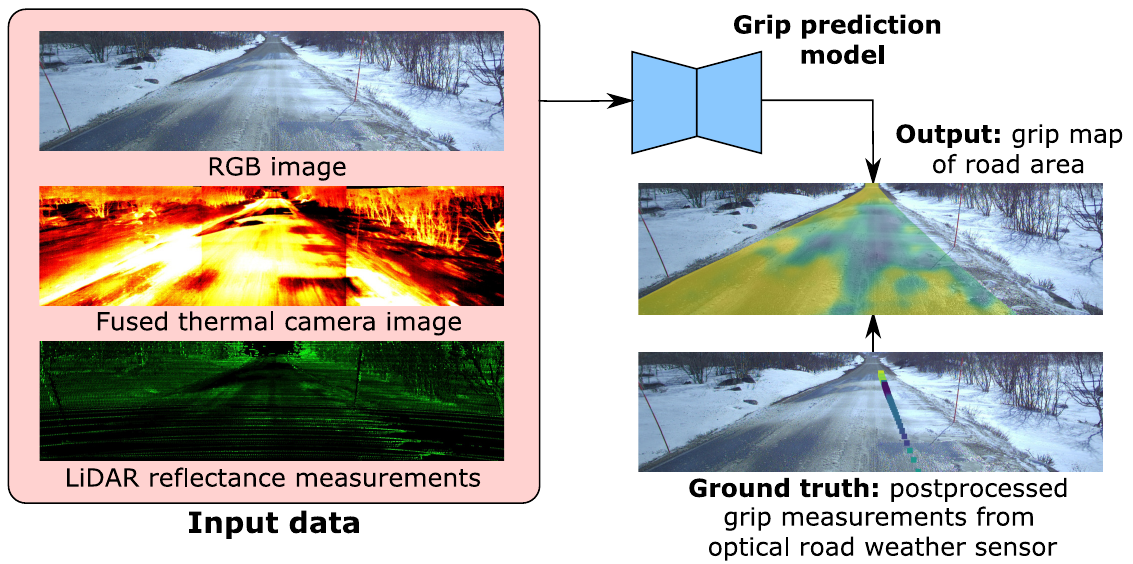}
    \caption{Our work presents a grip prediction model, which operates on pixelwise fused RGB camera, thermal camera, and LiDAR reflectance measurements and predicts a dense grip map of the road area. The ground truth for training is obtained with an optical road weather sensor that provides road grip measurements which are postprocessed with GNSS trajectories and external calibrations to match the input data.}
    \label{fig:scheme_idea}
\end{figure*}

Besides traditional vision by RGB cameras, even more accurate grip predictions could be achieved by combining measurements from other long-range sensors. For example, many LiDARs measure the return intensity of the sent infrared laser pulse, which could be used to differentiate ice and water on the road due to their different optical properties. In addition, a thermal camera might be able to differentiate some road surface layer types, such as snowy and clear areas on the road. These sensors are commonly used in autonomous vehicles, so their positive effect on grip prediction could be easily adopted by the industry.

In this paper, we introduce a pixelwise road surface grip prediction model based on a convolutional neural network (CNN) to generate dense grip map predictions based on fused images from front-facing RGB and thermal cameras as well as LiDAR reflectance measurements. The ground truth grip values were provided by an optical road weather sensor, the data of which were projected on the other sensor data during postprocessing using 3D transformations between different sensors and the postprocessed trajectory of the data collection. The main idea and different sensor modalities are presented in Figure~\ref{fig:scheme_idea}. We collected a 37~hour (1538~km) dataset with our autonomous research platform ARVO (Figure~\ref{fig:arvo}) within different adverse weather conditions and preprocessed it for the aims of this study.

This work extends the previous work by Pesonen~\cite{pesonen2023pixelwise} and provides a new ablation study of the grip prediction accuracy between different input data modalities. The previous approach was also improved with a more consistent training and validation setup and extended testing. The capability of the dense grip map prediction was measured using quantitative error measurements and qualitative analysis for road areas where ground truth measurements could not be obtained. The study shows that the dense grip predictions are improved with the fused RGB, thermal camera, and LiDAR inputs, while the model relying on the sole RGB inputs, already, greatly improves the resolution of any prior camera-based grip prediction methods. 

Our contributions to the state of the art are:
    1) We developed a novel method to collect and process a dataset with pixelwise matching of multimodal images and sparse road grip measurements.
    2) We proposed a model to predict a dense grip map of the road area in diverse weather conditions.
    3) We compared the grip prediction accuracies of models using RGB images, thermal images, and LiDAR reflectance measurements as model input modalities both separately and with every combination using multi-encoder-fusion.

We shared a demo of our models in a Gitlab repository to allow readers to test our methods.\footnote{\url{https://gitlab.com/fgi_nls/public/grip-prediction}}

\section{Background}

While dense road surface grip map prediction has only been proposed in our earlier work \cite{pesonen2023pixelwise}, methods for grip prediction have been proposed before using various sensor setups.
Road surface grip measurement methods can be roughly divided into non-contact and contact-based measurements, which have been addressed in surveys by Ma et al. \cite{ma2022current} and Acosta et al. \cite{acosta2017road} respectively. Even though contact-based grip sensors and models relying on vehicle information, such as wheel rotation speeds, are incapable of producing the required predictions for grip in front of the car, their use has been essential for evaluating the later-developed non-contact methods. 
According to the survey by Ma et al. \cite{ma2022current} the most prominent non-contact-based methods rely on infrared spectroscopy, computer vision, optical polarisation, or radar detection. %

Most of the proposed camera-based road surface grip prediction methods have relied on classification of different road surface conditions without providing scalar estimates of the surface grip \cite{panhuber2016recognition, roychowdhury2018machine} or by using a two-part process where the classification result is further used to generate a scalar estimate of the road surface grip \cite{du2019rapid, vsabanovivc2020identification, langstrand2023using, zhao2024road}. Few models have also been suggested for directly generating scalar grip estimates \cite{cech2021self, du2023pavement, ojala2023enhanced}. However, as a common limitation, all of the models rely on either generating a single prediction for the whole input image or for small regions of interest in predefined shapes. 
In some of the studies the ground truth labeling was generated by expert annotators \cite{roychowdhury2018machine, zhao2024road}, in one using a portable pendulum tester \cite{du2019rapid}, in one using friction wheel trailer measurements \cite{langstrand2023using}, in one with vehicle response \cite{cech2021self}, and in one with an optical sensor \cite{ojala2023enhanced}. 

Models generating pixelwise outputs have become popular in many tasks, such as semantic segmentation and monocular depth estimation. In semantic segmentation, models are trained to classify each pixel of the input image. 
Solutions proposed for the task, some of which have also found use in many other problems, include U-net \cite{ronneberger2015u}, FPN \cite{lin2017feature} and DeepLabV3+~\cite{deeplabv3plus2018}.

Monocular depth estimation is a task more similar to the one presented in this paper, as the labels are scalar distance values instead of discrete classes as in the case of segmentation. In addition, the training labels could originate from sparse measurements such as LiDAR readings. Such weak supervision has been applied to monocular depth estimation with sparse labels by Guizilini et al. \cite{guizilini2020robust} showing similarity to our grip prediction task due to the comparable sparsity of the ground truth labels.

As both optical road weather sensors and LiDARs use lasers to measure the return intensity of the measured object, the use of LiDARs for road surface condition prediction shows potential. Ruiz-Llata et al. \cite{8234230} and Shin et al. \cite{shin2019characteristics} showed that different road surface conditions can be detected using LiDAR measurements. Sebastian et al. \cite{sebastian2021rangeweathernet} proposed the use of a LiDAR-based CNN for simultaneous road condition and weather classification. While the use of 3D LiDARs was proposed in the studies, their use for dense road surface grip prediction was not investigated in depth.

As noted, the prior literature is concentrated on low-resolution grip predictions using individual sensor inputs. This study aims to fill the gap by introducing both data and methods for generating dense predictions using multimodal input data.

\section{Data}

In this section, we describe the collection and preprocessing of the data used to train and evaluate the proposed grip prediction methods.

\subsection{Dataset Collection}

We collected a 37~hour or 1538~km dataset of different driving conditions with the sensor setup in our autonomous driving research vehicle ARVO, of which an older version is presented in our earlier study~\cite{maanpaa2021multimodal}. The dataset includes various driving conditions, such as daytime and nighttime, snowfall, snow-covered roads, slushy conditions, rain, and wet roads. It also contains data from several road types, such as highways, urban roads, and paved and unpaved local roads. The dataset was collected mostly in the capital region of Finland during fall and winter, with a smaller part collected in Western Lapland during spring. The dataset was postprocessed to contain samples at a frequency of 2~fps and after automatic filtering of low-quality data, the dataset had 237\ 067 samples.

\begin{figure}[b!]
    \centering
    \includegraphics[width=0.7\linewidth]{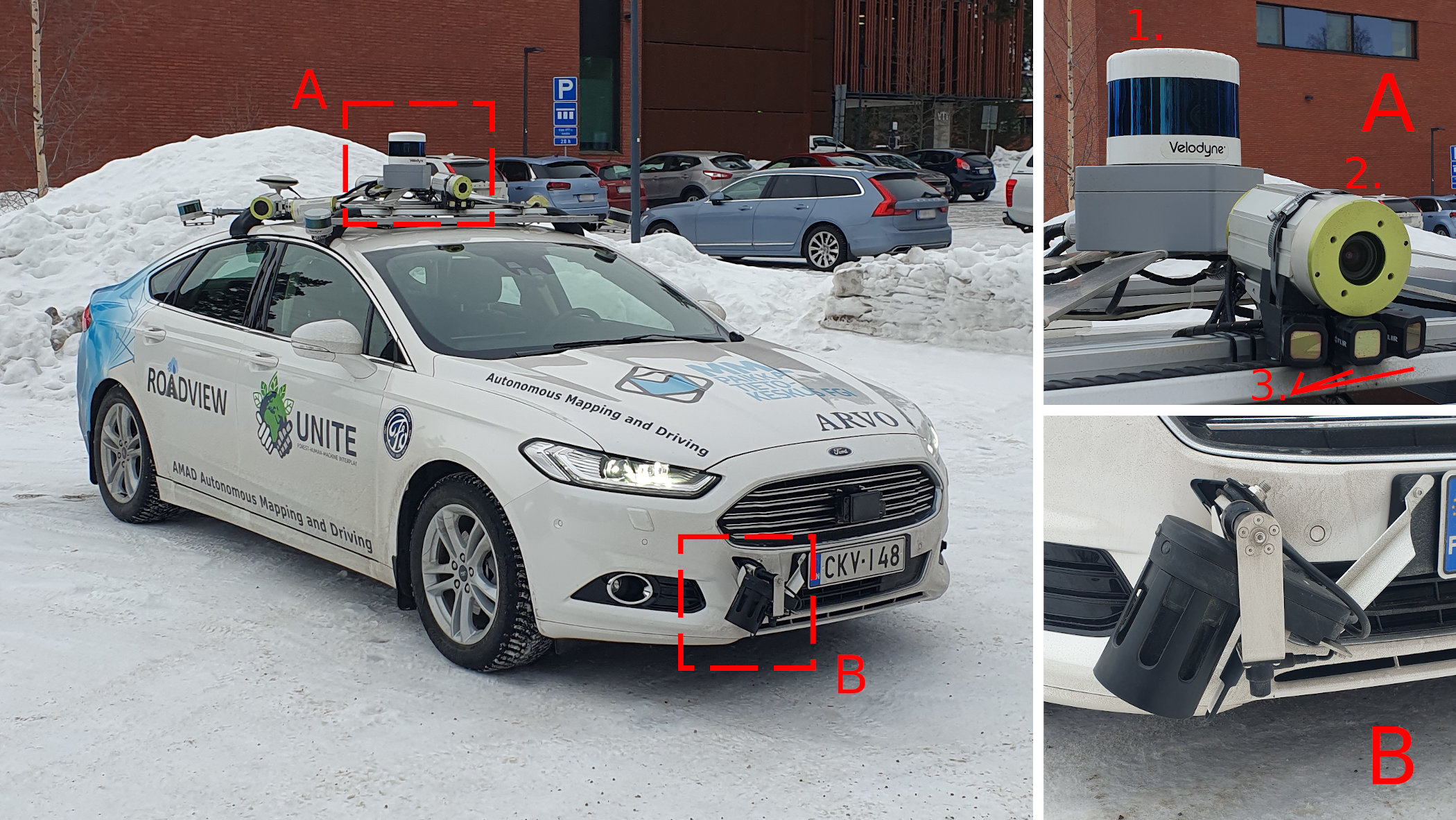}
    \caption{The research vehicle ARVO used for data collection. The long-range sensors shown in box A are 1. LiDAR, 2. RGB camera in a weatherproof housing and 3. thermal cameras. The road weather sensor is shown in box B.}
    \label{fig:arvo}
\end{figure}

For this project, we used data from a forward-facing RGB camera (Basler MED ace 2.3 MP 164 color), three forward-facing thermal cameras (FLIR ADK, 24\textdegree~FOV), a roof-mounted 128-beam rotating LiDAR (Velodyne Alpha Prime VLS-128), a GNSS Inertial Navigation System (INS) (Novatel PwrPak7-E1), and a mobile road weather sensor (Vaisala Mobile Detector MD30). An image of the car with highlighted sensor locations is shown in Figure~\ref{fig:arvo}. 
The left and right thermal cameras were horizontally tilted approximately 23 degrees outwards from the center camera to achieve a combined field of view covering a larger horizontal angle. All sensors were synchronized with GNSS INS triggering signals except the road weather sensor, which was synchronized manually during postprocessing.

We chose these long-range sensor modalities for this research due to their common use in autonomous driving development. We also noted in our preliminary studies that different types of snow can have type-specific features in thermal cameras. LiDAR reflectance measurements (laser pulse return intensity amplitude which is normalized with distance internally by LiDAR sensor) could also provide single-band spectral information on the surface material as the LiDAR sensor used in this study uses 903~nm wavelength lasers, which has different extinction coefficients between water and ice according to the data by Palmer and Williams \cite{palmer1974optical} and the data by Warren and Brandt \cite{warren2008optical}. 

The road weather sensor Vaisala Mobile Detector MD30 is an optical sensor that estimates water, ice, and snow layer thicknesses using three laser intensity measurements in different wavelengths. The operating principle of the sensor is not publicly available, but an earlier sensor prototype is presented in a master's thesis~\cite{kivi2019algorithmic}. The sensor uses an internal model for calculating the grip estimate of the road, most likely based on the three surface layer thickness values. Based on earlier studies on optical sensors~\cite{malmivuo2013comparison, malmivuo2023optisten}, we assume that an optimistic upper limit of the sensor grip estimate accuracy is~$0.1$. 
However, the grip estimate can be more accurate within clear conditions with constant grip.
The surface layer thickness and grip estimates are measured with a 40~fps sampling rate. In addition, the sensor provides road surface condition class, road and air temperatures, and other meteorological measurements.
In our analysis, we have assumed that the sensor's grip measurements are sufficiently accurate that it is reasonable to imitate the sensor measurements for a dense grip map, even though the sensor grip values are likely to contain some inaccuracies as the grip between the tires and the road surface is a complex physical phenomenon.

\begin{figure}[t]
    \centering
    \includegraphics[width=0.77\linewidth]{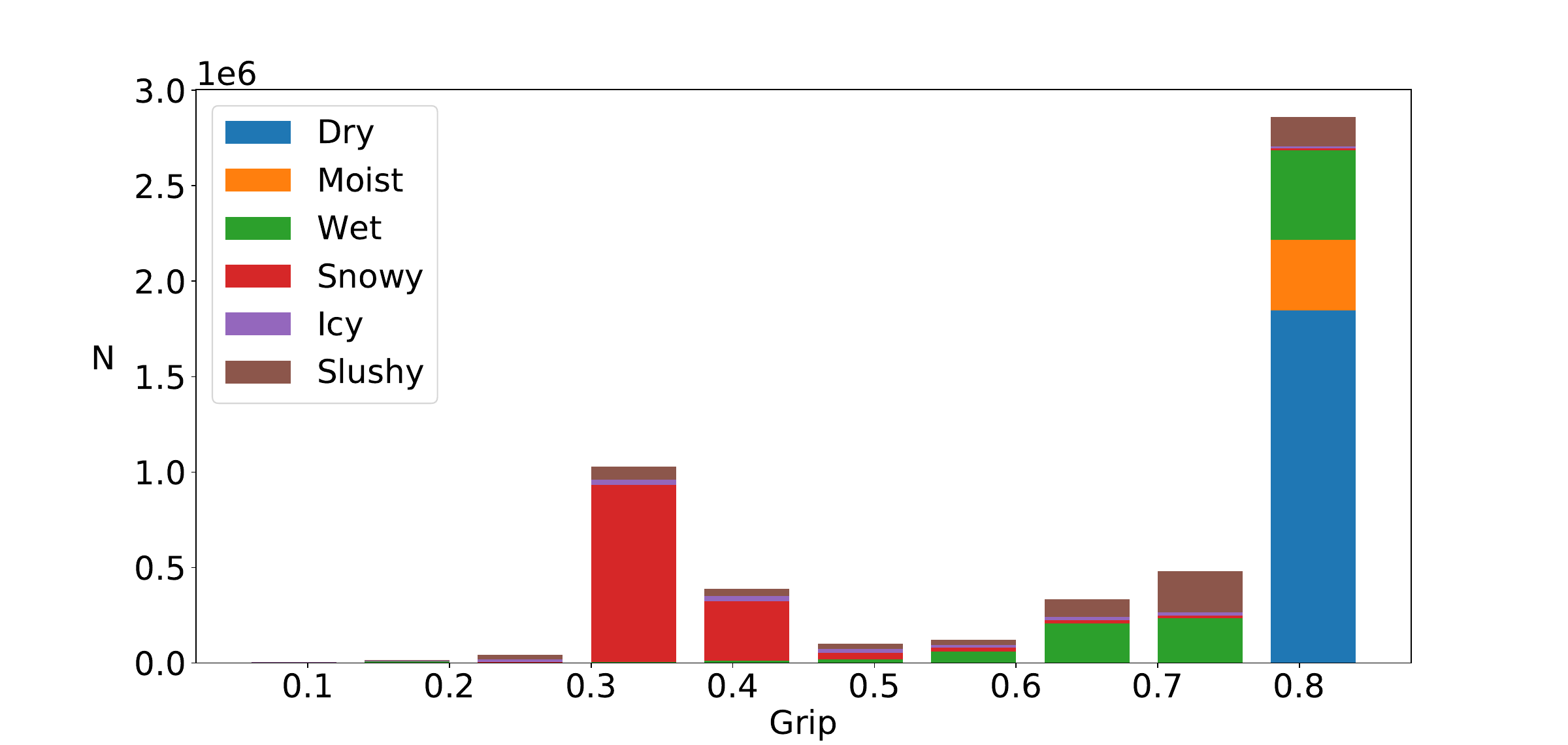}
    \caption{The distribution of grip values and road surface states provided by the road weather sensor measurements in the complete unprocessed dataset. We observe that most of the data is collected within dry, wet, or snowy conditions.}
    \label{fig:gripstate}
\end{figure}

The grip and road surface condition distribution in our dataset is visualized in Figure~\ref{fig:gripstate}. We observe that the two most prominent road states are dry and snowy conditions. These conditions have two distinct grip coefficients, which are $0.82$~for dry road and $0.35$~for snowy road. Additionally, we note that wet roads usually have a grip of less than~$0.8$, and the smallest grip of $0.1$ is observed on icy roads or roads with very thick layers of water.

\subsection{Datasplit}

The dataset includes data collections from 18 days, many of which shared the same data collection locations. To ensure that the training, validation, and test sets were collected from different locations while maintaining similar weather condition distributions, we used a geofencing-based approach to choose the validation and test sets from the full dataset. We chose circular areas within approximately one-kilometer intervals from the data collection area so that all samples collected within these areas are included either in the validation or the test set. The rest of the data is included in the training set, except for any positions less than 55 meters from any border of the chosen circular areas. This 55-meter gap assures that no observations are shared between validation, testing, or training.  With this data split, we achieved a qualitatively similar distribution of weather conditions between training, validation, and test sets. Additional qualitative data filtering was also done at this stage. In the end, the training set has 159\ 801 samples (79.1\%), the validation set has 15\ 343 samples (7.6\%) and the test set has 26\ 783 samples (13.3\%).

There is a possibility that some conditions of the input data, such as illumination, would allow the model to fit to these conditions and learn the general grip conditions on specific data collection dates. Therefore, we used three separate data collections, with 16\ 139 samples in total, as additional test drives to demonstrate the accuracy of the model regardless of this effect. 

\subsection{Pixelwise Matching of Modalities}

To obtain pixelwise pairs of image data and ground truth road weather measurements, we used the following preprocessing approach. We calibrated all cameras intrinsically and extrinsically and measured the 3D locations and orientations of each sensor. Due to the hardware-based synchronization, we also know the time correspondences between each of the sensors. The GNSS trajectory was postprocessed using base-station data 
to increase its accuracy.

We chose the RGB camera image as the reference frame of the data as it has the highest resolution regarding the front area of the car. The road weather sensor measurements were overlaid on the RGB images with the following procedure: first, we used the postprocessed trajectory and the external transformation between the INS reference frame and the road weather sensor measurement locations to project the road weather sensor measurement positions to a 3D trajectory. This trajectory of the measurements is then transformed to the RGB camera coordinates and projected to the RGB camera image plane. Therefore, we obtained RGB camera images where the road weather measurement points, which were recorded soon after the RGB camera capture time, are overlaid. To improve the data quality, we only included road weather measurement points within 50 meters of the cameras and excluded the measurement points behind any obstacles.

The LiDAR point clouds were motion-corrected with the postprocessed trajectory and projected to the RGB camera pixel coordinates. We also accumulated more LiDAR points from the lower part of the three previous scans to include more reflectance measurements from the nearby road area. 

The thermal cameras required a more complex pixelwise matching with the RGB camera. First, we generated approximate range images from the LiDAR point clouds projected on the RGB camera. Then for each RGB camera pixel, we searched a corresponding 3D point from the range image, projected that point to a single thermal camera frame, and determined the corresponding thermal value from the thermal camera image. As a result, we obtained thermal camera images projected to each RGB camera frame. We normalized the left and right thermal camera pixel values to the scale of the center of the thermal camera with the value distribution close to the shared image border.

As the thermal pixel values correspond to the thermal flux in the pixel with a varying scale due to online calibration, the raw thermal values were not considered suitable for this work. Therefore, we normalized the thermal camera pixel values within each frame so that a sample area from the road has a consistent distribution with zero mean and unit variance. In addition, the borders of the thermal camera images had lower values within cold conditions due to the operation of the thermal camera sensor. We alleviated this effect by determining the systematic error distribution for each thermal camera image and subtracting that error to obtain an image with a more homogenous value distribution. The data preprocessing is described in more detail in the preliminary results of our work~\cite{pesonen2023pixelwise}.

An example of the pixelwise matched sensor data can be seen in Figure~\ref{fig:scheme_idea}. 
The pixelwise matching quality varies between frames and occasionally distant areas or tall objects closer to the camera might appear unaligned between different sensors. We considered this effect negligible for this work, as the road surface is mostly well aligned and the road surface is usually large and homogenous, alleviating any problems that could be caused by the slight unalignment.

\section{Methods}

In this section, we present our model for the grip prediction, the training setup, and the performance evaluation methods.

\subsection{Model}

To generate dense predictions of the road surface grip using the multimodal input data, we propose using a convolutional neural network trained with the sparse pixelwise matched road weather measurements as the ground truth labels. Our models are based on Feature Pyramid Network~(FPN)~\cite{lin2017feature} which is adapted to predict pixelwise scalar values for regression. The FPN model was chosen as it was shown efficient for the task in our preliminary studies.

We trained our models with every combination of the collected input modalities to measure their effect on grip prediction accuracy. The models utilizing a single input modality are based on the standard FPN implementation which takes an image tensor as the input. However, the multimodal models include separate encoders for each input modality, and their features are concatenated channel-wise within each feature scale before being forwarded to the decoder. We implemented this feature-level fusion approach due to finding occasional lower-quality samples in some input modalities, meaning it was useful for the model to learn to discard these features in the corresponding situations. For each of the model encoders, we used ResNet-18~\cite{he2016deep}.

\begin{figure}[b!]
    \centering
    \includegraphics[width=\linewidth]{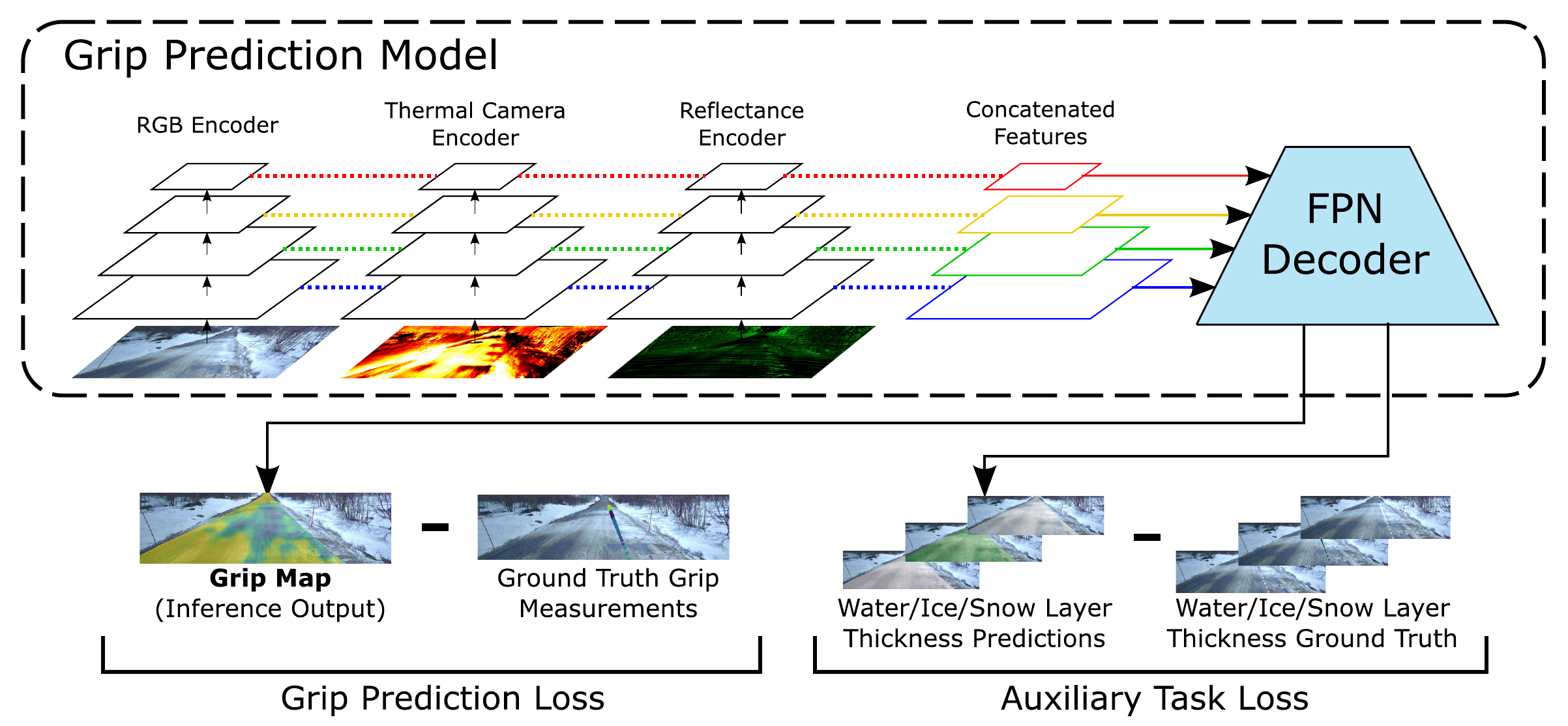}
    \caption{The model architecture and training scheme for the model using all data modalities. Each input data modality has a separate encoder and their features are concatenated within each feature scale before the FPN decoder. The loss is evaluated both for the grip and the surface layer thickness prediction tasks.}
    \label{fig:model}
\end{figure}

The outputs of the model are the predicted grip and the predicted water, ice, and snow layer thicknesses for each pixel. The grip prediction is the primary task of the model and the prediction of different surface layer thicknesses is used as an auxiliary task to support the learning, as it has been shown to improve the prediction accuracy of the obtained model in our prior experiments~\cite{pesonen2023pixelwise}. The model architecture and the training scheme are illustrated in Figure~\ref{fig:model}.

In most frames, more road weather measurement points were visible further away from the car. These distant points also contain less information as the resolution of the RGB camera and other sensors concerning the road surface was smaller. We alleviated the effect of these distant points by weighting the road weather measurements within each image based on their $y$-coordinate in the RGB image plane: the weight of each measurement point decreases linearly from the bottom of the image to the estimated horizon level. With this approach, we could approximately balance the prediction accuracy over the whole road area. For validation and testing the pixelwise weights were normalized within each frame so that their mean is one. For the training, the normalization was performed on the unfiltered road weather measurements, which included some overlapping positions, leading to slightly larger weights on average.

The predicted grip and surface layer thickness values are compared to the sparse ground truth values from the postprocessed road weather sensor data with the following loss function: %
\begin{align}\label{eq:loss}
\begin{split}
    \scalebox{0.90}{$\mathcal{L}(x,y_p,y_a,w|\theta) =  \frac{1}{N}\sum_{i=1}^N w^i (y_p^i - f_p^i(x|\theta))^2  +  \frac{\lambda}{3N}\sum_{l=1}^3 \sum_{i=1}^N w^i ( y_a^{l,i} - f_a^{l,i}(x|\theta))^2,$}
\end{split}
\end{align}
\noindent where $x$ is the input image tensor, $N$ is the number of pixels containing ground truth grip values in the sample, $w^i$ is the weight for pixel~$i$, $y_p^i$, and $f_p^i(x|\theta)$ are the ground truth and model output for grip value at pixel~$i$ and $y_a^{l,i}$ and $f_a^{l,i}(x|\theta)$ are the ground truth and model output for surface layer~$l$ value at pixel~$i$. The first term denotes the weighted mean square error for the grip prediction task and the second term denotes the similarly weighted mean square error for the prediction of surface layer thicknesses. The parameter~$\lambda$ is used to adjust the effect of the supportive auxiliary task and in our experiments, it was set to~1.0.

\subsection{Training Setup}

Each model was trained using the Adam optimizer \cite{kingma2014adam} for 38 epochs with a batch size of 32 and a learning rate of $1e-3$. The FPN model used a dropout rate of 20\% in its last layer. The models were compared using the instances which achieved the best validation loss during the training.

As our method was designed to predict the grip and the layer thicknesses based on the surface appearance, we avoided excessive augmentation to maintain accurate predictions. Some augmentations were still used to ensure appropriate generalization. Therefore, it was chosen to apply small random scale and rotation augmentation with a 30\% probability, horizontal flip with a 50\% probability, small random blur with a 30\% probability, and random color jitter to the RGB images with a 30\% probability.

\subsection{Performance Evaluation}

The model performance was evaluated with a root mean square error (RMSE) between the predicted and the ground truth grip values. A similar weighting as in the training loss~\eqref{eq:loss} was applied during validation and test error evaluation as we wanted to measure the grip prediction accuracy balanced over the road area. Due to this weighting, the mean square error was evaluated for each frame separately, and these sample-wise square errors were averaged before evaluating the square root. The test accuracy is reported both for the test set from the main data collection and the three extra test drives with no correspondence to the main dataset.

However, the error evaluation alone could not show if the grip predictions are valid over the road areas which rarely contain ground truth measurements. Therefore we also performed qualitative analysis on the model output to estimate how well the grip map follows the slipperiness expected by human drivers.

\section{Results}

In this section, we first analyze the quantitative errors from our validation and test sets and then inspect the qualitative performance of different models.

\subsection{Validation and Test Set Errors}

We performed the experiments by training the model with different sensor modalities as the input to observe the effect of each sensor on the grip prediction accuracy. The validation set, test set, and separate test drive dataset information and RMSEs achieved with each model are found in Table~\ref{tab:all_results}. 

For the validation and test set all obtained errors are significantly smaller than the standard deviation of the dataset, which insists that the models could learn to predict useful grip values. In most experiments, the best or second-to-best results are achieved with the model that uses all data modalities. Using RGB images provides the best accuracy when compared to other data modalities, but the model using only the LiDAR reflectance achieves comparable results with the RGB model. 
While the combination of RGB and thermal images does not improve performance over the sole RGB data, the combination of thermal and reflectance data provides similar improvements as the combination of RGB and reflectance, indicating that the RGB and thermal information may overlap significantly but also provide information unavailable from the LiDAR reflectance alone. Thus, almost all of the best or second-to-best results in Table~\ref{tab:all_results} are achieved using some combination of LiDAR reflectance and a higher-resolution image input.

\begin{table}[b!]
\centering
\caption{Dataset information and grip prediction RMSE for different models on the validation set, test set, and separate test drives. Different data modalities are abbreviated where RGB denotes RGB camera, T denotes thermal camera and R denotes LiDAR reflectance measurements. The best-achieved error in each set is in bold and the second-to-best is underlined.}
\label{tab:all_results}
\begin{tabular}{|l|r|r|r|r|r|} 
\cline{2-6}
\multicolumn{1}{l|}{}                                                        & \multicolumn{1}{l|}{\textbf{Validation set}} & \multicolumn{1}{l|}{\textbf{Test set}} & \multicolumn{1}{l|}{\textbf{Test drive 1}}                                          & \multicolumn{1}{l|}{\textbf{Test drive 2}} & \multicolumn{1}{l|}{\textbf{Test drive 3}}  \\ 
\hline
\begin{tabular}[c]{@{}l@{}}\textbf{Weather}\\\textbf{condition}\end{tabular} & \multicolumn{1}{c|}{Varying}                 & \multicolumn{1}{c|}{Varying}           & \multicolumn{1}{c|}{\begin{tabular}[c]{@{}c@{}}Snowy,\\snowfall, dark\end{tabular}} & \multicolumn{1}{c|}{Snowy}                 & \multicolumn{1}{c|}{Wet, slushy}            \\ 
\hline
\textbf{Grip mean}                                                           & 0.6474                                          & 0.659                                    & 0.399                                                                                 & 0.557                                        & 0.649                                         \\ 
\hline
\textbf{Grip SD}                                                            & 0.2037                                          & 0.201                                    & 0.104                                                                                 & 0.140                                        & 0.159                                         \\ 
\hline
\textbf{\# samples}                                                            & 15\ 343                                          & 26\ 783 
& 5\ 746                                                                                 & 2\ 042                                        & 8\ 351                                         \\ 
\hline
\textbf{Modalities}                                                          & \multicolumn{1}{l|}{\textbf{RMSE}}           & \multicolumn{1}{l|}{\textbf{RMSE}}     & \multicolumn{1}{l|}{\textbf{RMSE}}                                                  & \multicolumn{1}{l|}{\textbf{RMSE}}         & \multicolumn{1}{l|}{\textbf{RMSE}}          \\ 
\hline
RGB                                                                          & 0.0657                                         & 0.0589                                   & 0.1041                                                                                 & 0.1497                                        & 0.1062                                         \\ 
\hline
T                                                                            & 0.0794                                          & 0.0772                                    & 0.1248                                                                                 & 0.1670                                        & 0.1361                                         \\ 
\hline
R                                                                            & 0.0677                                          & 0.0591                                    & \underline{0.0992}                                                                                 & 0.1262                                        & 0.0944                                         \\ 
\hline
RGB + T                                                                      & 0.0655                                          & 0.0605                                    & 0.1024                                                                                 & 0.1416                                        & 0.1069                                         \\ 
\hline
RGB + R                                                                      & \underline{0.0638}                                          & \textbf{0.0565}                                    & 0.1038                                                                                 & 0.1418                                        & \underline{0.0917}                                         \\ 
\hline
T + R                                                                      & 0.0664                                          & 0.0586                                    & 0.1056                                                                                 & \textbf{0.1038}                                        & \textbf{0.0906}                                         \\
\hline
RGB + T + R                                                                  & \textbf{0.0632}                                          & \underline{0.0575}                                    & \textbf{0.0974}                                                                                 & \underline{0.1118}                                        & 0.0994                                         \\
\hline
\end{tabular}
\end{table}

The separate test drive results confirm that the use of several data modalities improves the accuracy and the models have not noticeably overfit to the training data. Even though the standard deviation in each test drive is close to the model errors, it should be noted that the conditions in a single test drive are mostly constant and the models have predicted at least the general conditions in the test drive. However, there is variance and inconsistency in the separate test drive results as the amount of data is small, and adverse effects in one modality could decrease the performance of a single model. Some differences between the results could also be explained by the specific driving conditions, as the dark conditions in Test drive 1 might benefit the performance of the models using reflectance.

\begin{figure}[t]
    \centering
    \subfloat{\includegraphics[width=0.4\textwidth]{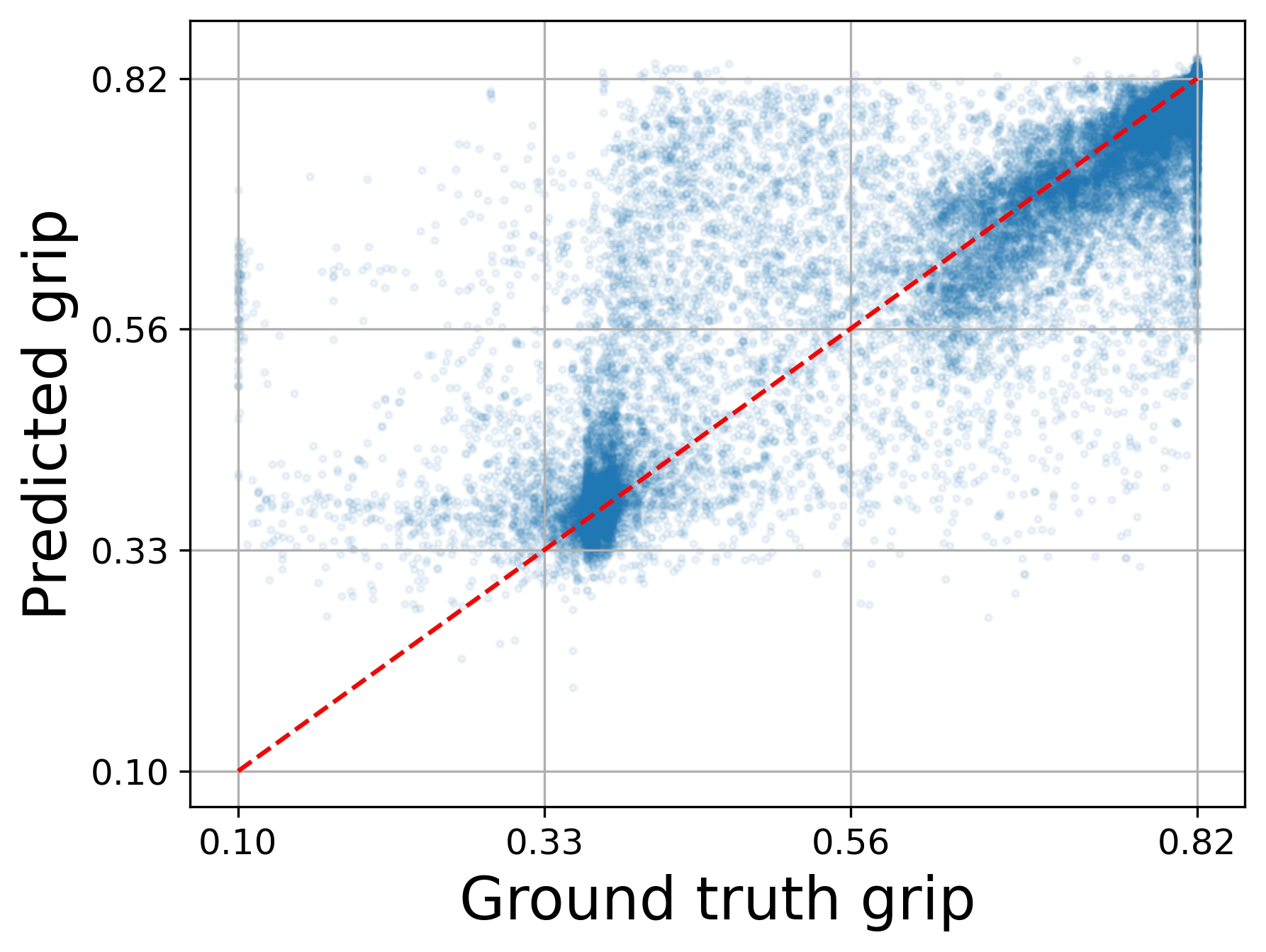}}
    \hspace{0.5cm}
    \subfloat{\includegraphics[width=0.4\textwidth]{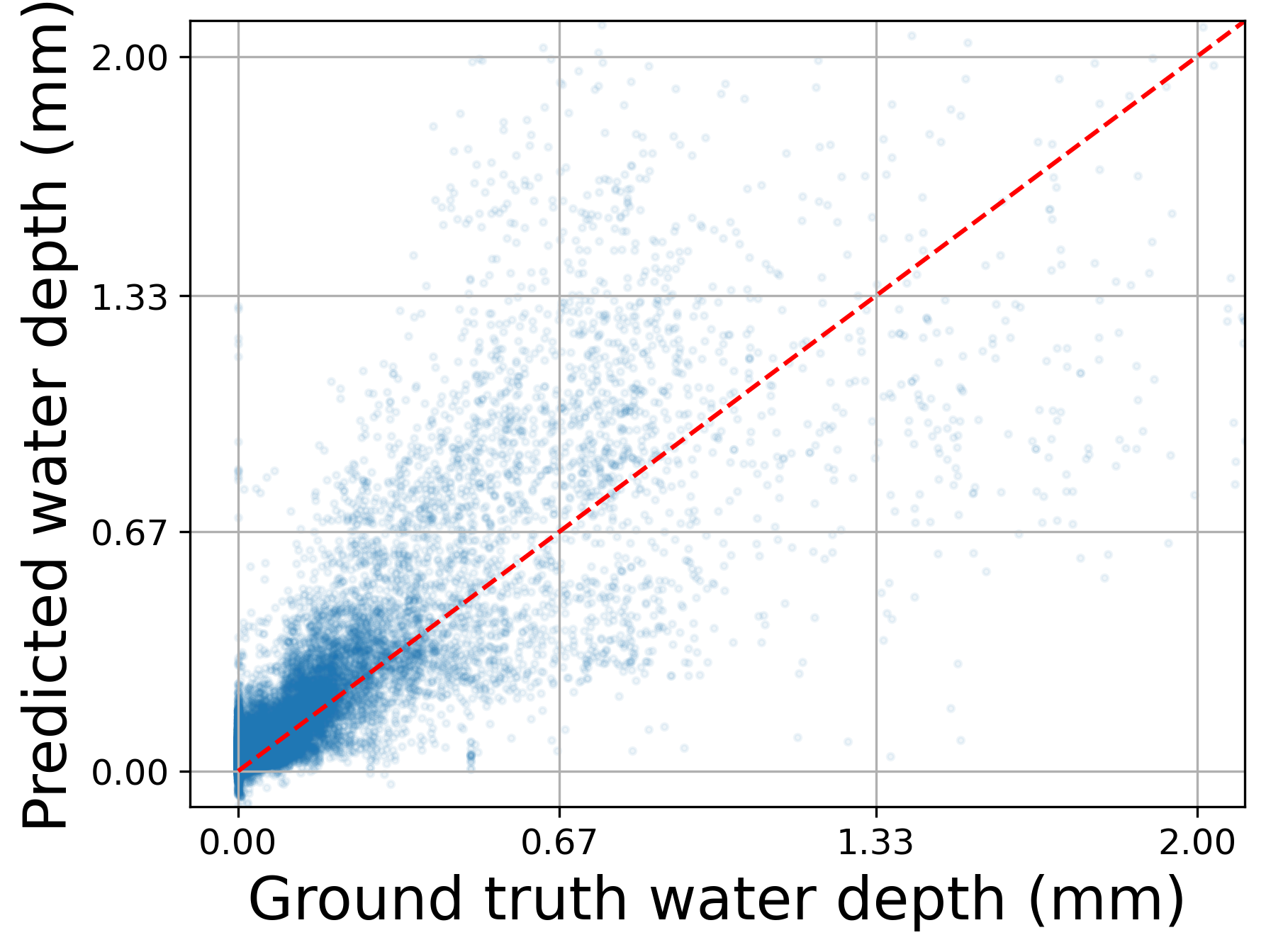}}\\
    \vspace{-0.2cm}
    \subfloat{\includegraphics[width=0.4\textwidth]{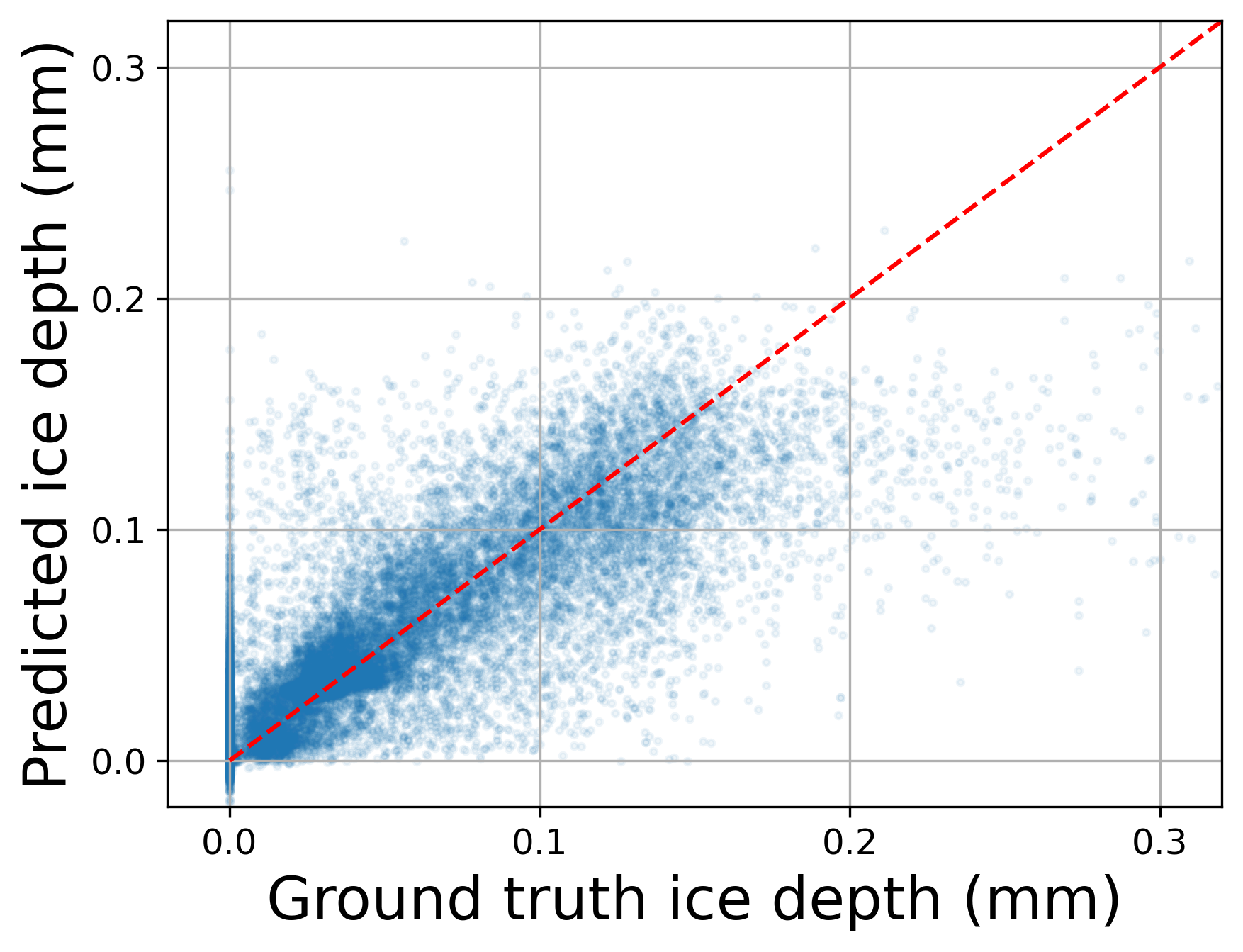}}
    \hspace{0.5cm}
    \subfloat{\includegraphics[width=0.4\textwidth]{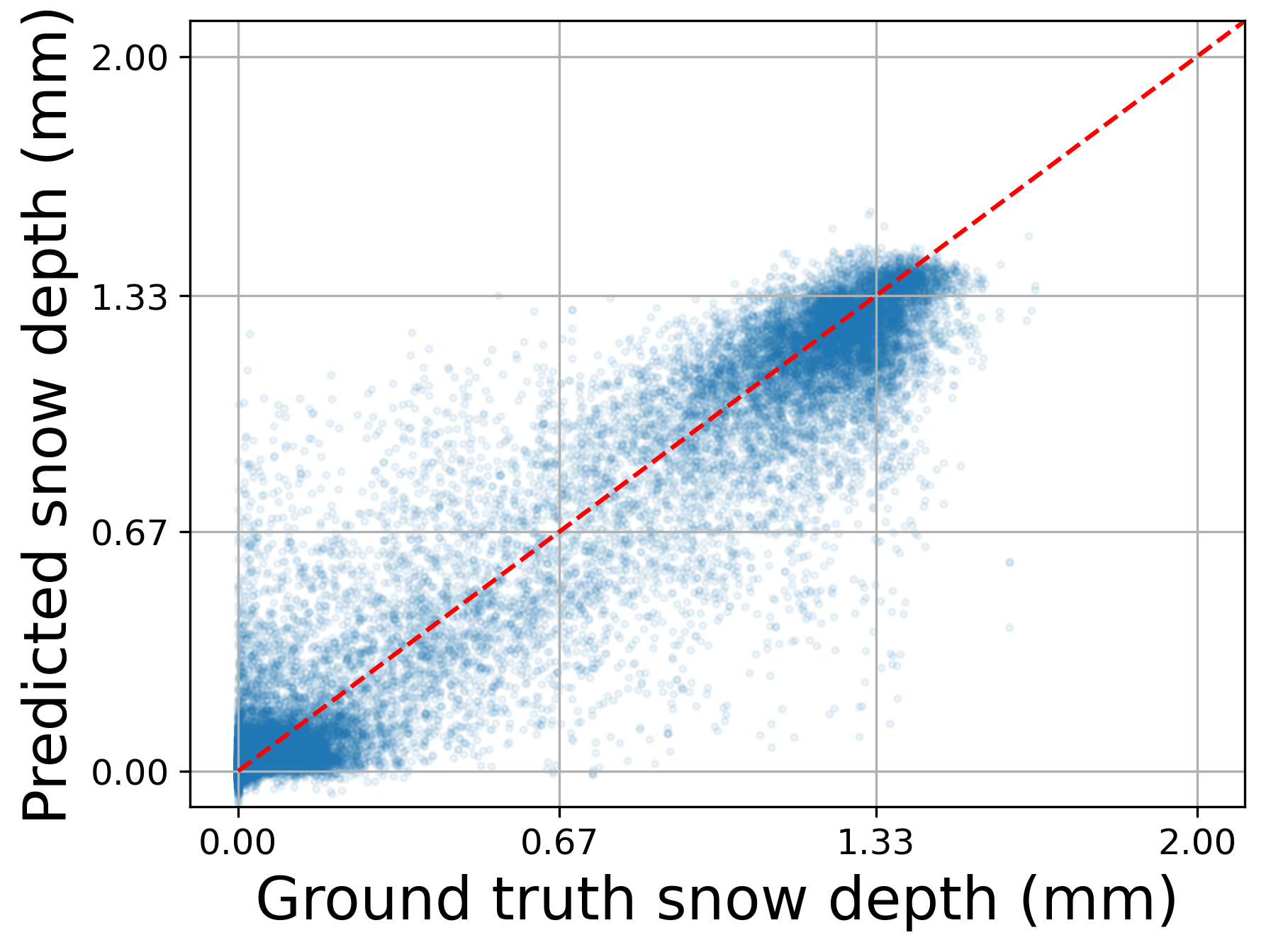}}
    \caption{Scatter plots of predicted grips and layer thicknesses produced by the best, proposed model (RGB+T+R). The x-axis represents the ground truth values and the y-axis the predictions. The plots were generated using 50\ 000 random measurements and corresponding predictions from the test set. The red dashed line represents the position of correct predictions.}
    \label{fig:scatters}
\end{figure}

In addition, a scatter plot of the grip and different surface layer thickness predictions in the test set is shown in Figure~\ref{fig:scatters}. The surface layer thickness predictions mostly follow the ground truth values within a relatively small error range while the predicted grip values have a larger error distribution, partly due to misinterpretation of snowy conditions.

\subsection{Qualitative Performance}

\begin{figure}[t]
    \centering
    
    \begin{tabular}{cccc}
    Ground truth grip &
    Predicted grip &
    Ground truth grip &
    Predicted grip 
     \\
    
    \includegraphics[width=0.24\textwidth]{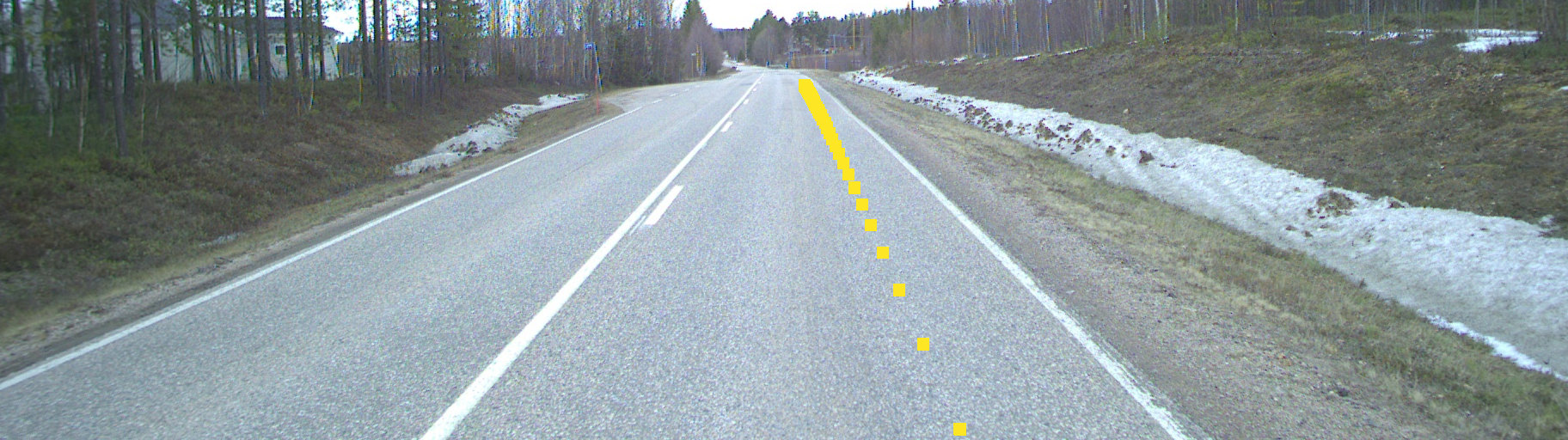} &
    \includegraphics[width=0.24\textwidth]{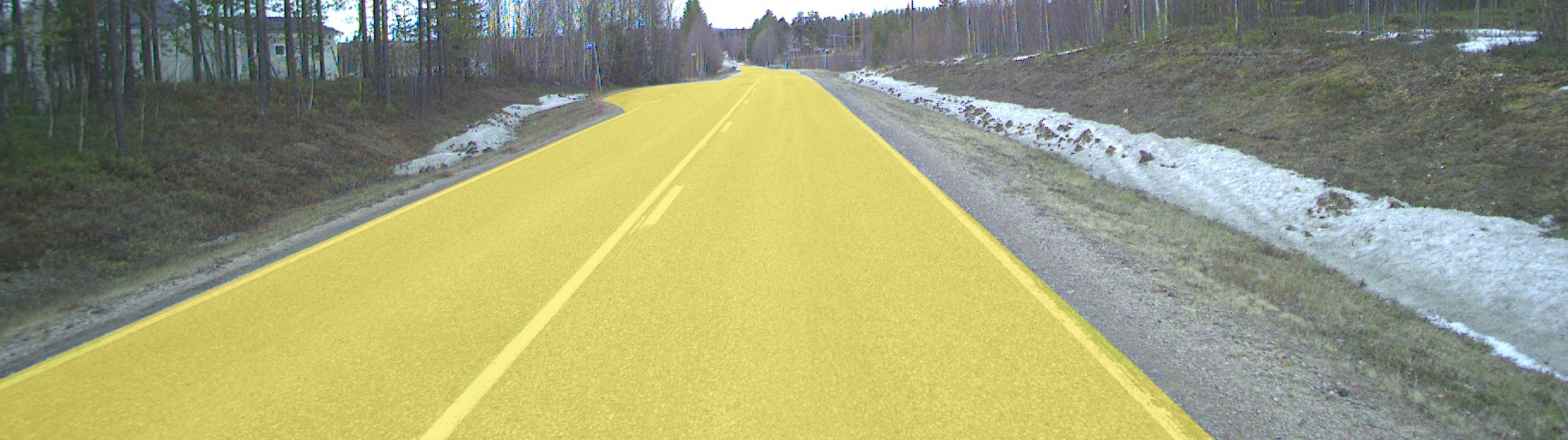}
    &
    \includegraphics[width=0.24\textwidth]{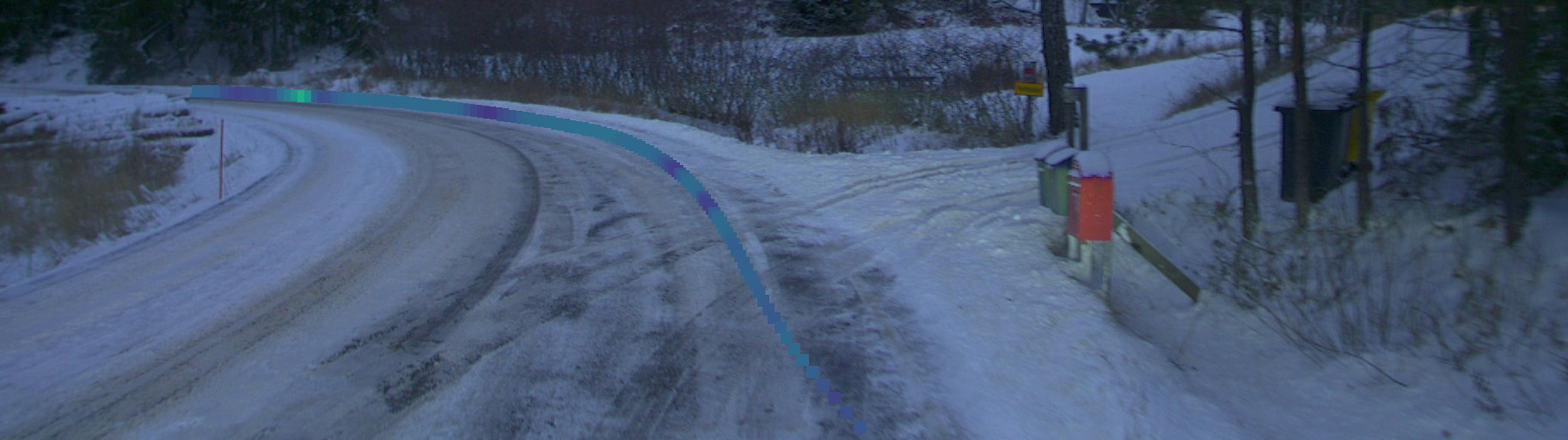}&
    \includegraphics[width=0.24\textwidth]{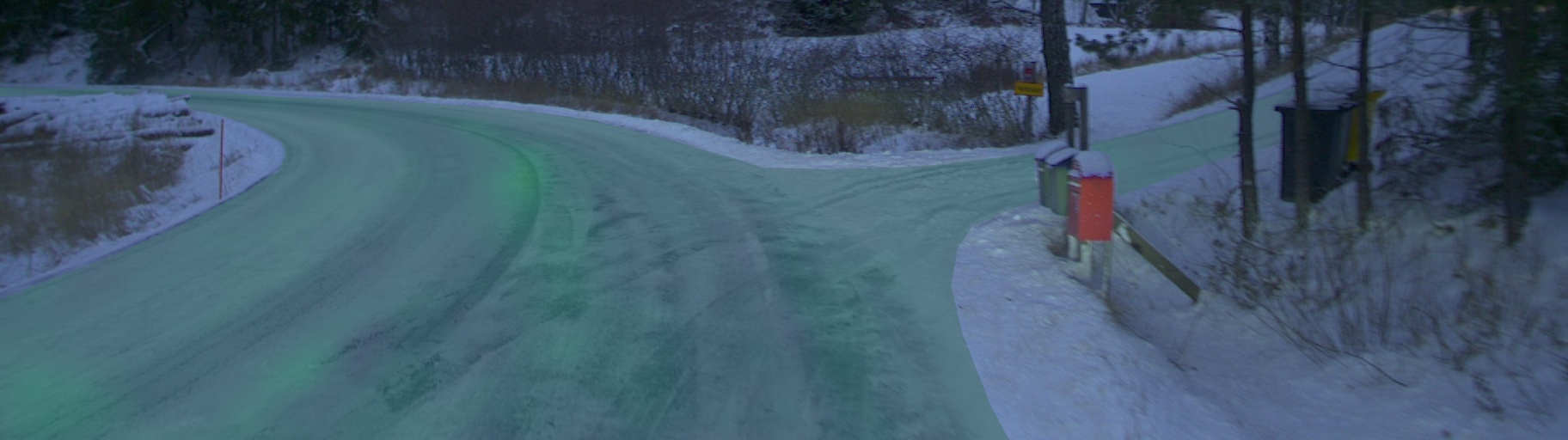}
     \\
    
    \includegraphics[width=0.24\textwidth]{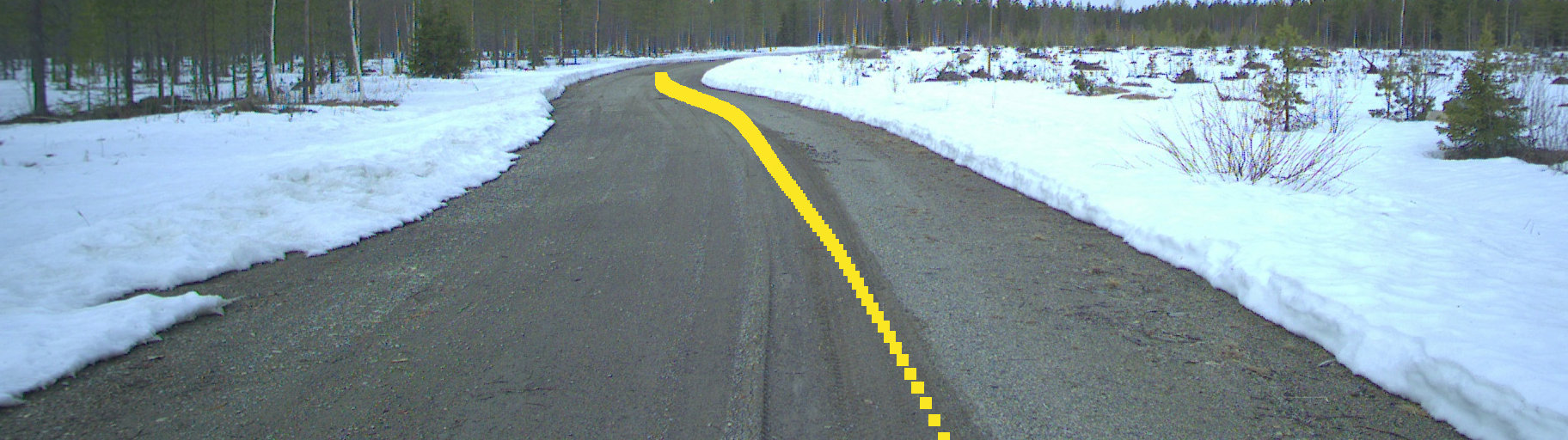}&
    \includegraphics[width=0.24\textwidth]{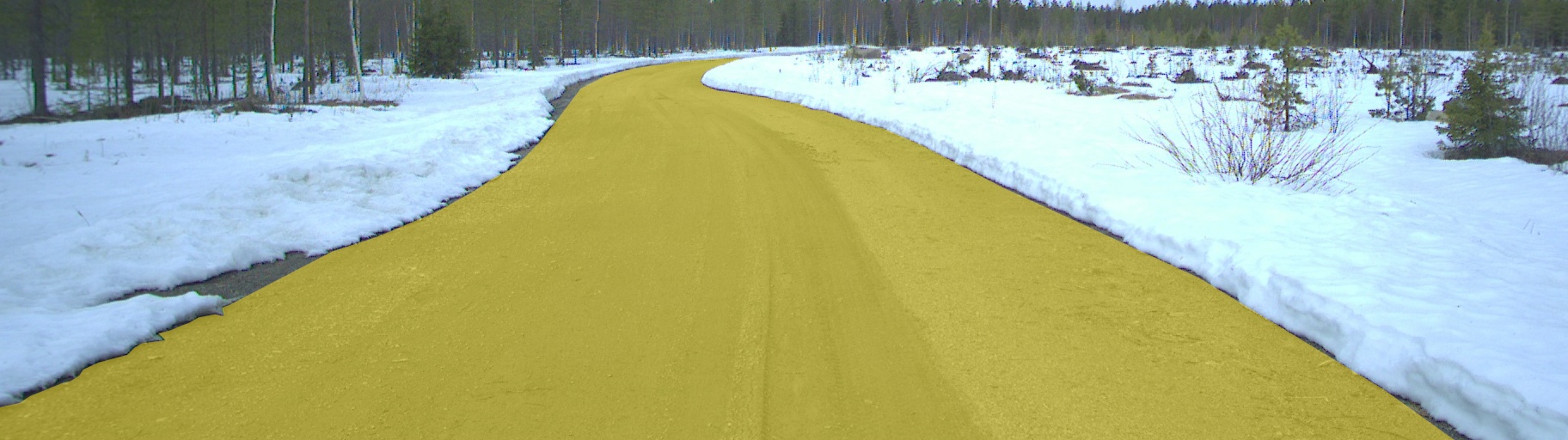}
    &
    \includegraphics[width=0.24\textwidth]{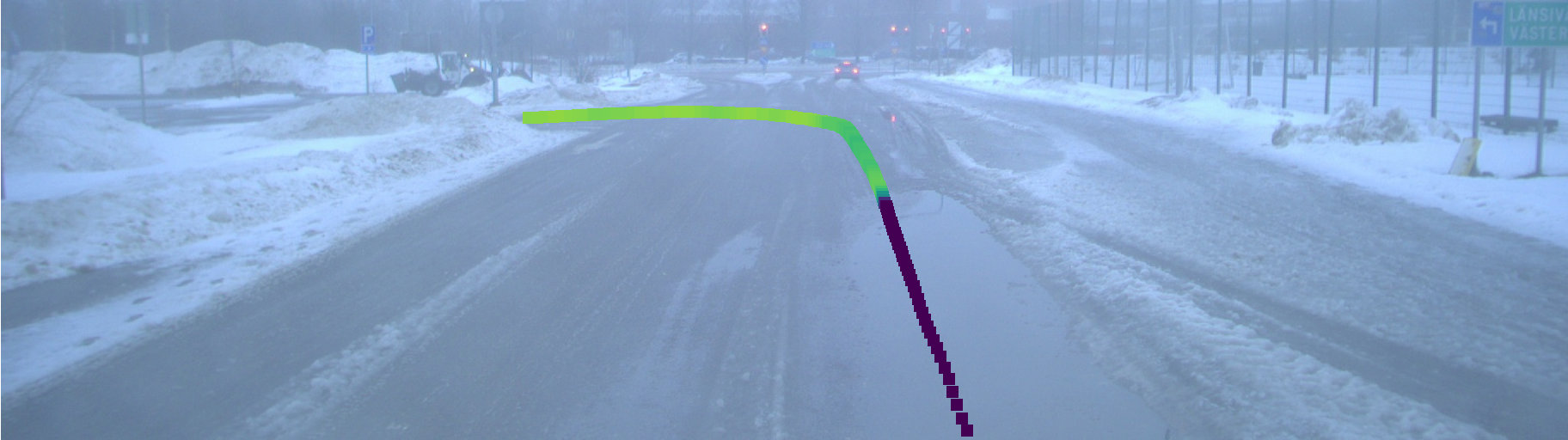}&
    \includegraphics[width=0.24\textwidth]{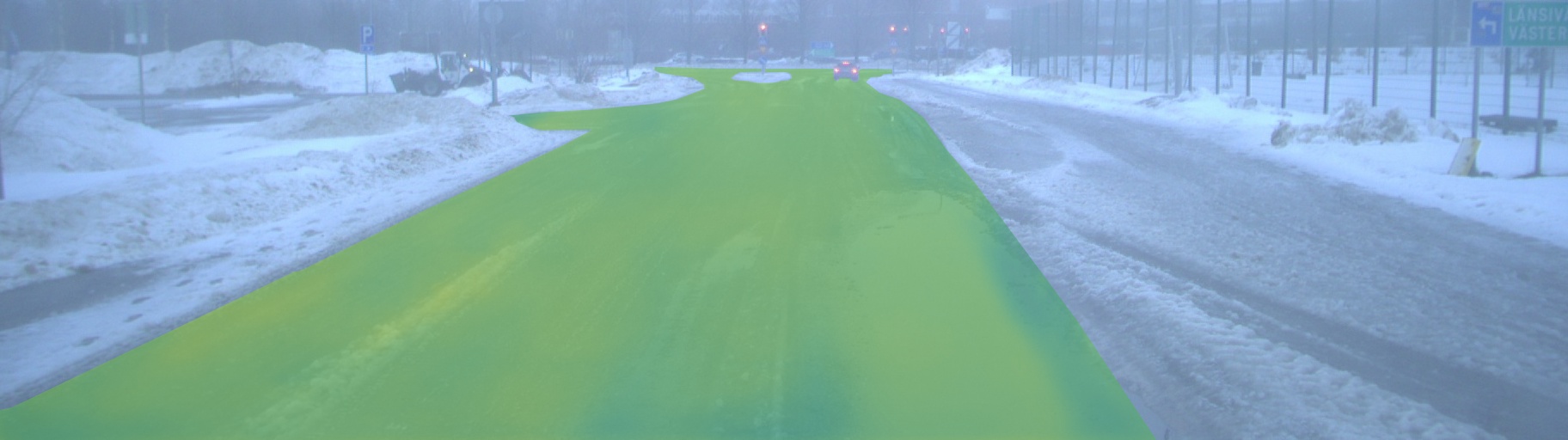}
     \\
    
    \includegraphics[width=0.24\textwidth]{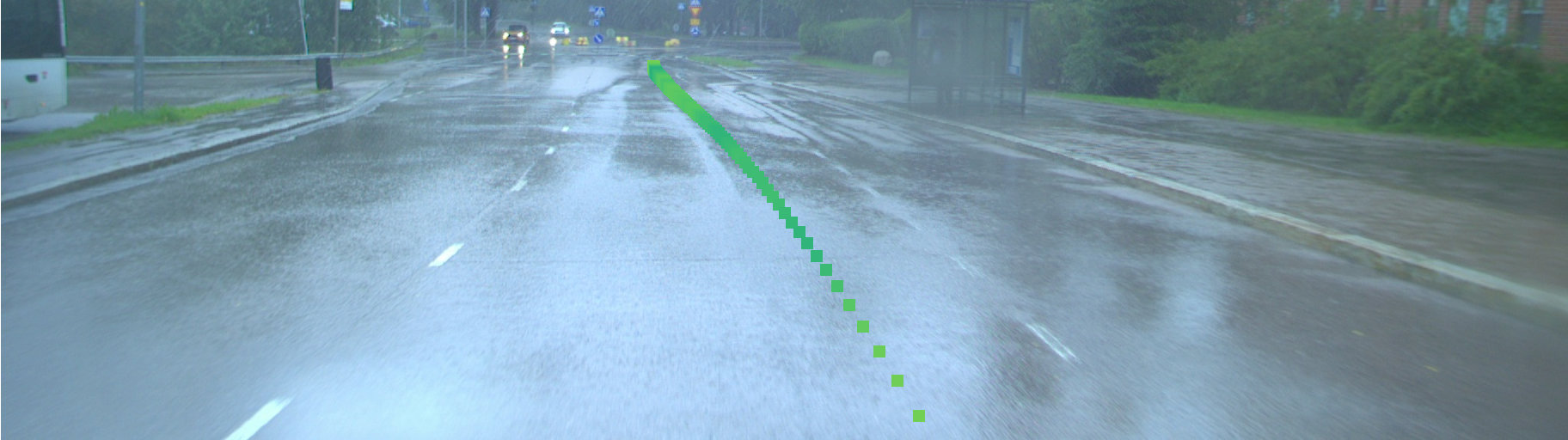} &
    \includegraphics[width=0.24\textwidth]{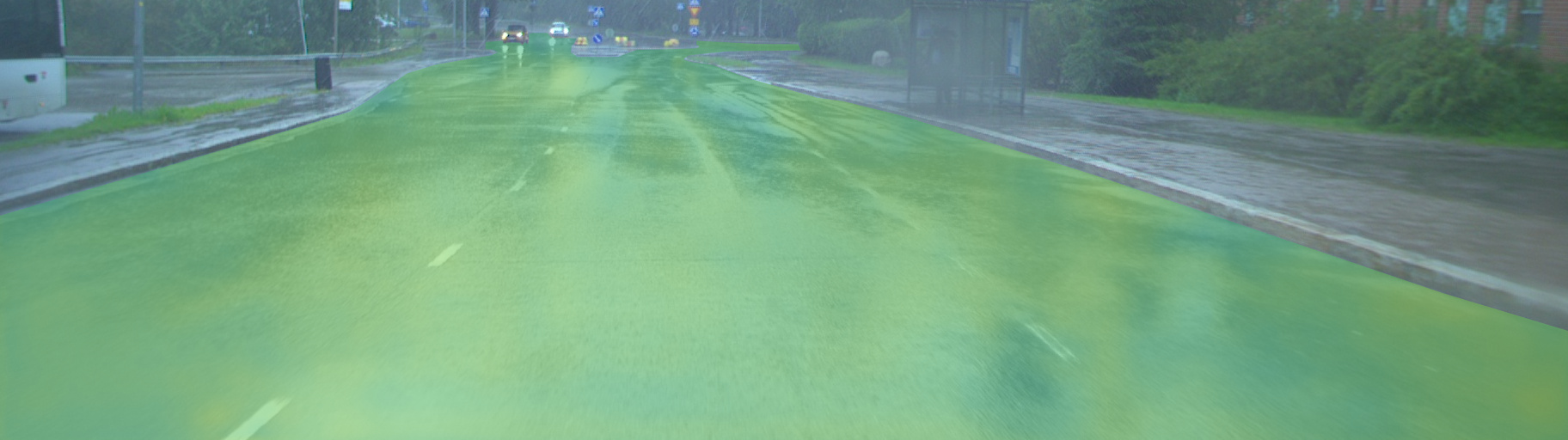}
    &
    \includegraphics[width=0.24\textwidth]{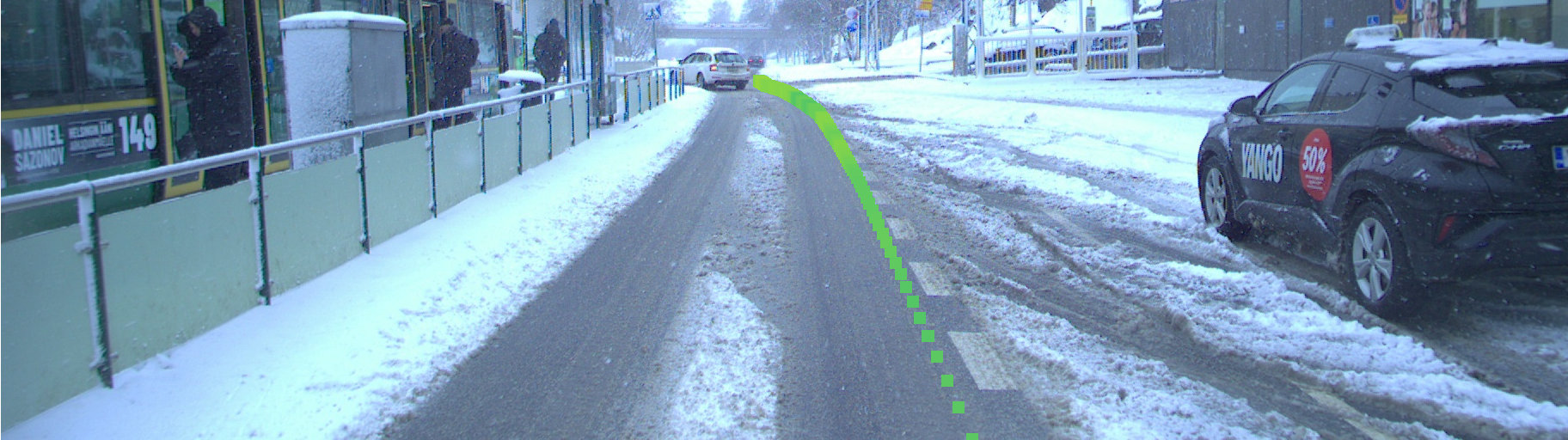}&
    \includegraphics[width=0.24\textwidth]{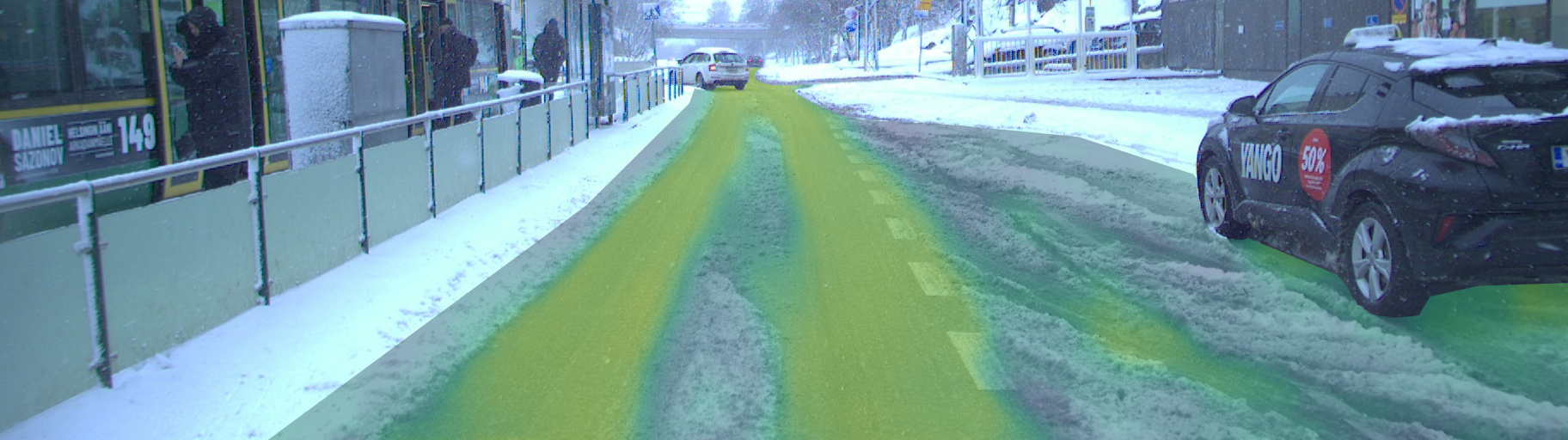}
     \\

    \includegraphics[width=0.24\textwidth]{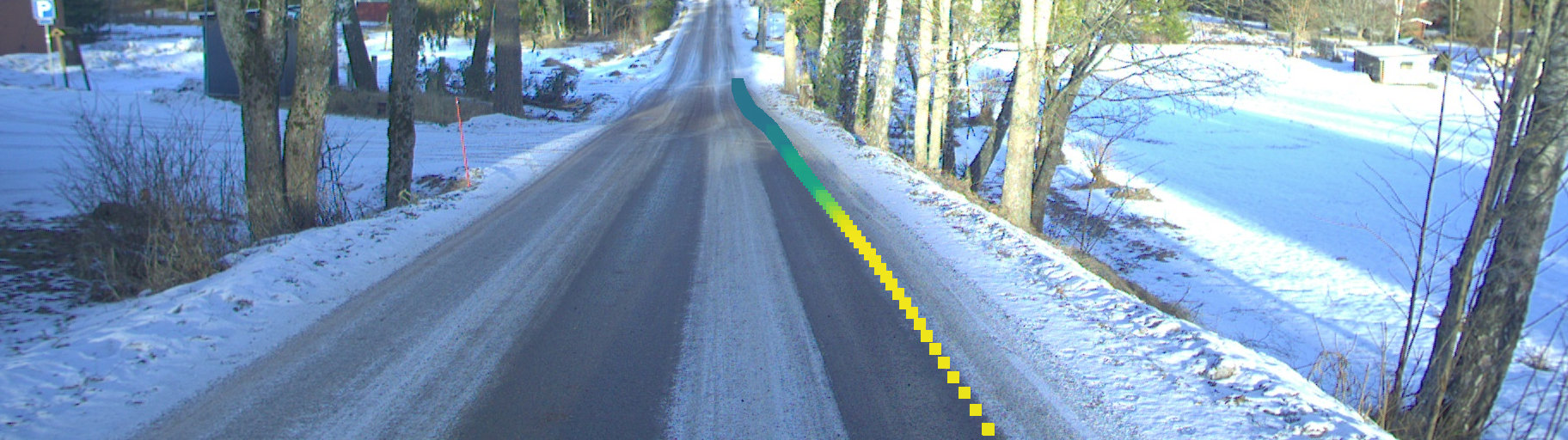} &
    \includegraphics[width=0.24\textwidth]{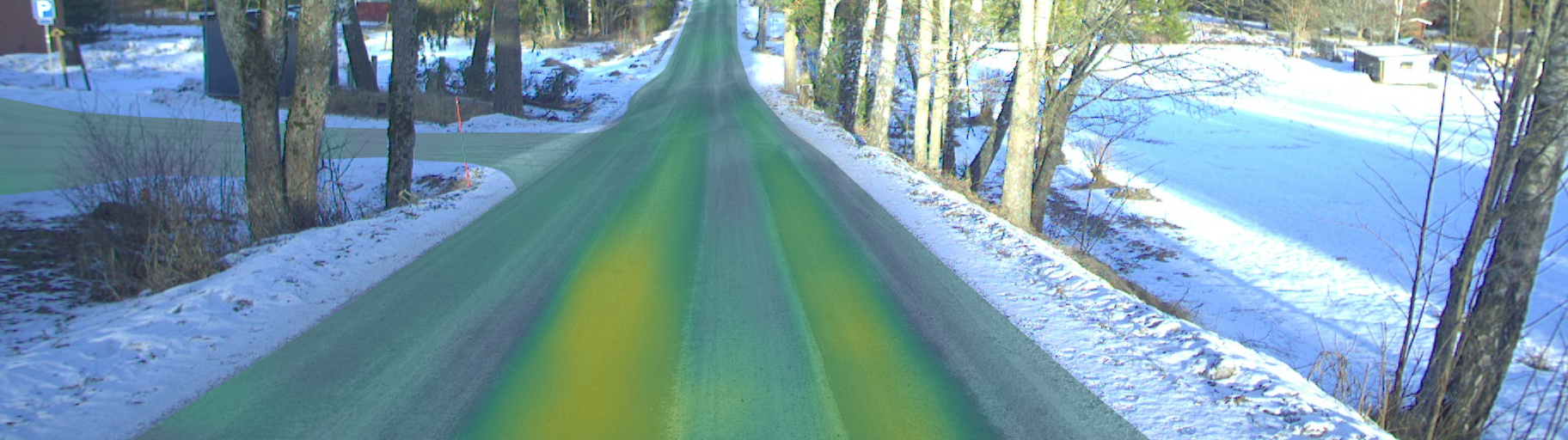}
    &
    \includegraphics[width=0.24\textwidth]{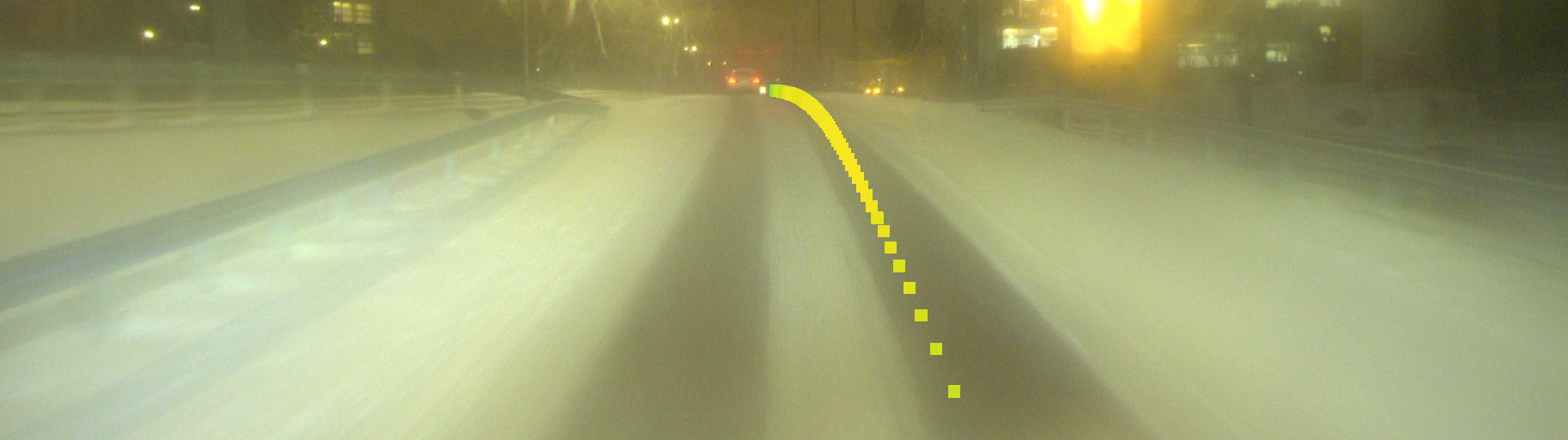}&
    \includegraphics[width=0.24\textwidth]{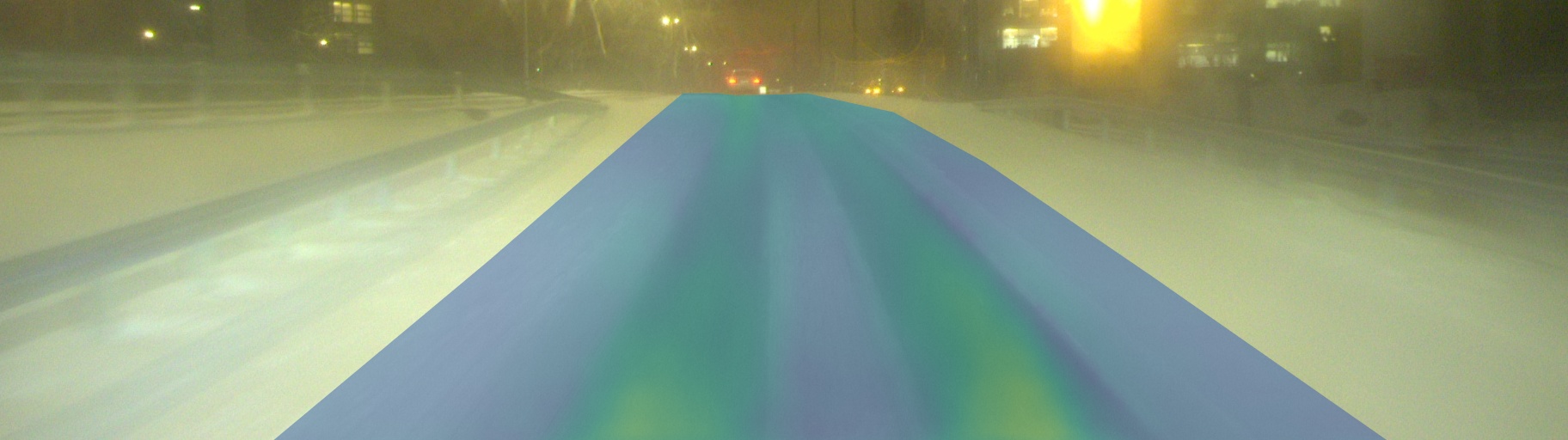}
     \\
    
    \end{tabular}
    \includegraphics[width=0.5\textwidth]{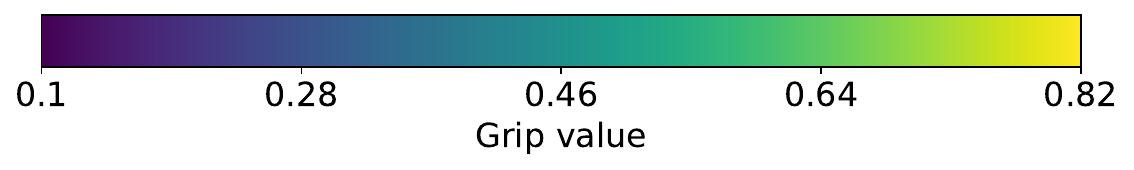}
    
    \caption{Visualisations of the qualitative performance of the final model (RGB+T+R). The ground truth labels are shown using 14-by-14-pixel colored squares drawn on the RGB input image.}
    \label{fig:best_outputs}
\end{figure}

Besides the error evaluation based on the ground truth road weather sensor measurements, we evaluated the grip map prediction over the complete road area qualitatively. Several example scenarios and grip map predictions from the final proposed model in different road weather conditions are shown in Figure~\ref{fig:best_outputs}. Additionally, examples from the other introduced models and a comparison of the impact of different modalities on the qualitative results are shown in Figure~\ref{fig:qualitative_mods} and in the supplementary material. In all figures in this work, the road area is segmented manually as the model does not differentiate the road area from the input data.

In general, we observe that the model output is smooth and is often constant when there are no variations in road weather conditions, such as when the road is completely dry, completely wet, or completely covered in snow. The model predictions could also mostly follow the boundaries between snowy and clear areas as seen in scenarios presented in Figure~\ref{fig:best_outputs} where clear tire tracks can be seen on otherwise snowy roads. Some conditions are still difficult to detect, such as the second scenario on the right column, in which the model could not detect the low grip of an area covered with deep water.

In addition, the model performance is unclear in some conditions that are further from the usual ground truth data locations, such as on the adjacent lane. The model output also could not follow sharp changes in grip values as the model seems to average the grip on relatively large prediction areas. This is likely due to the sparsity and varying data quality of the ground truth road weather measurements.

In Figure~\ref{fig:qualitative_mods} we show performance differences between models using different data modalities as inputs. In some examples, the thermal and reflectance-based single modality models misclassify the grip conditions of the whole scene as the data modality can not differentiate the current condition correctly. However, the model using each data modality seems to combine the correct predictions from the single modalities into a consistent representation of the grip map.

\begin{sidewaysfigure}
    \centering
    \begin{tabular}{ccccc}
Ground truth & RGB & Thermal & Reflectance & RGB+T+R \\

\subfloat{
\includegraphics[width=0.19\linewidth]{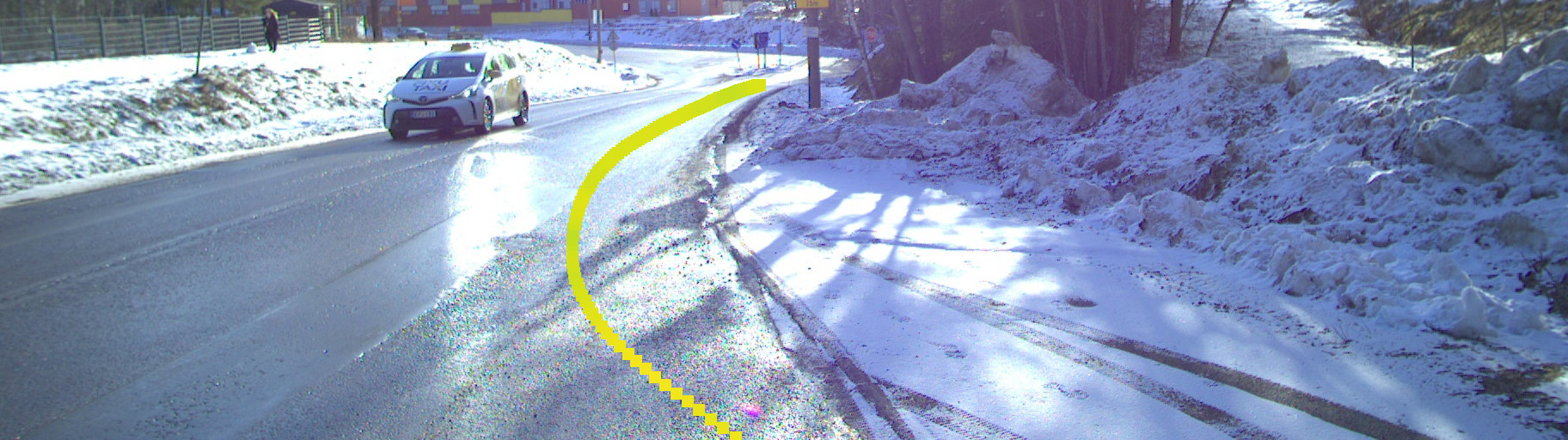}
}&
\subfloat{
\includegraphics[width=0.19\linewidth]{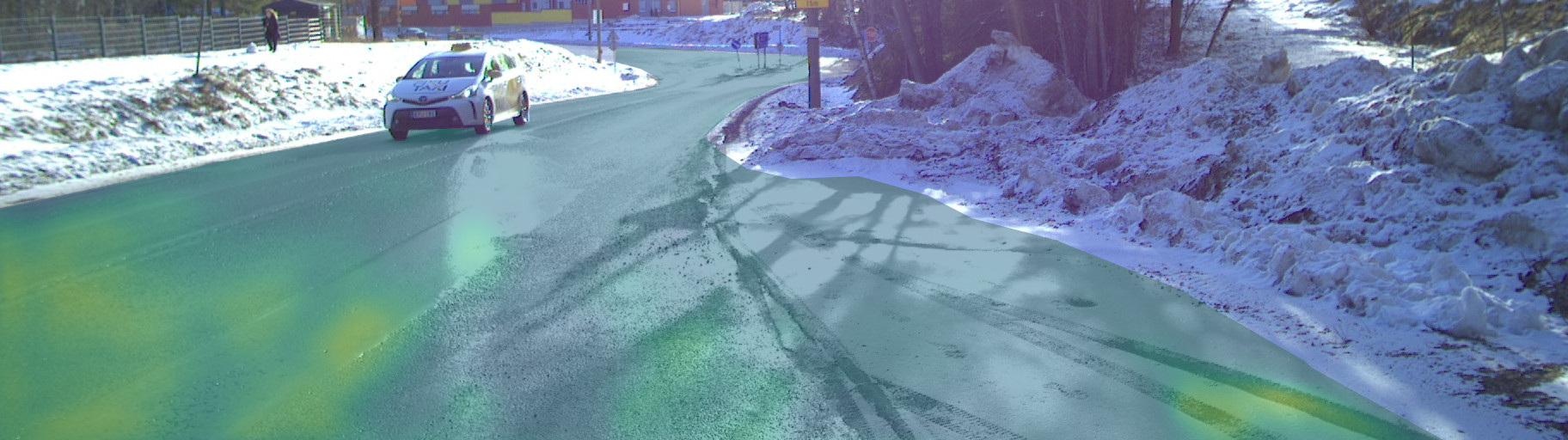}
}&
\subfloat{
\includegraphics[width=0.19\linewidth]{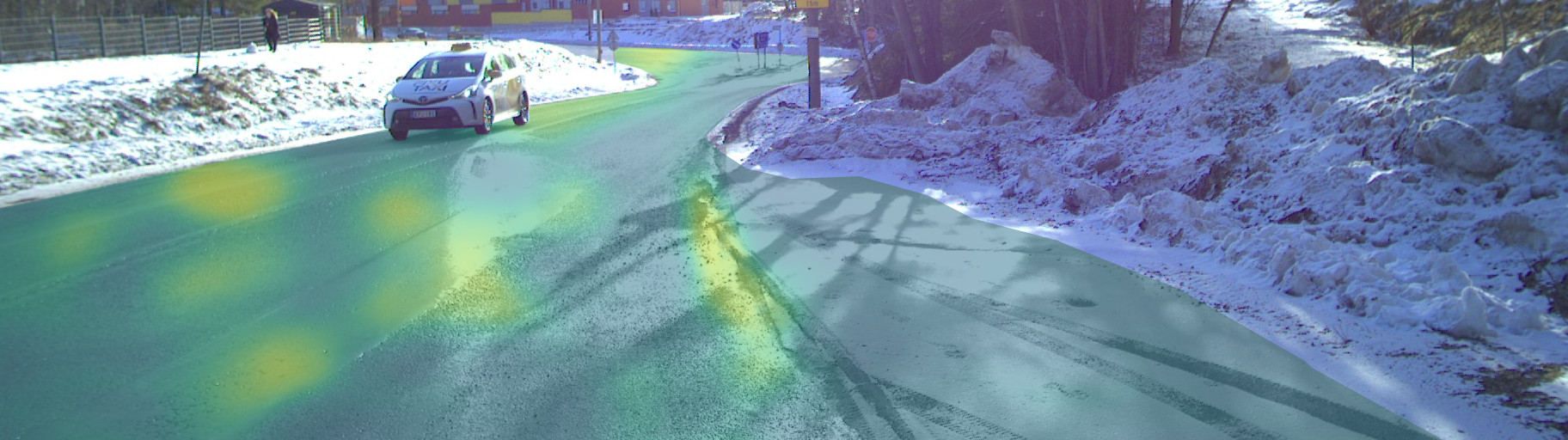}
}&
\subfloat{
\includegraphics[width=0.19\linewidth]{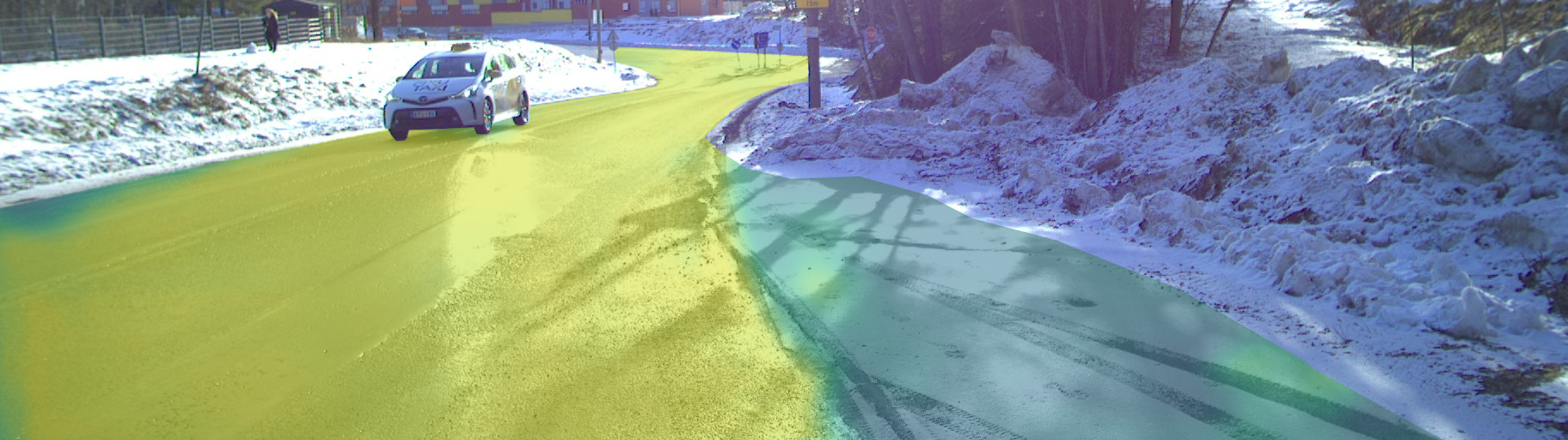}
}&
\subfloat{
\includegraphics[width=0.19\linewidth]{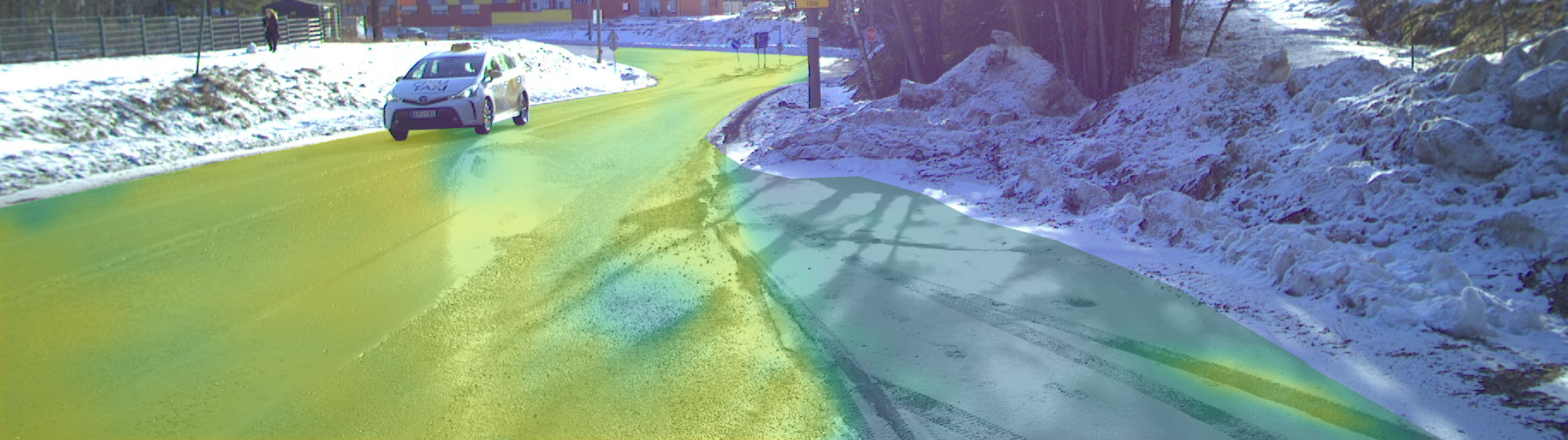}
} \\

\subfloat{
\includegraphics[width=0.19\linewidth]{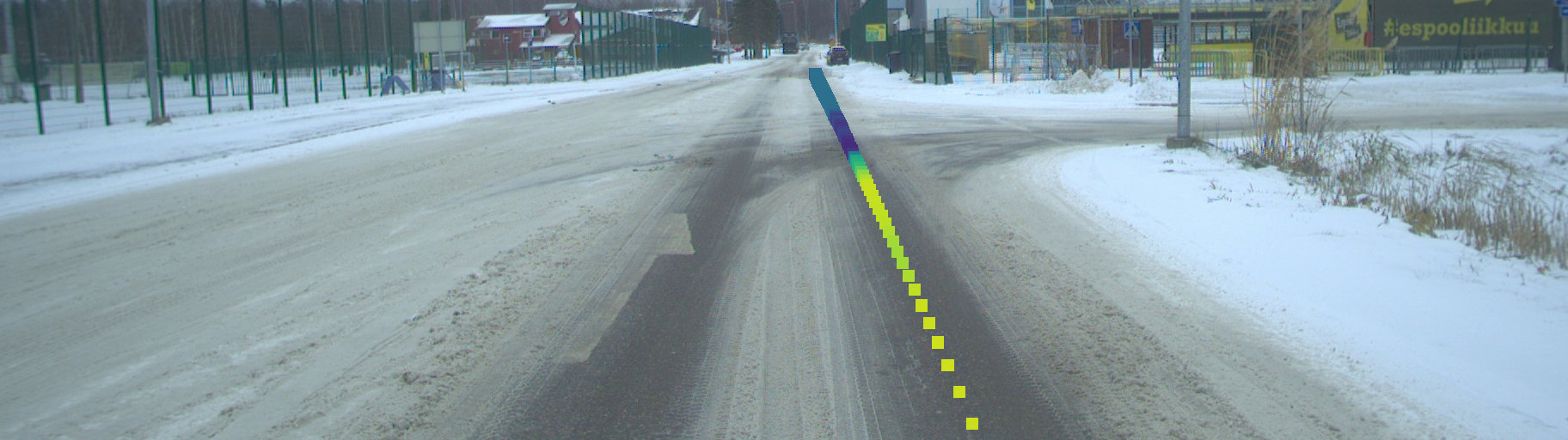}
}&
\subfloat{
\includegraphics[width=0.19\linewidth]{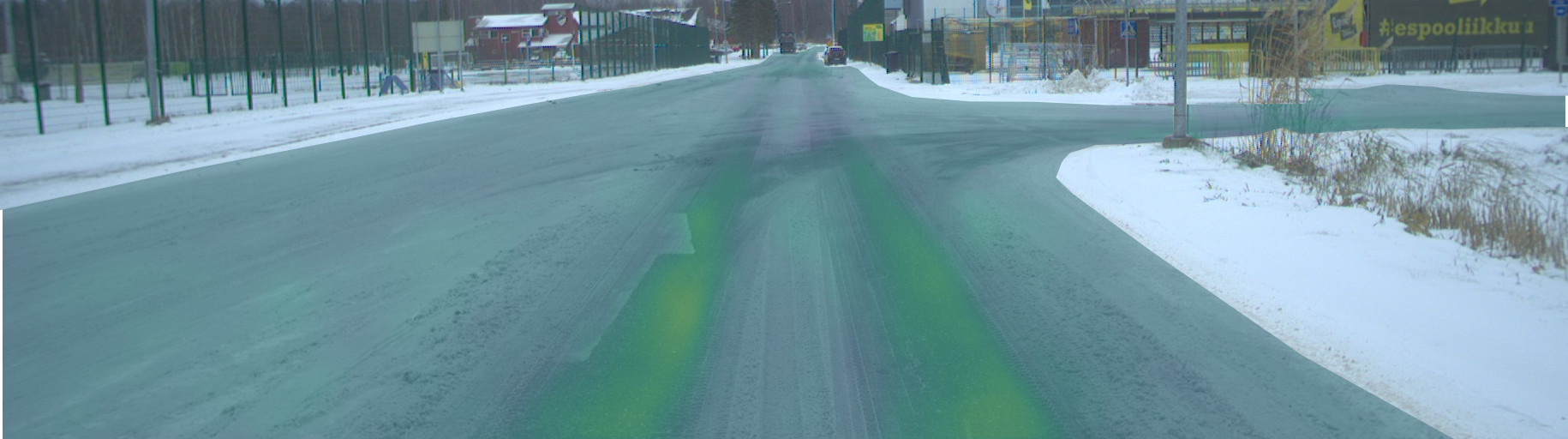}
}&
\subfloat{
\includegraphics[width=0.19\linewidth]{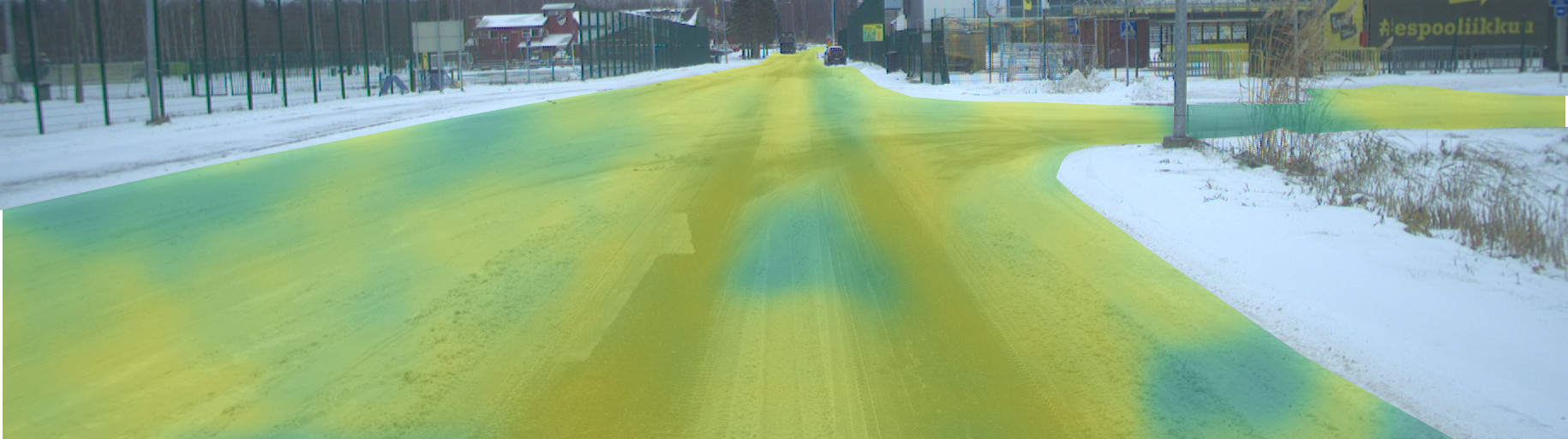}
}&
\subfloat{
\includegraphics[width=0.19\linewidth]{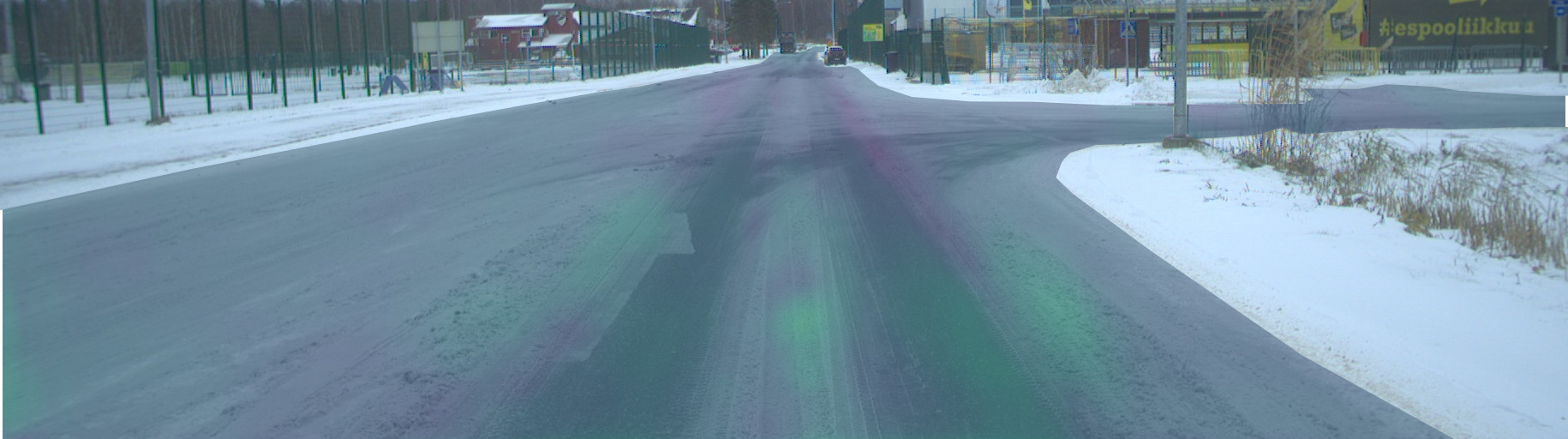}
}&
\subfloat{
\includegraphics[width=0.19\linewidth]{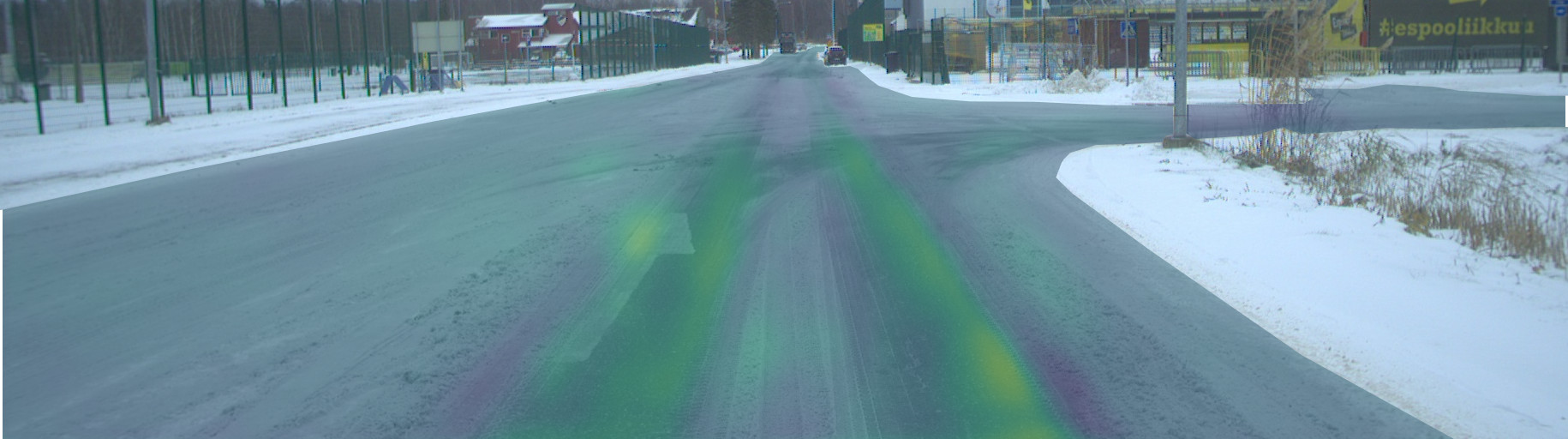}
} \\

\subfloat{
\includegraphics[width=0.19\linewidth]{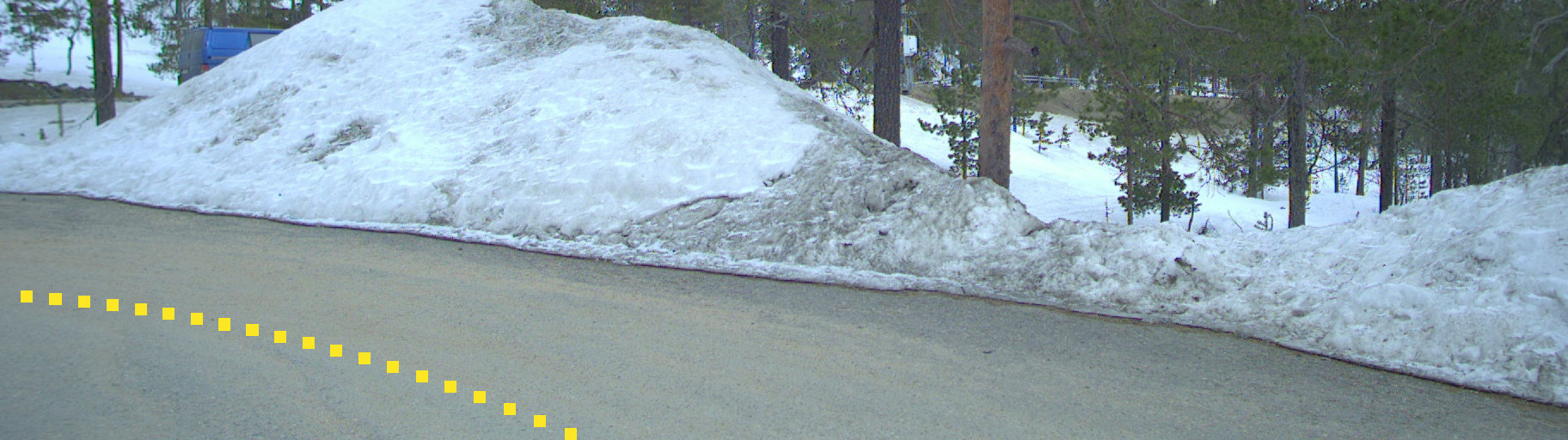}
}&
\subfloat{
\includegraphics[width=0.19\linewidth]{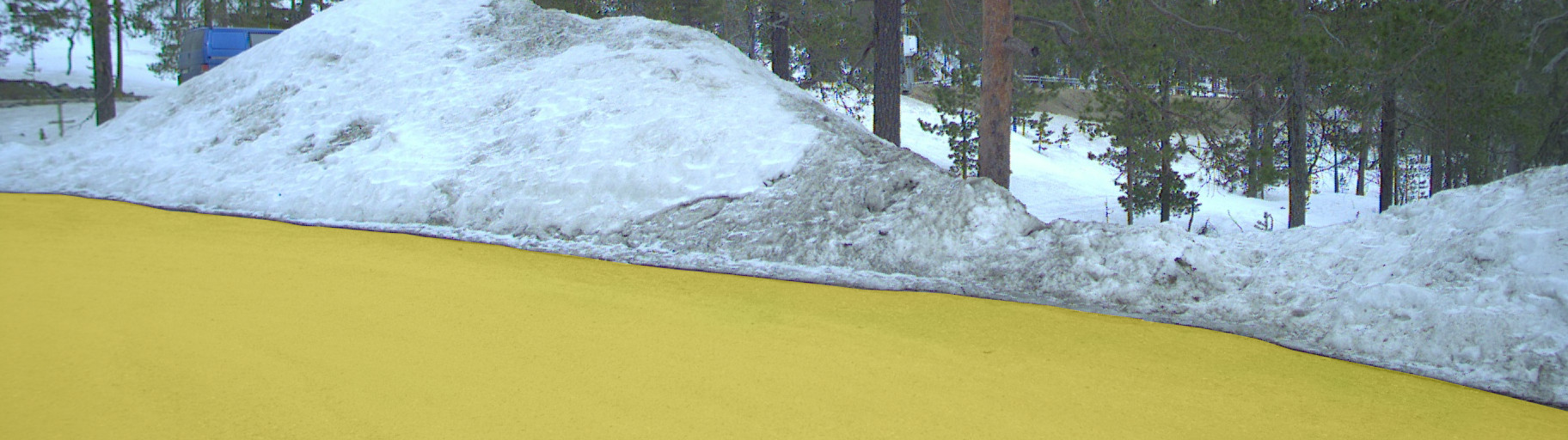}
}&
\subfloat{
\includegraphics[width=0.19\linewidth]{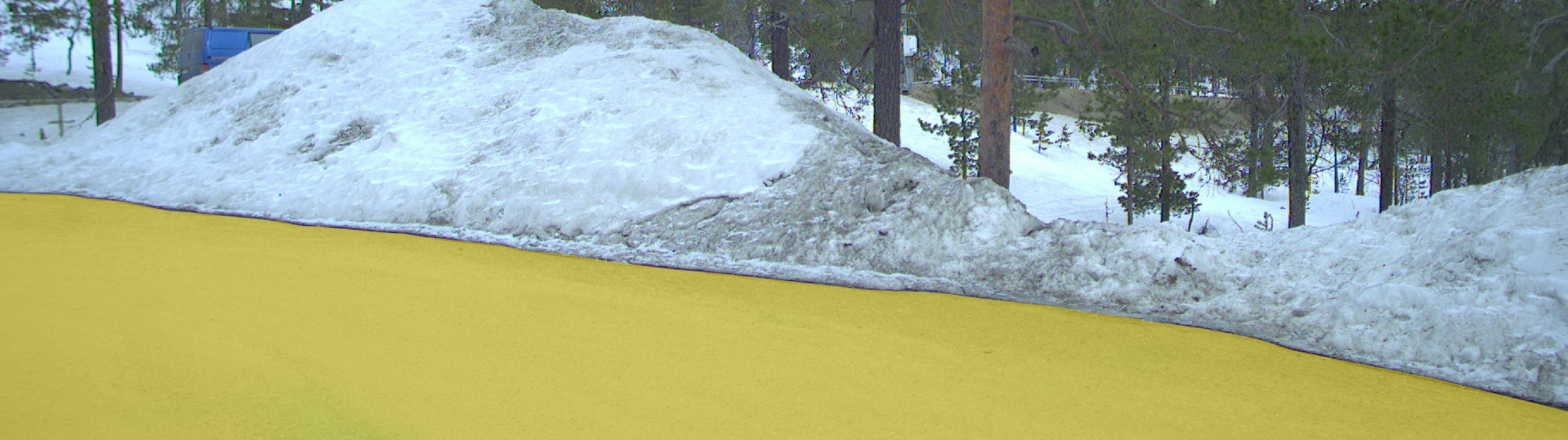}
}&
\subfloat{
\includegraphics[width=0.19\linewidth]{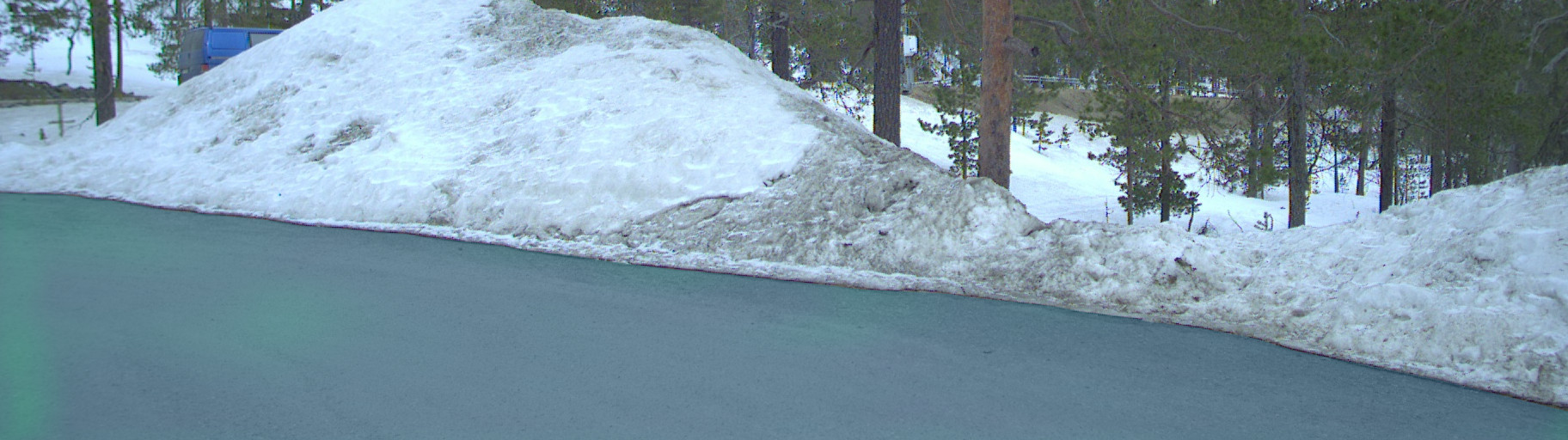}
}&
\subfloat{
\includegraphics[width=0.19\linewidth]{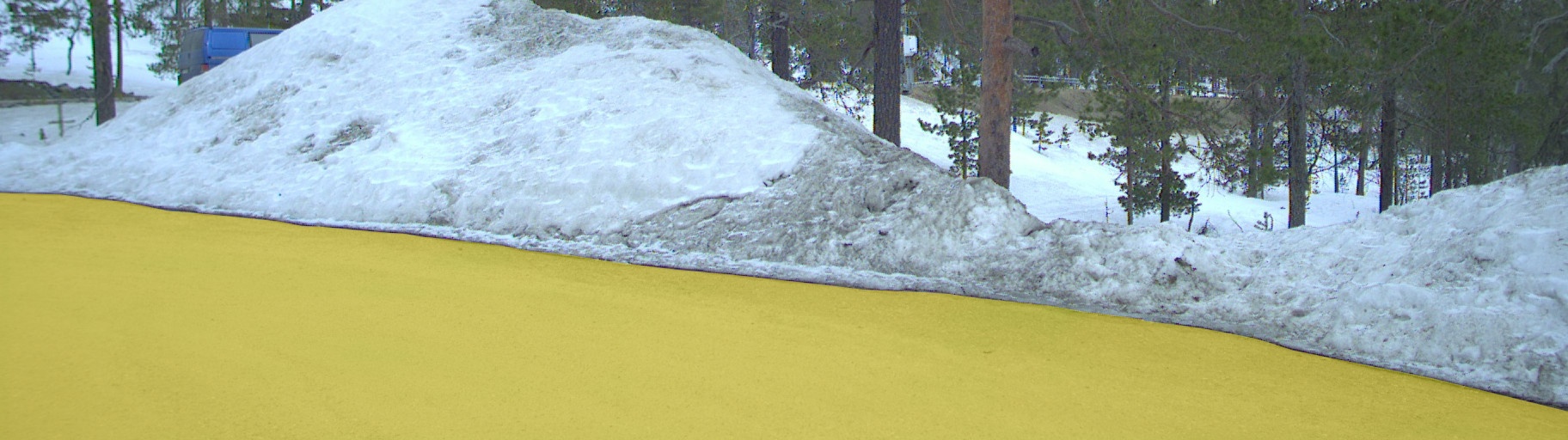}
} \\

\subfloat{
\includegraphics[width=0.19\linewidth]{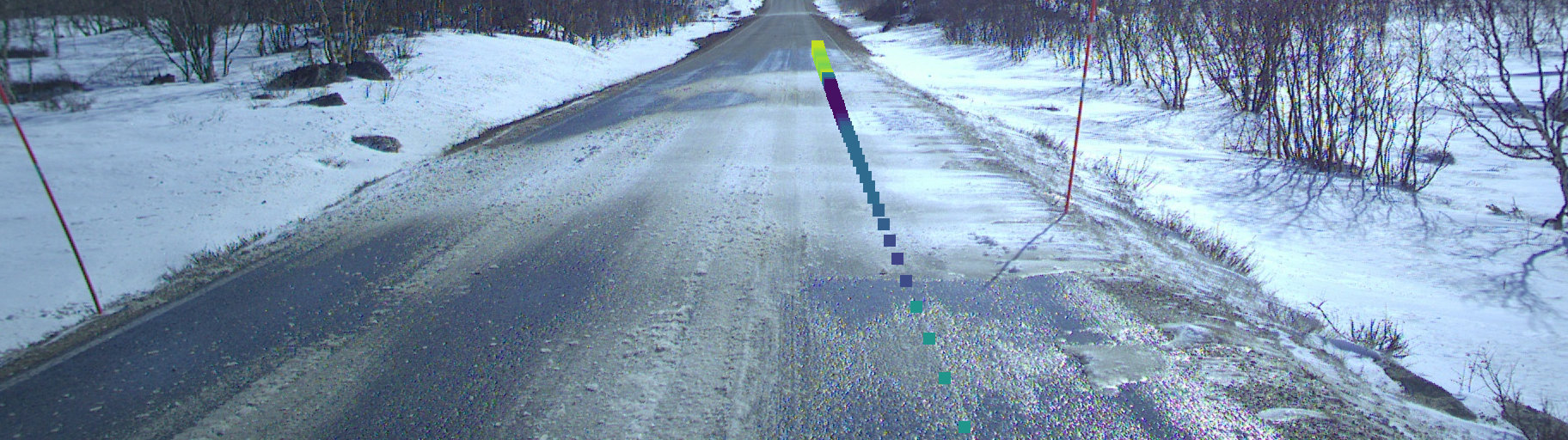}
}&
\subfloat{
\includegraphics[width=0.19\linewidth]{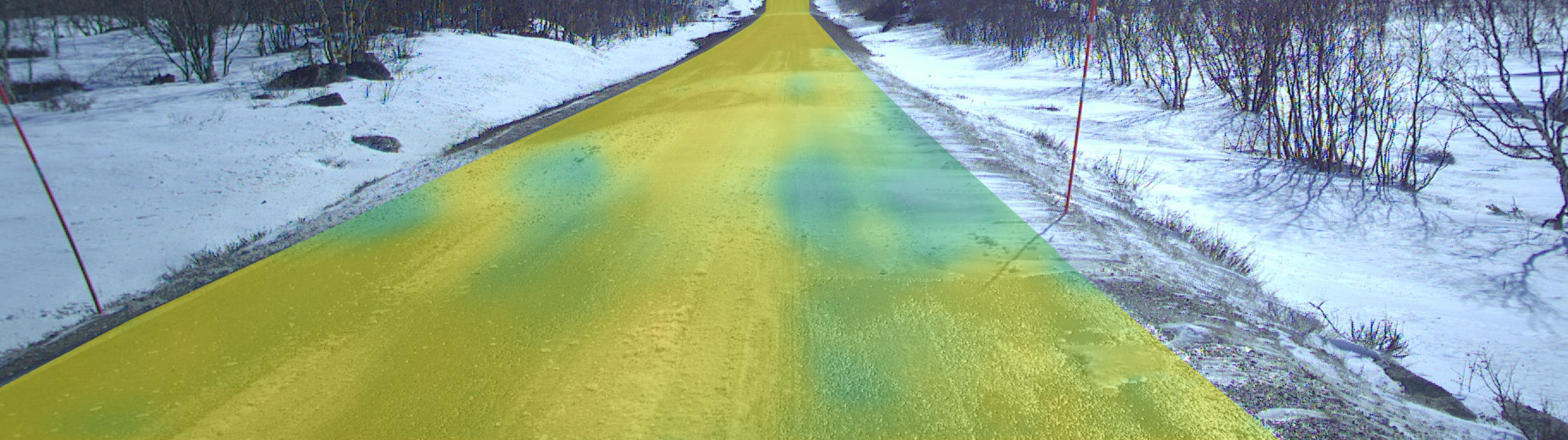}
}&
\subfloat{
\includegraphics[width=0.19\linewidth]{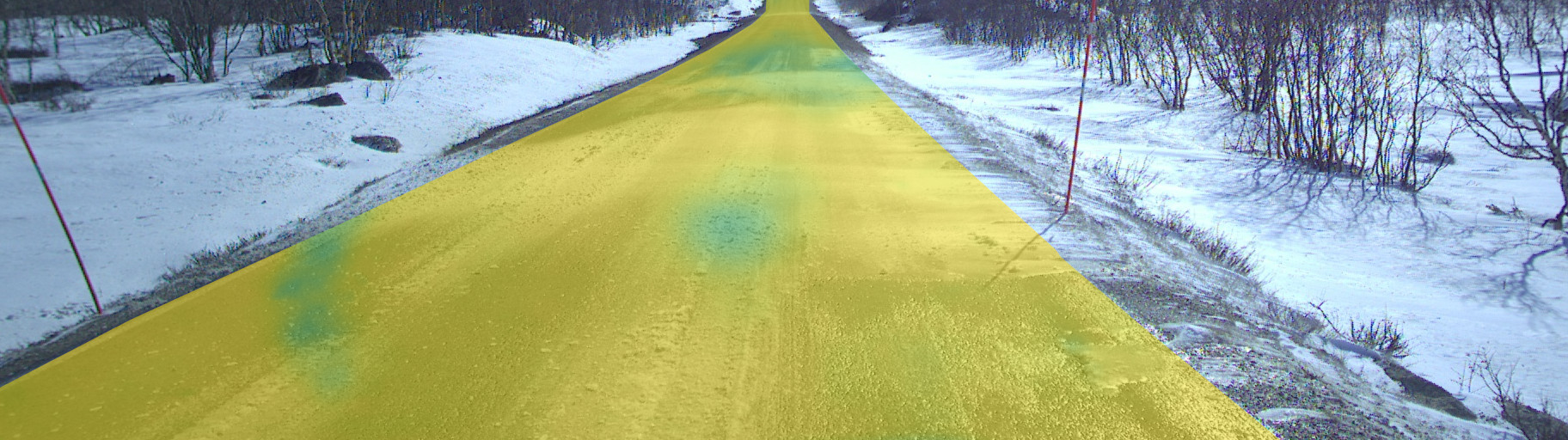}
}&
\subfloat{
\includegraphics[width=0.19\linewidth]{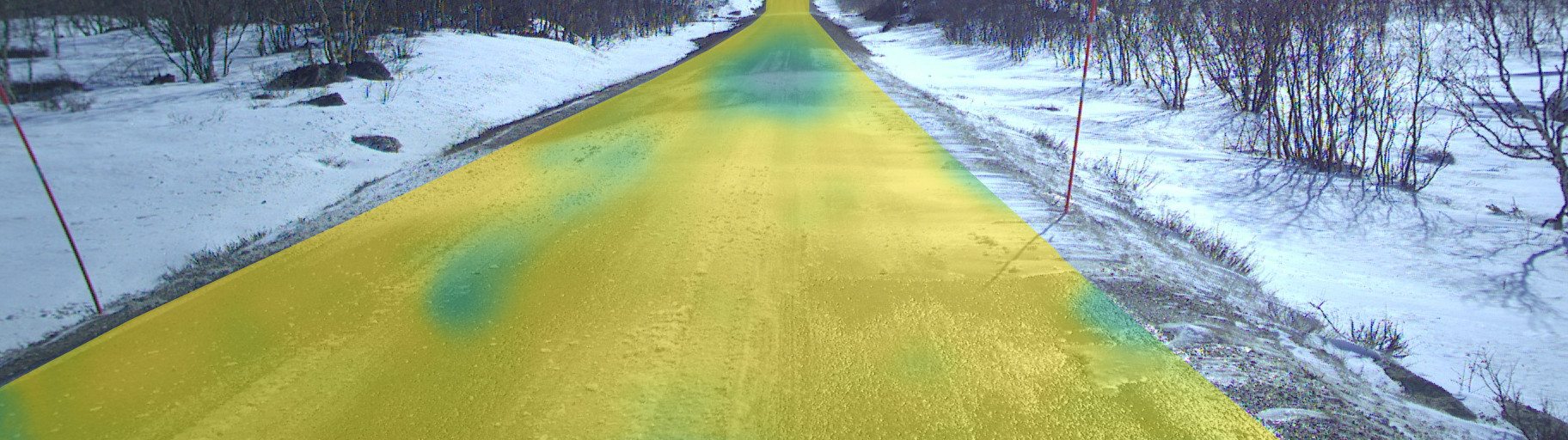}
}&
\subfloat{
\includegraphics[width=0.19\linewidth]{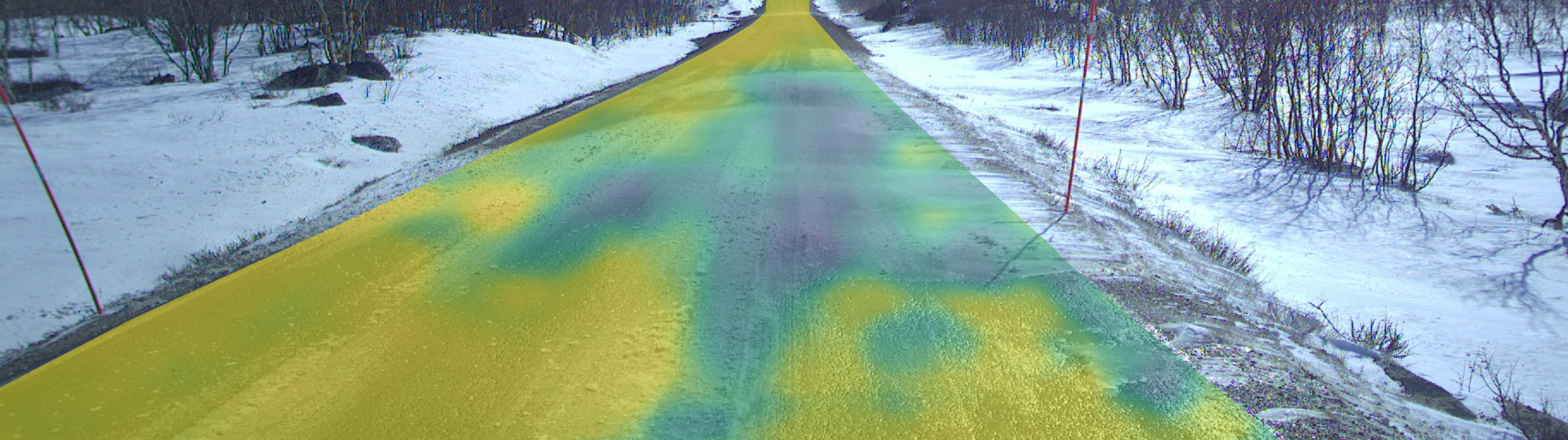}
} \\

\subfloat{
\includegraphics[width=0.19\linewidth]{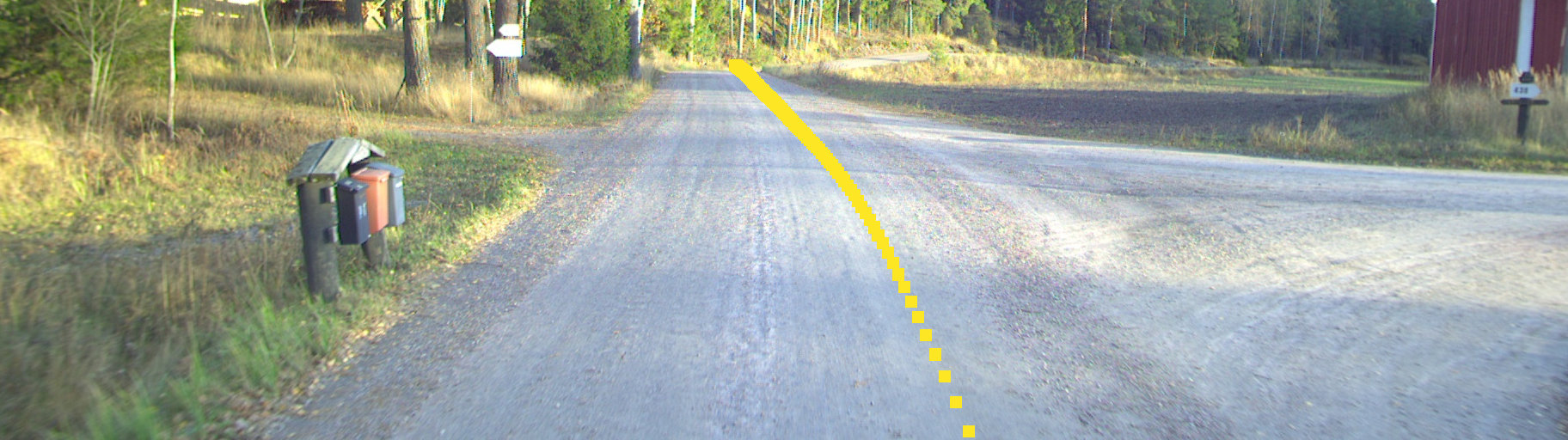}
}&
\subfloat{
\includegraphics[width=0.19\linewidth]{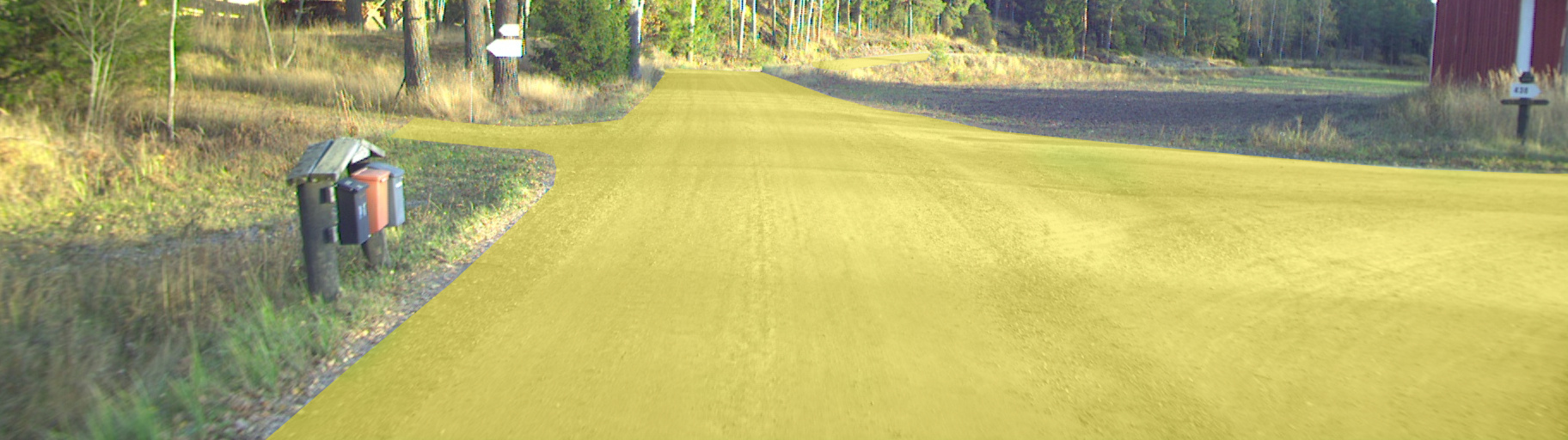}
}&
\subfloat{
\includegraphics[width=0.19\linewidth]{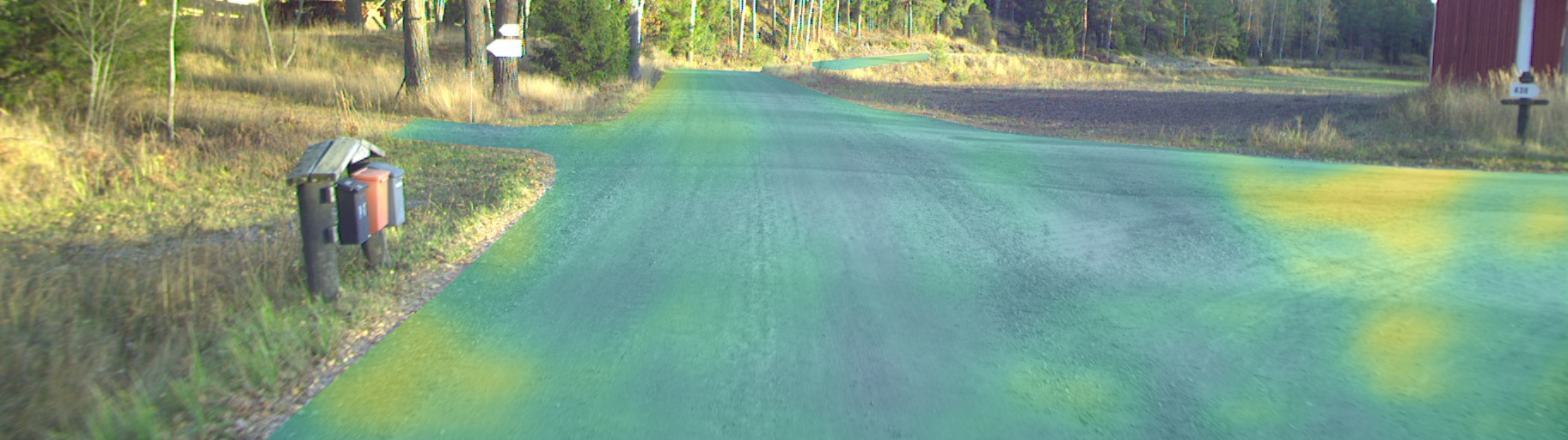}
}&
\subfloat{
\includegraphics[width=0.19\linewidth]{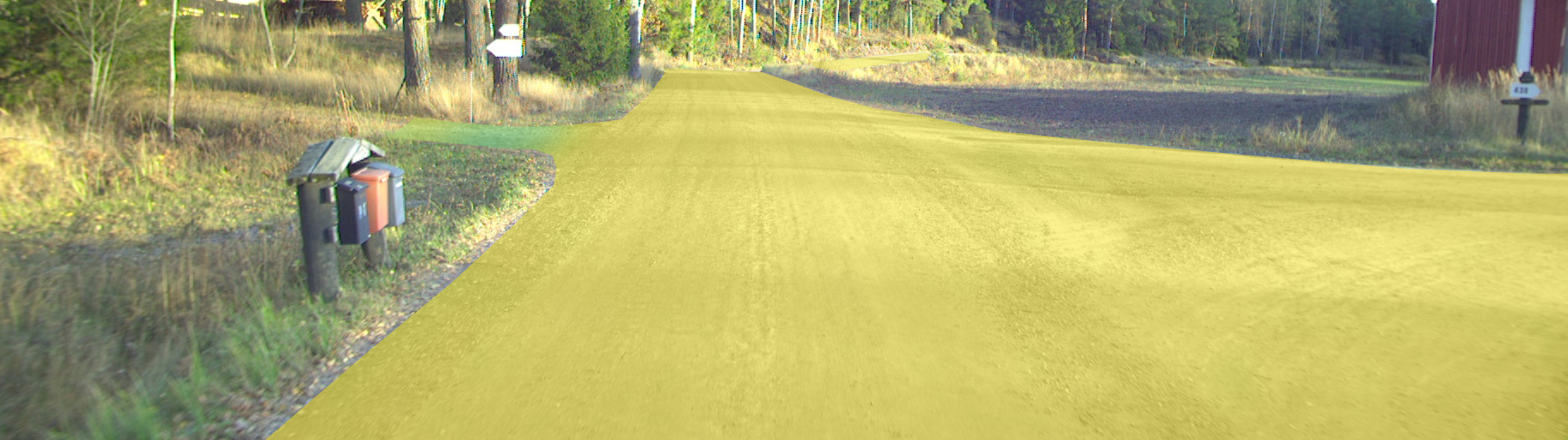}
}&
\subfloat{
\includegraphics[width=0.19\linewidth]{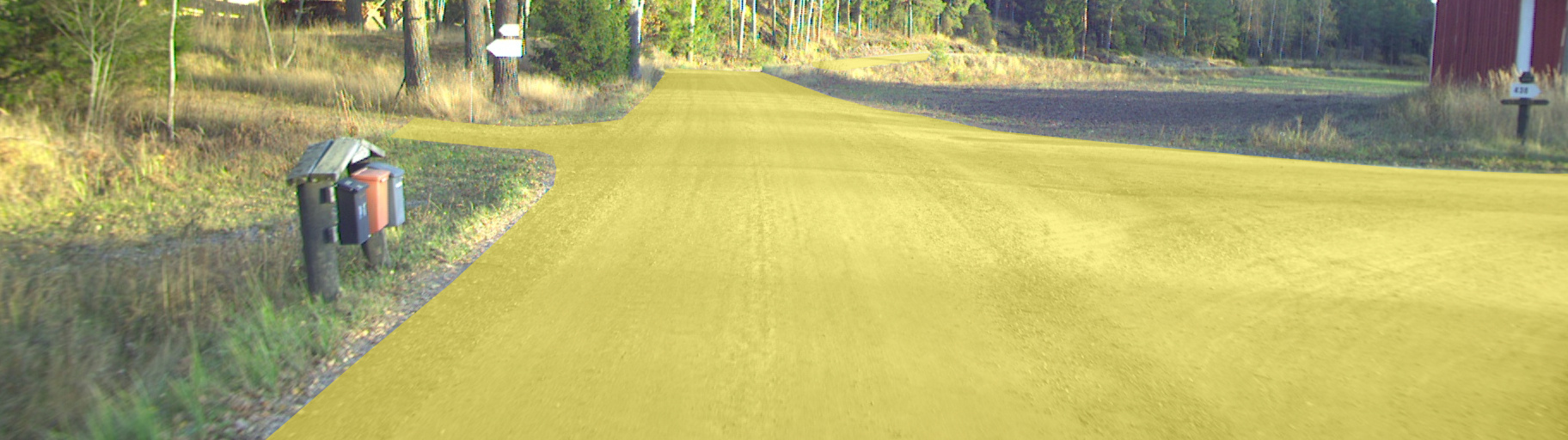}
} \\

\subfloat{
\includegraphics[width=0.19\linewidth]{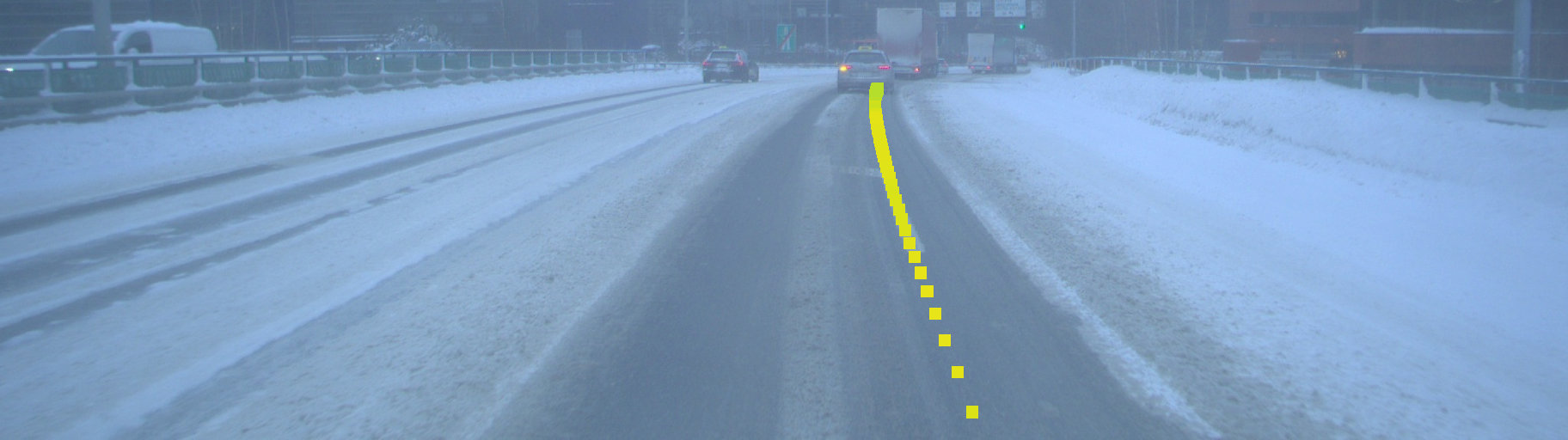}
}&
\subfloat{
\includegraphics[width=0.19\linewidth]{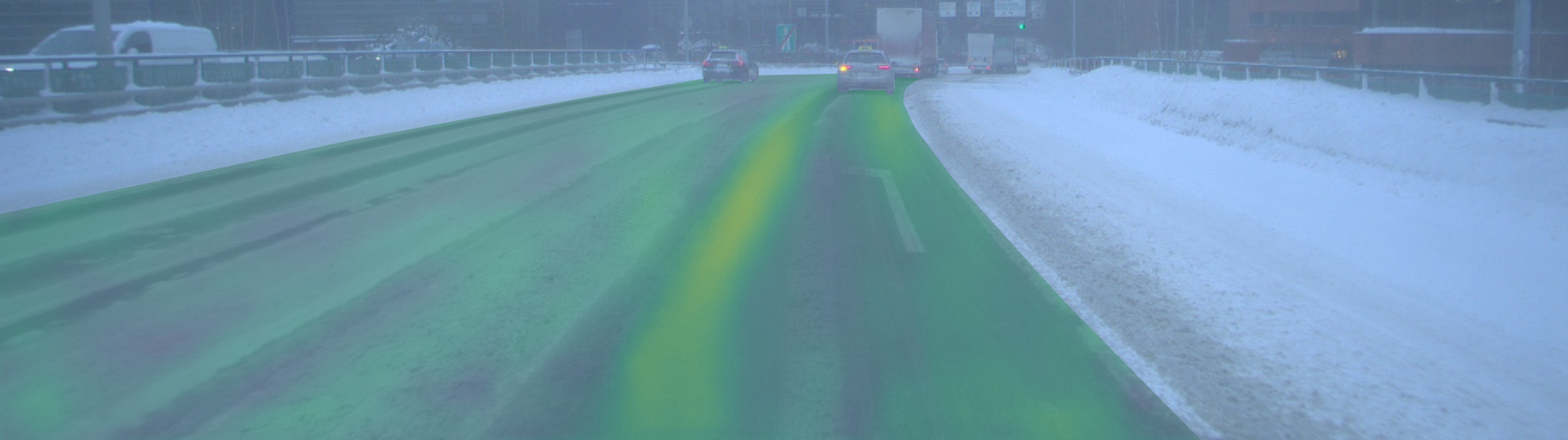}
}&
\subfloat{
\includegraphics[width=0.19\linewidth]{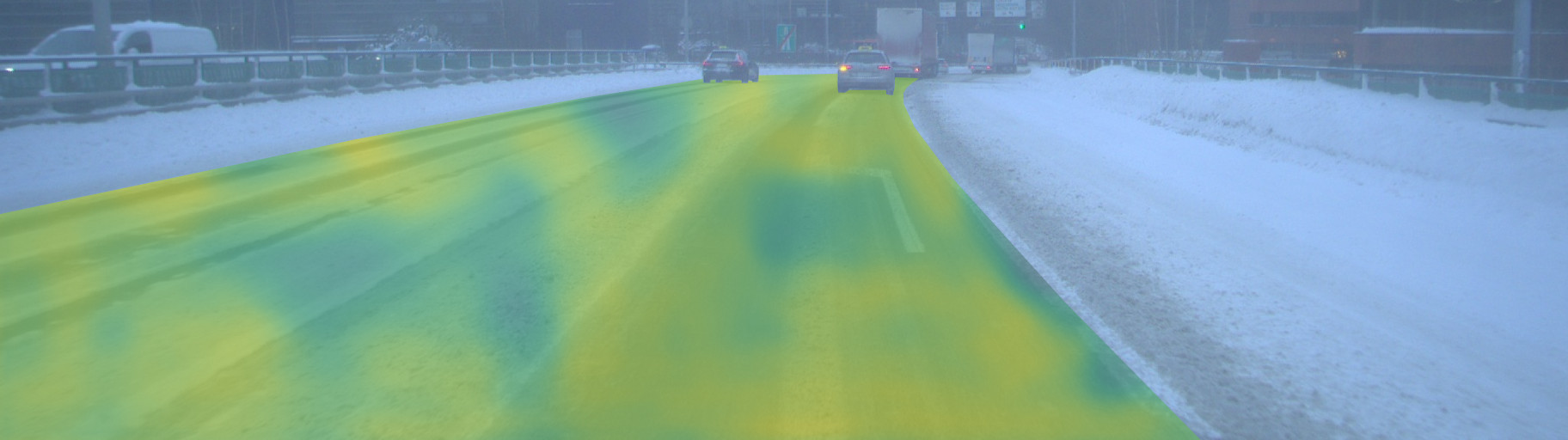}
}&
\subfloat{
\includegraphics[width=0.19\linewidth]{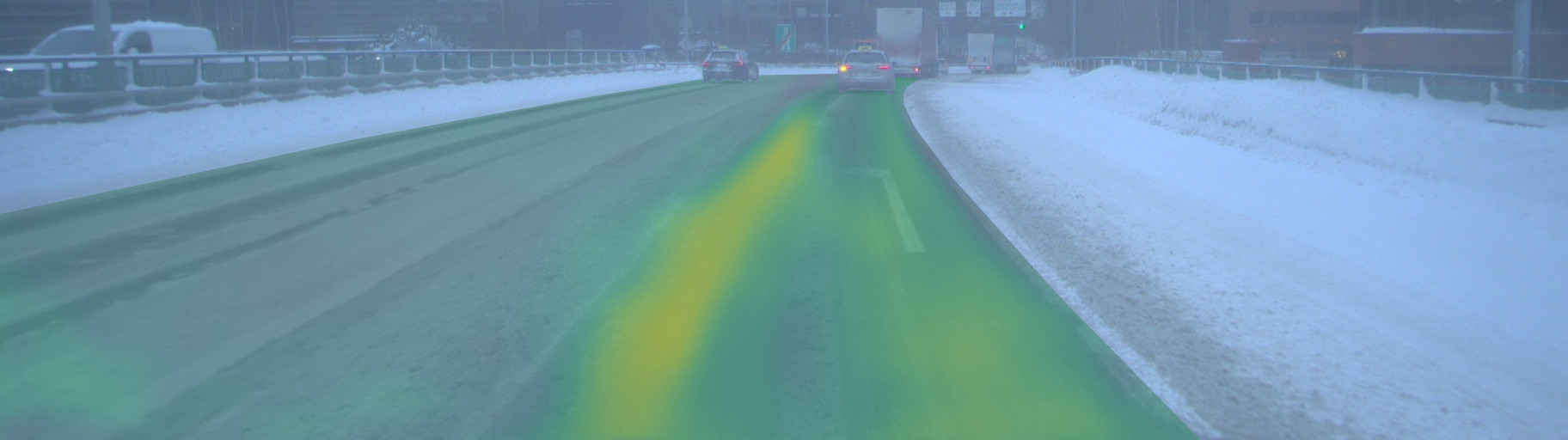}
}&
\subfloat{
\includegraphics[width=0.19\linewidth]{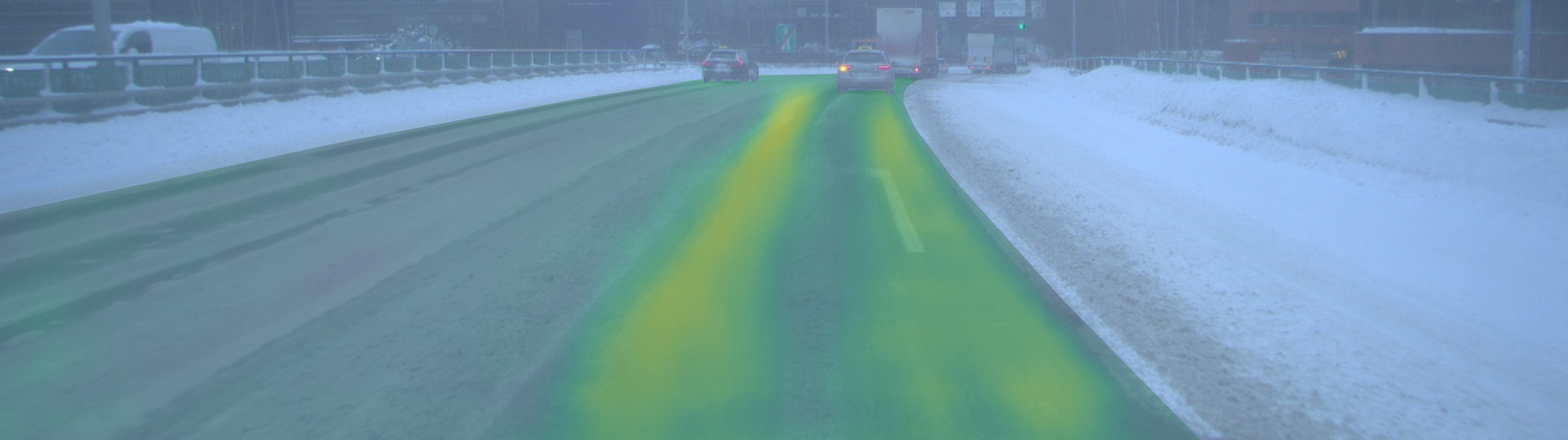}
} \\

\subfloat{
\includegraphics[width=0.19\linewidth]{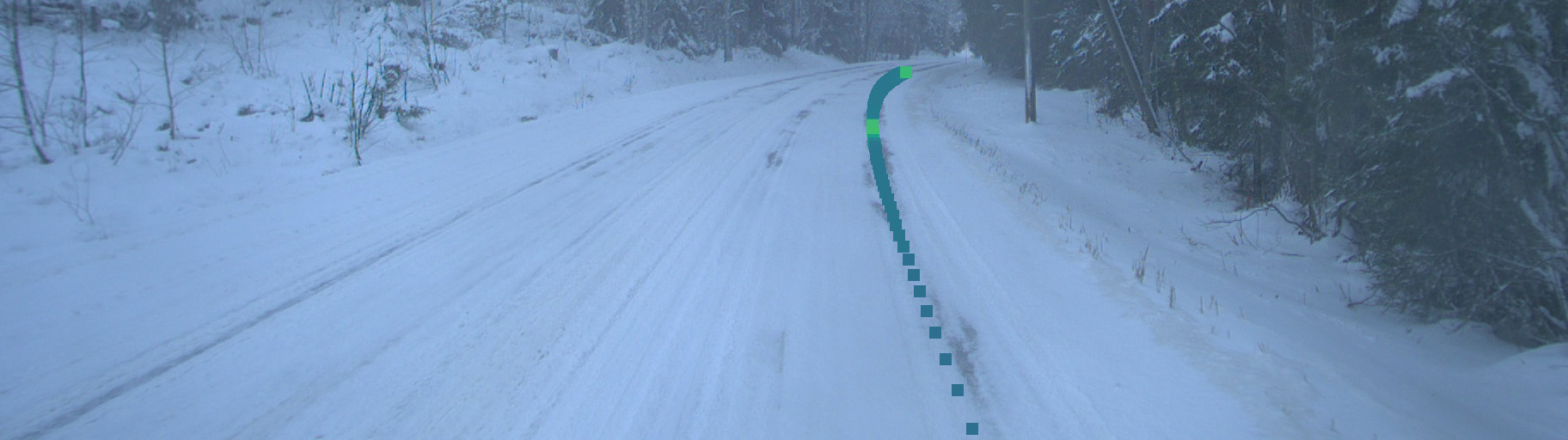}
}&
\subfloat{
\includegraphics[width=0.19\linewidth]{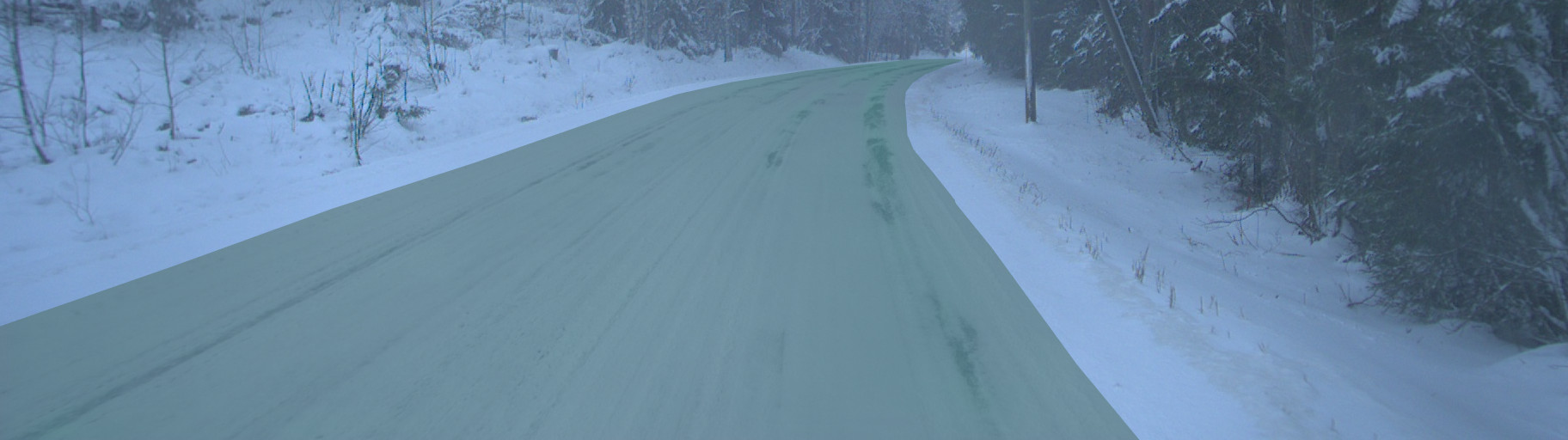}
}&
\subfloat{
\includegraphics[width=0.19\linewidth]{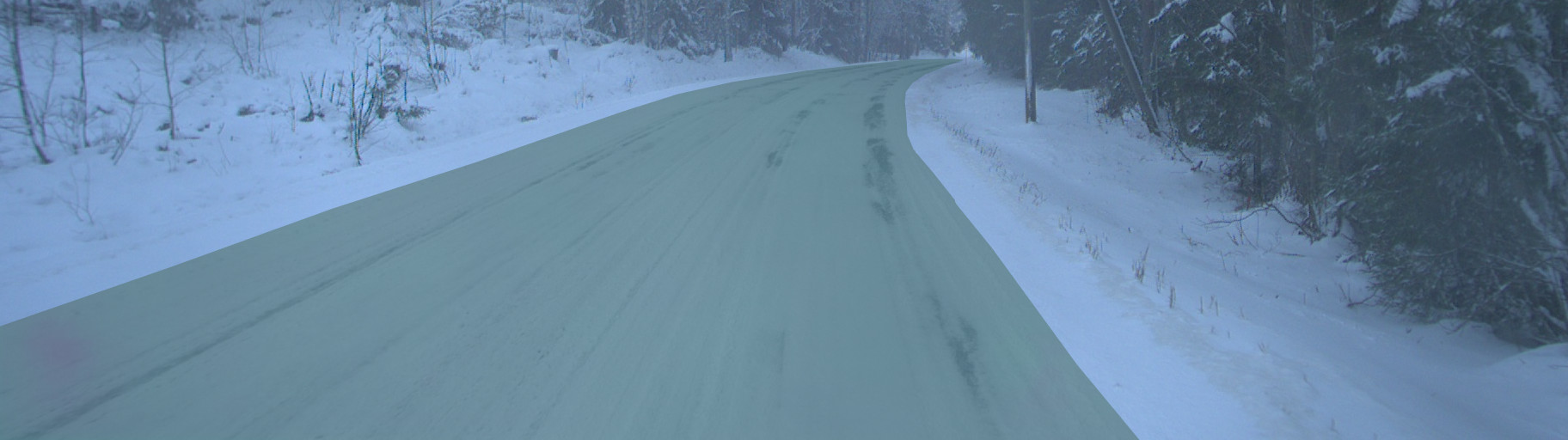}
}&
\subfloat{
\includegraphics[width=0.19\linewidth]{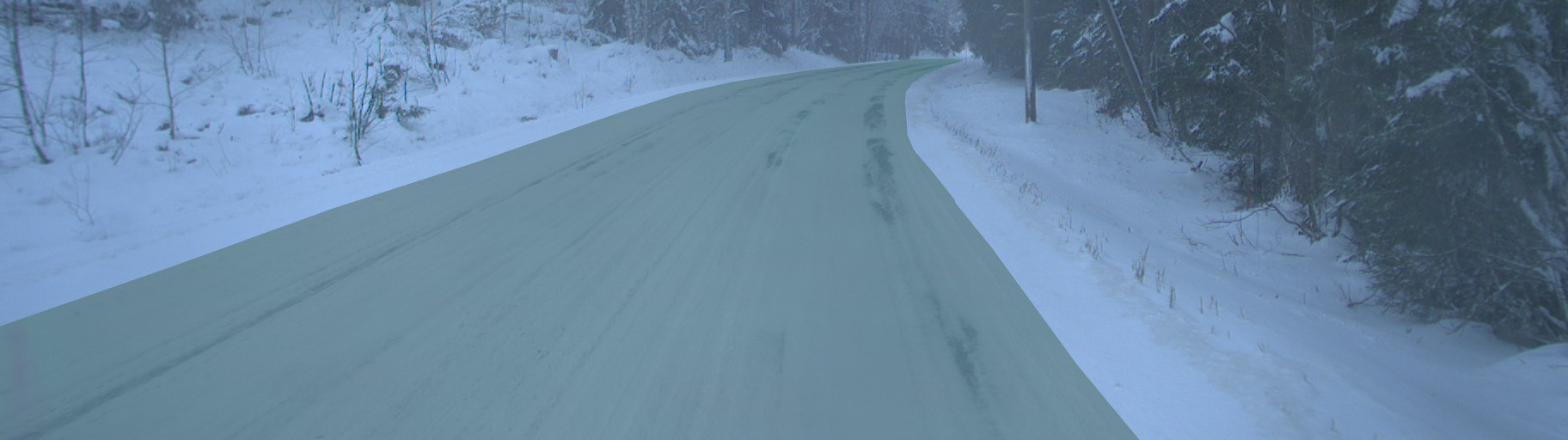}
}&
\subfloat{
\includegraphics[width=0.19\linewidth]{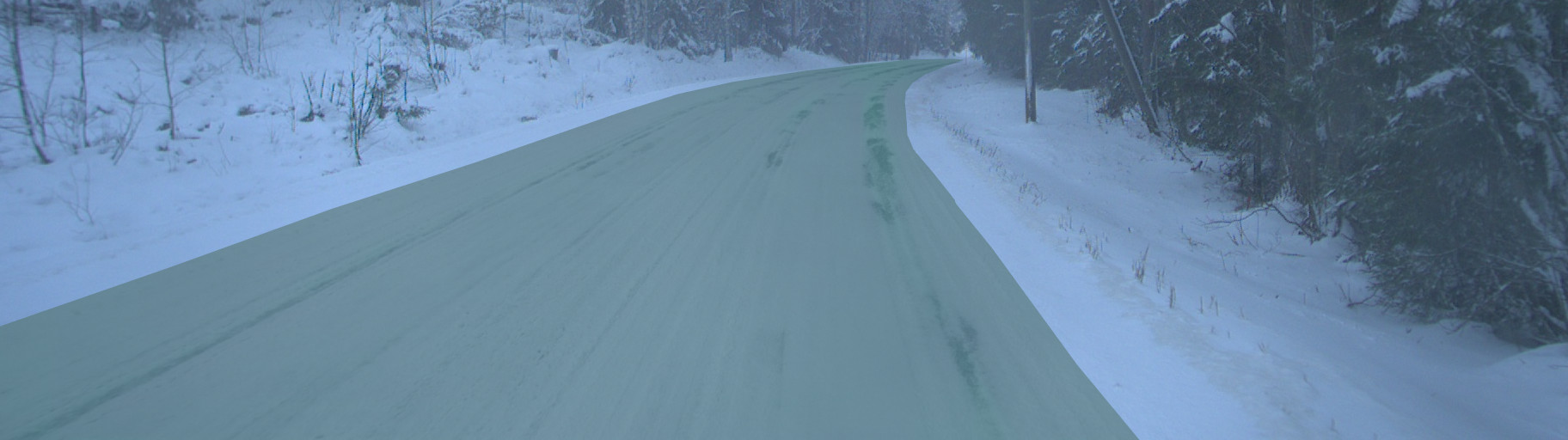}
} \\

\subfloat{
\includegraphics[width=0.19\linewidth]{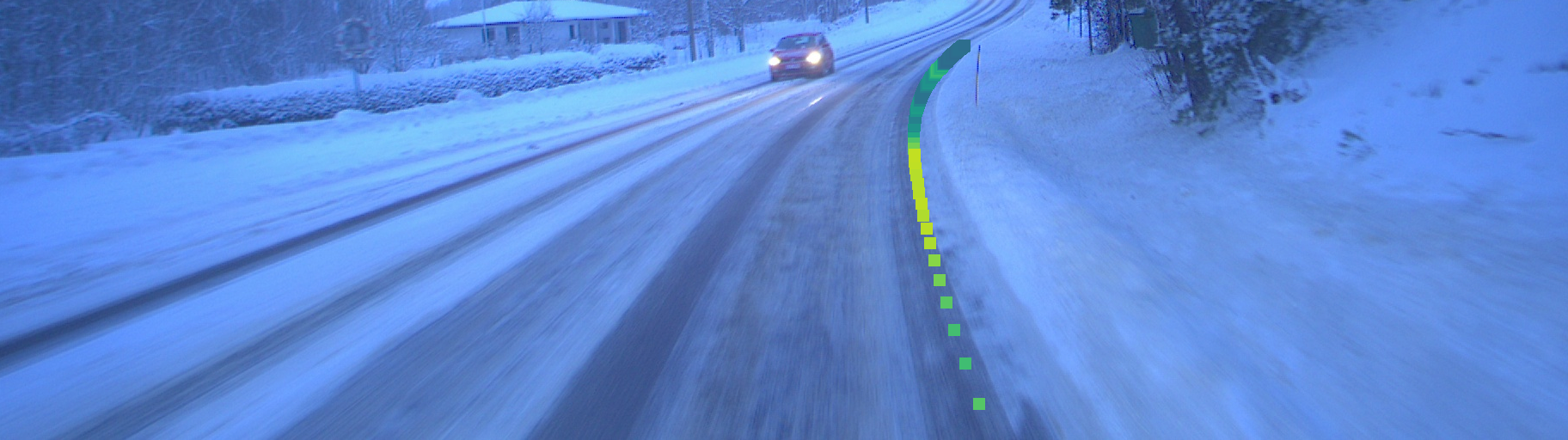}
}&
\subfloat{
\includegraphics[width=0.19\linewidth]{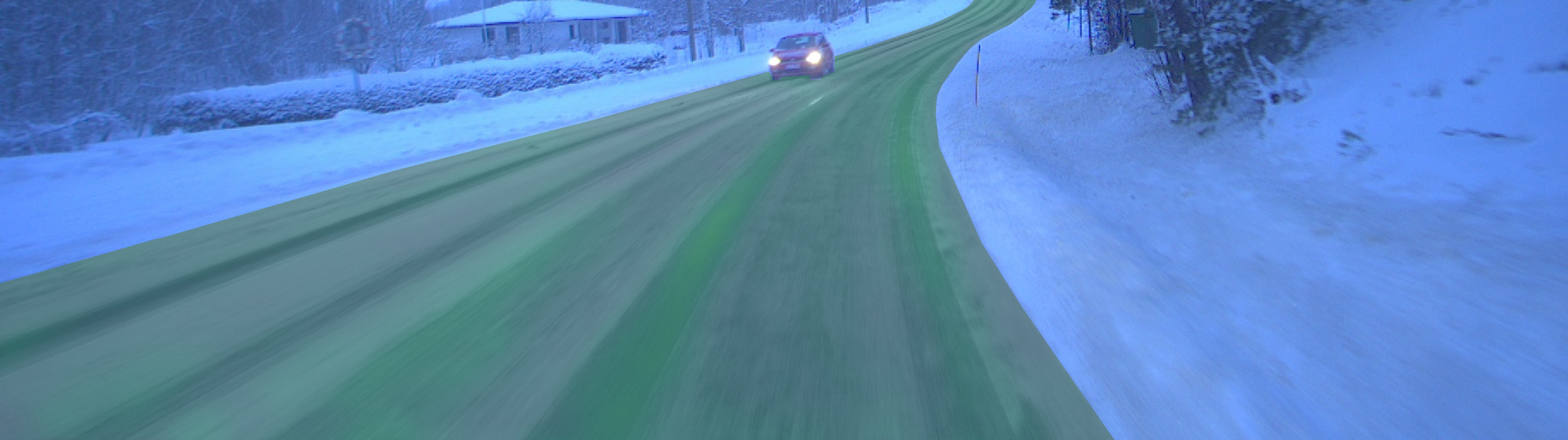}
}&
\subfloat{
\includegraphics[width=0.19\linewidth]{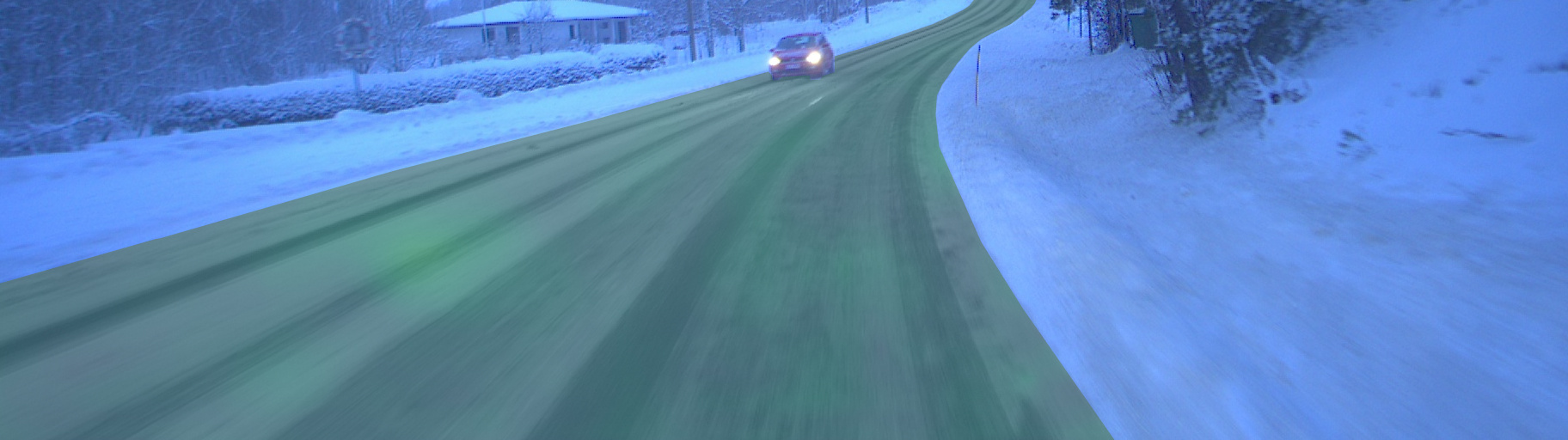}
}&
\subfloat{
\includegraphics[width=0.19\linewidth]{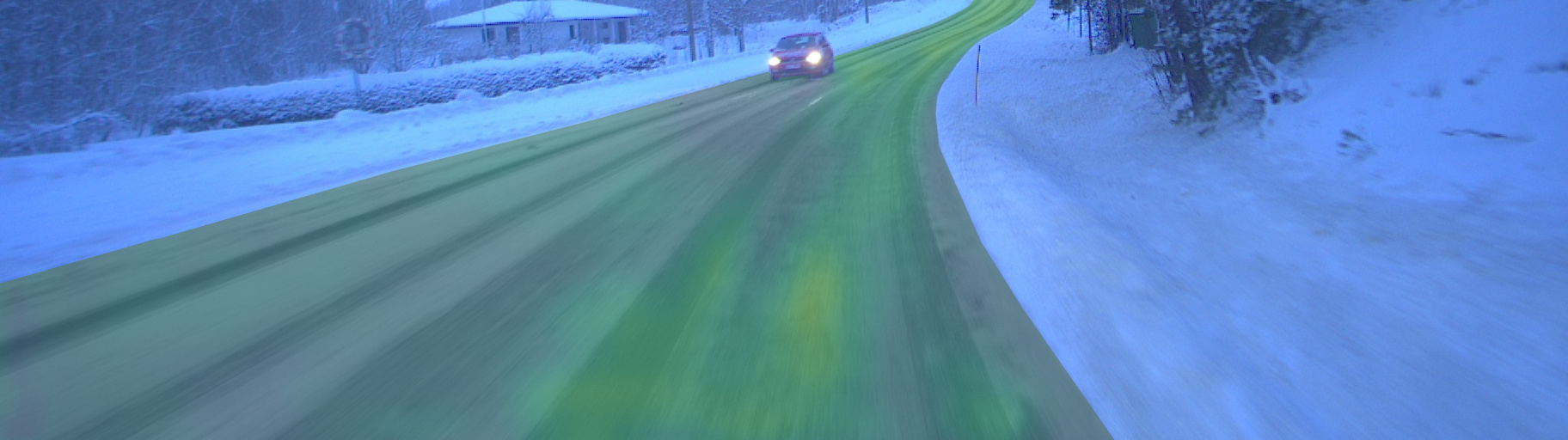}
}&
\subfloat{
\includegraphics[width=0.19\linewidth]{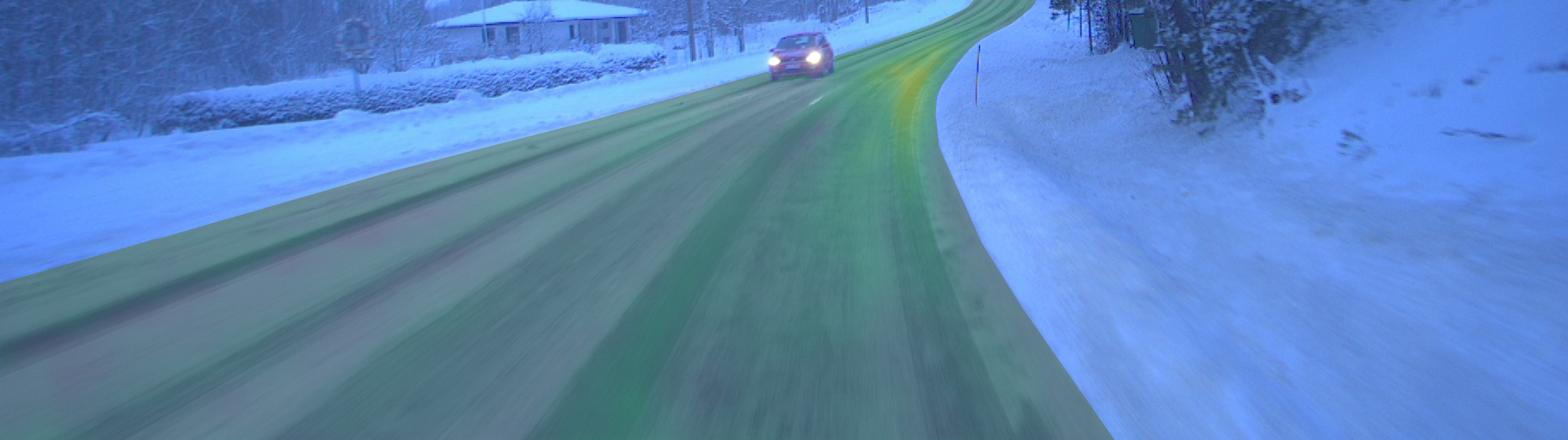}
} \\

\end{tabular}

\includegraphics[width=0.4\textwidth]{imgs/cbar.pdf}

\caption{Example grip prediction visualizations for each single-modality model and the RGB+T+R model. Some samples were chosen to highlight the performance differences between models.}
\label{fig:qualitative_mods}

\end{sidewaysfigure}

\section{Discussion}

The results support the original hypothesis on the accuracy of the dense grip map and the benefit of additional data modalities besides the RGB camera. Even though a large portion of the accuracy is obtained by separating the dry and snowy conditions from each other, it seems the model can perform in other road weather conditions as well.

However, one has to consider several error sources, as the optical road weather sensor is not designed to measure all the complex phenomena that could affect the grip between the tires and the road. In addition, the correct synchronization and alignment of the road weather sensor data was challenging. There is also a risk that the data split into training, validation, and test sets could cause some samples in different sets to have too many similarities meaning it's possible that some overfitting could not be observed from the validation and test set results. However, the results on the data from the three extra test drives defend the validation and test set results. In general, one would need an even larger representation of different weather conditions in the dataset to obtain a model with less bias and higher accuracy in several real-life road weather conditions. Despite these limitations, our results show evidence of the performance of our method.

\section{Conclusions}

This study presents a novel method to predict a dense grip map of the road area from multimodal image data with a convolutional neural network. The models using RGB or 3D LiDAR reflectance measurements provide the best baseline results, whereas the highest accuracy predictions are achieved with sensor fusion using modality-wise encoders. The use of thermal camera images also shows potential, while their contribution is smaller than that of the RGB and reflectance measurements. The results follow those of earlier studies in proving that the RGB camera is a powerful tool for detecting road surface conditions while also providing major steps in using 3D LiDAR reflectance measurements for dense grip prediction both alone and alongside RGB cameras.

The best model configuration using a combination of all three input modalities achieves an RMSE of 0.0632 and an RMSE of 0.0575 on the diverse validation and test sets respectively. The results from separate test drives also prove the system's usability in unseen conditions. In addition, the qualitative results show the model recognizing various shapes of snow, ice, and water layer distributions affecting the grip prediction. These qualitative results were also improved with the model that uses multimodal inputs with the fusion of encoder features.

To achieve a reliable implementation of this method for autonomous driving, one should collect a large dataset with improved sensor data quality and an even more diverse and balanced set of road and weather conditions. It could also be investigated if one could improve the prediction accuracy by switching the reference image plane from the presented RGB camera frame to another plane, such as the bird's-eye view of the road area or even the 3D frame of the LiDAR. Finally, one should develop methods to predict the uncertainty of the grip prediction output to fuse the output from this method reliably with autonomous driving systems.

\subsubsection*{Acknowledgements}

Funded by the European Union (grant no. 101069576). Views and opinions expressed are however those of the author(s) only and do not necessarily reflect those of the European Union or the European Climate, Infrastructure and Environment Executive Agency (CINEA). Neither the European Union nor the granting authority can be held responsible for them.

The contribution of the authors is as follows: Maanp{\"a}{\"a} and Pesonen planned the study, performed experiments, and wrote the manuscript. Maanp{\"a}{\"a} also preprocessed the dataset and Pesonen performed preliminary experiments before this study. Manninen, Maanp{\"a}{\"a} and Hyyti developed the research vehicle for data collection with the research group and Maanp{\"a}{\"a} collected the dataset. Hyyti, Melekhov, and Kannala advised in the planning of the study. Kukko and Hyypp{\"a} supervised the project.

In addition, we would like to thank Eugeniu Vezeteu for his help in data collection and sensor calibration and Paula Litkey for participating in the vehicle development. We would also like to thank Eero Ahokas for GNSS trajectory processing and Josef Taher for their advice during this work.

%
%
%
\bibliographystyle{splncs04}
\bibliography{sample}

\input{supplementary_material}

\end{document}

%% file: supplementary_material.tex
\begin{landscape}
\thispagestyle{empty}
\begin{smashminipage}
\section*{Supplementary Material}
    \centering
    \begin{tabular}{cccc}
Ground truth & RGB+T & RGB+R & R+T \\

\includegraphics[width=0.24\linewidth]{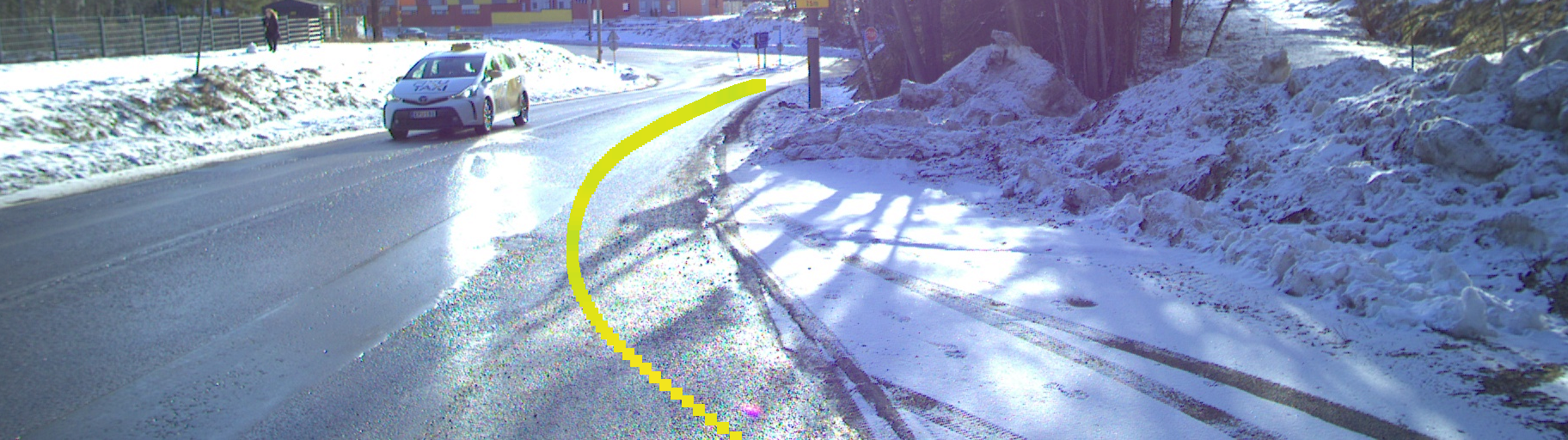}
&
\includegraphics[width=0.24\linewidth]{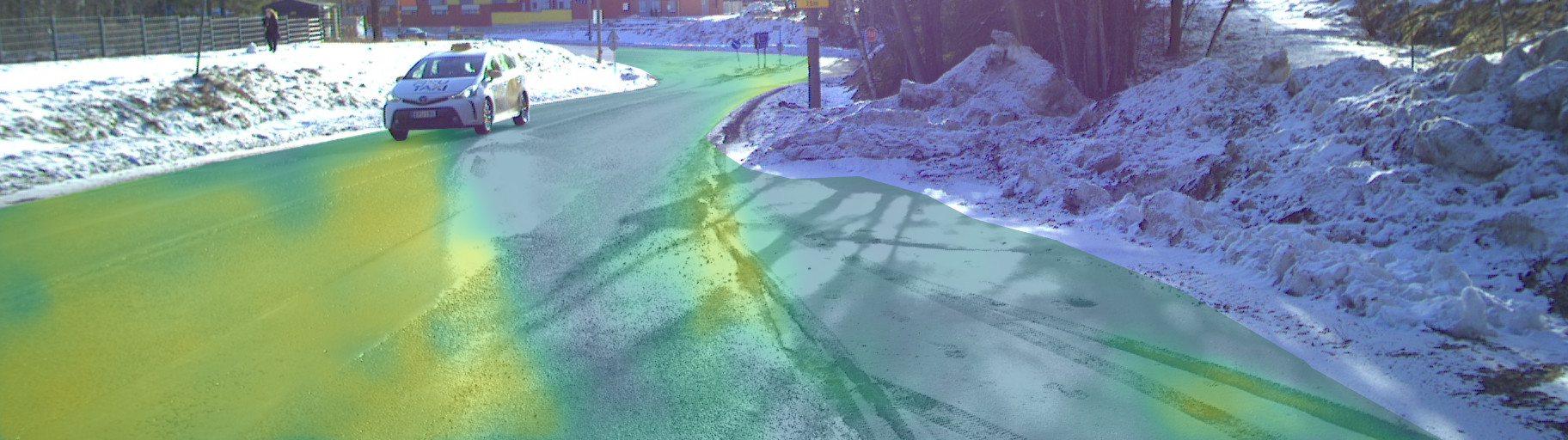}
&
\includegraphics[width=0.24\linewidth]{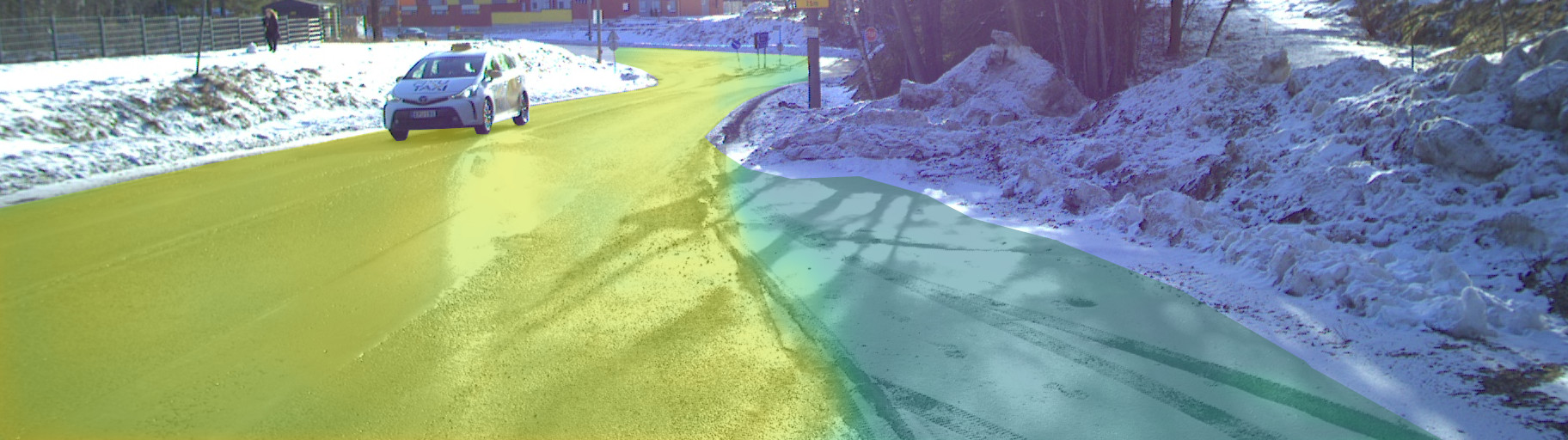}
&
\includegraphics[width=0.24\linewidth]{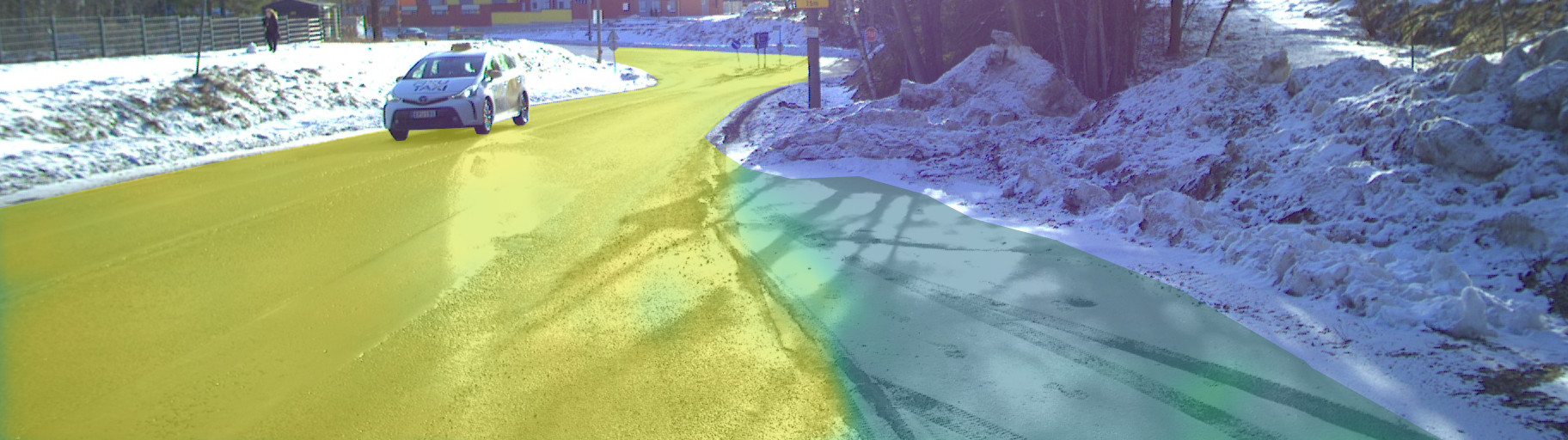}
\\

\includegraphics[width=0.24\linewidth]{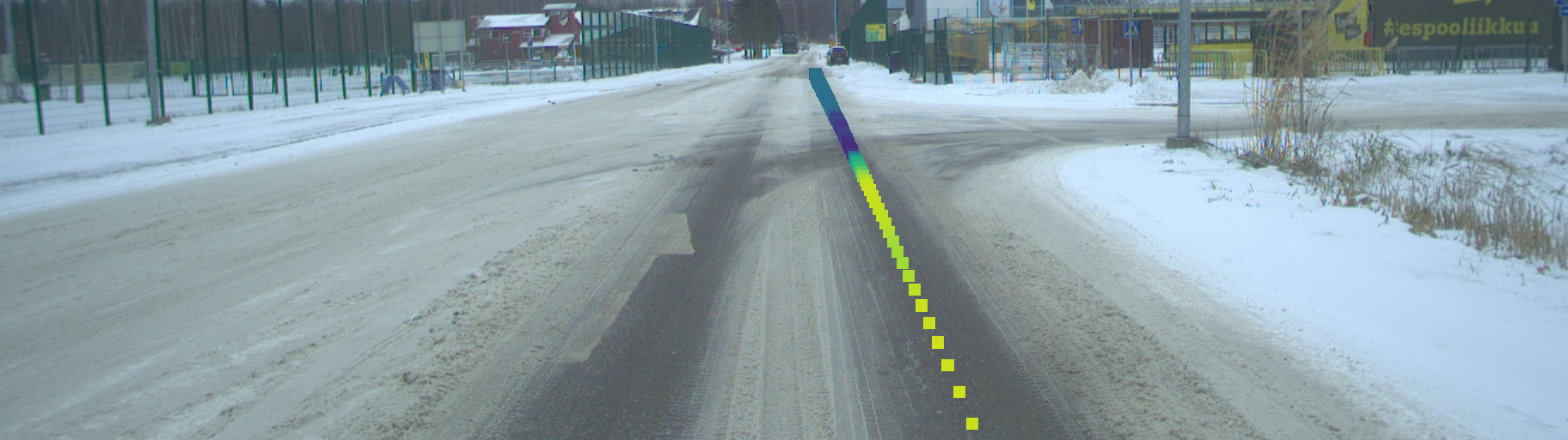}
&
\includegraphics[width=0.24\linewidth]{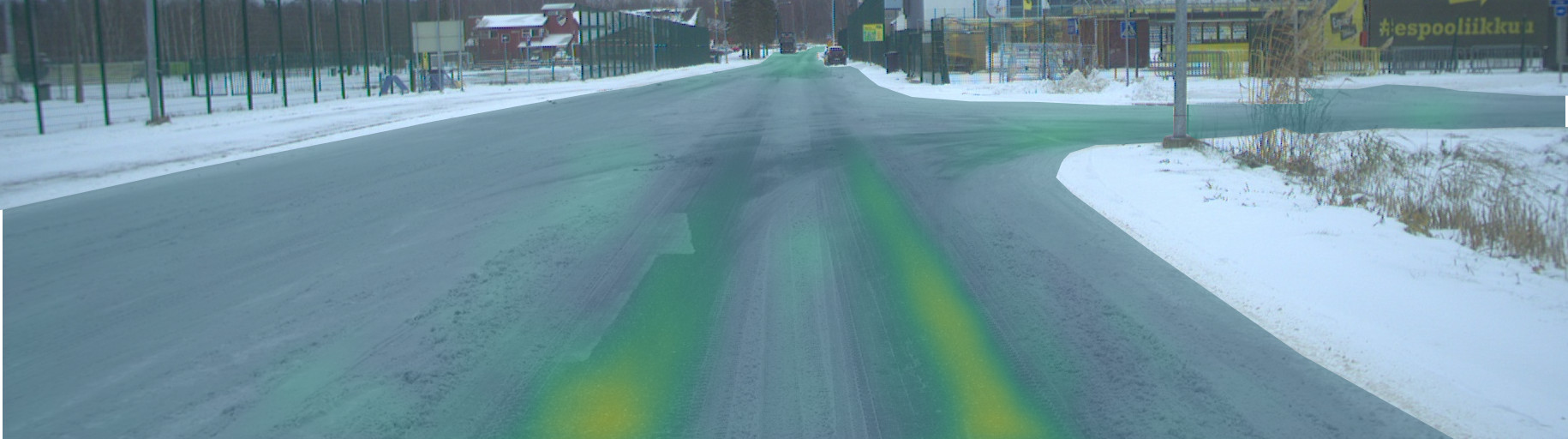}
&
\includegraphics[width=0.24\linewidth]{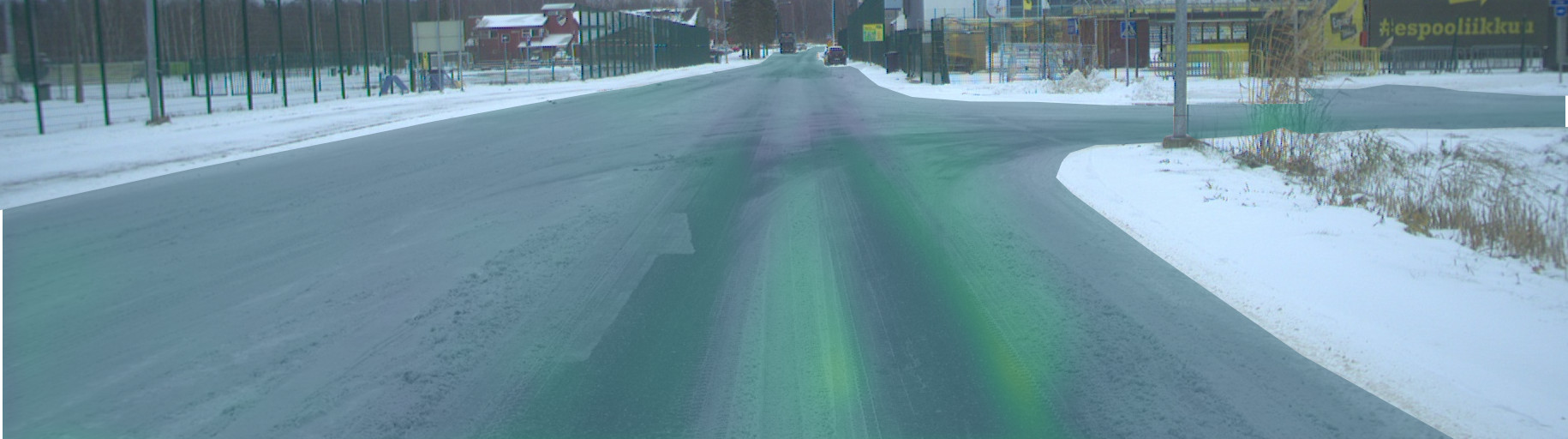}
&
\includegraphics[width=0.24\linewidth]{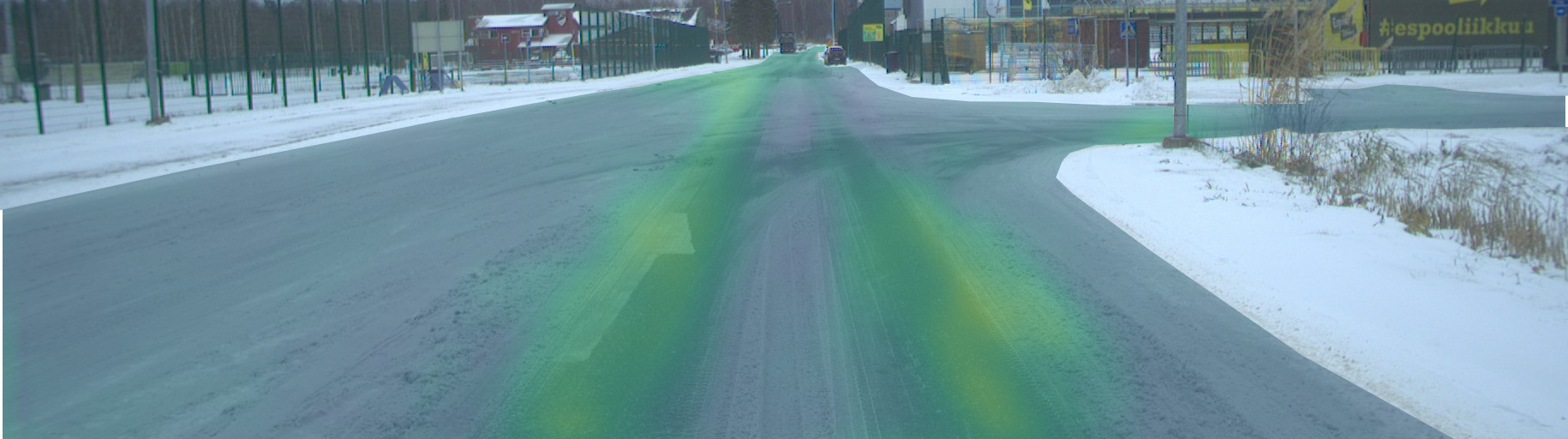}
\\

\includegraphics[width=0.24\linewidth]{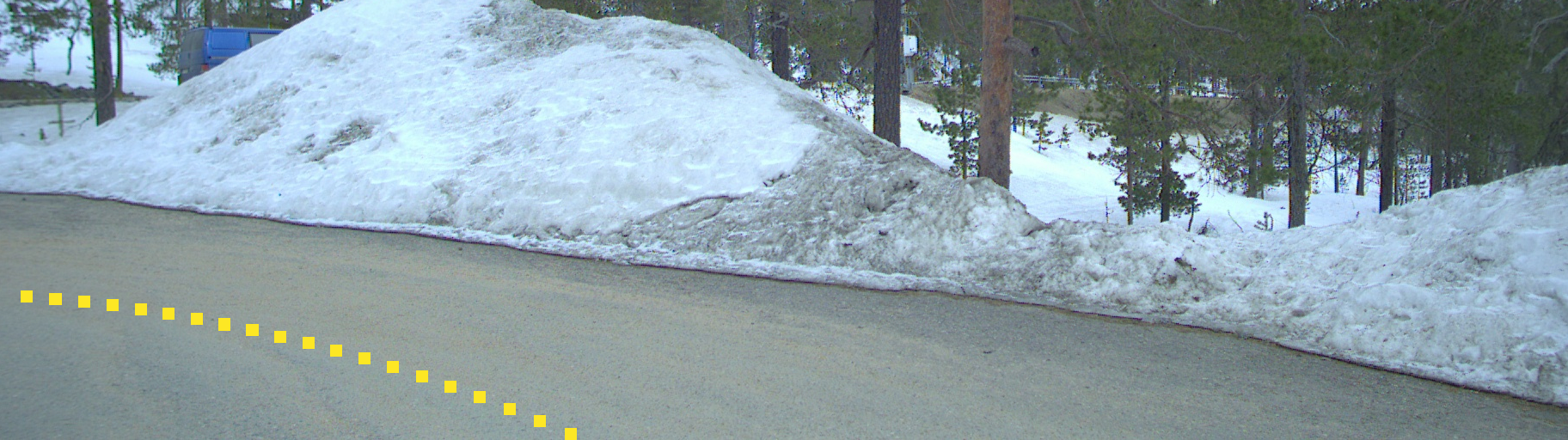}
&
\includegraphics[width=0.24\linewidth]{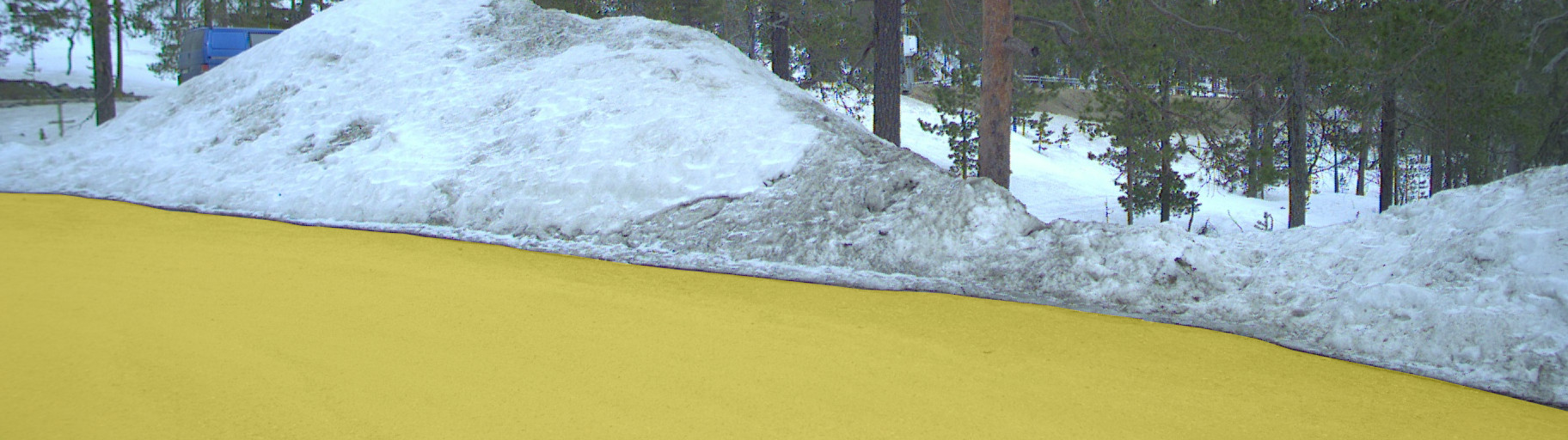}
&
\includegraphics[width=0.24\linewidth]{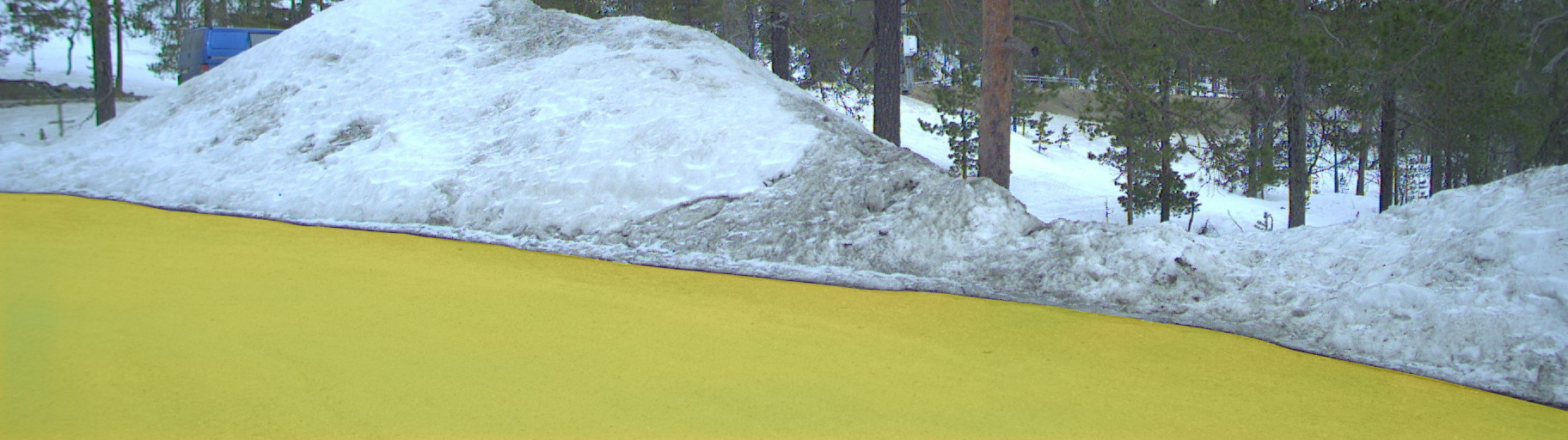}
&
\includegraphics[width=0.24\linewidth]{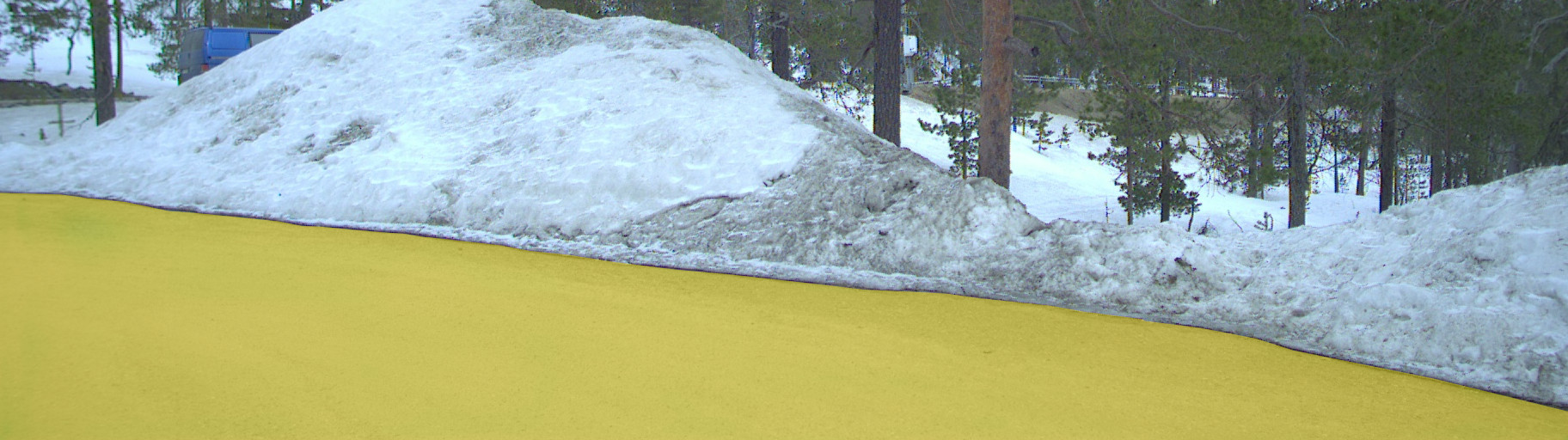}
\\

\includegraphics[width=0.24\linewidth]{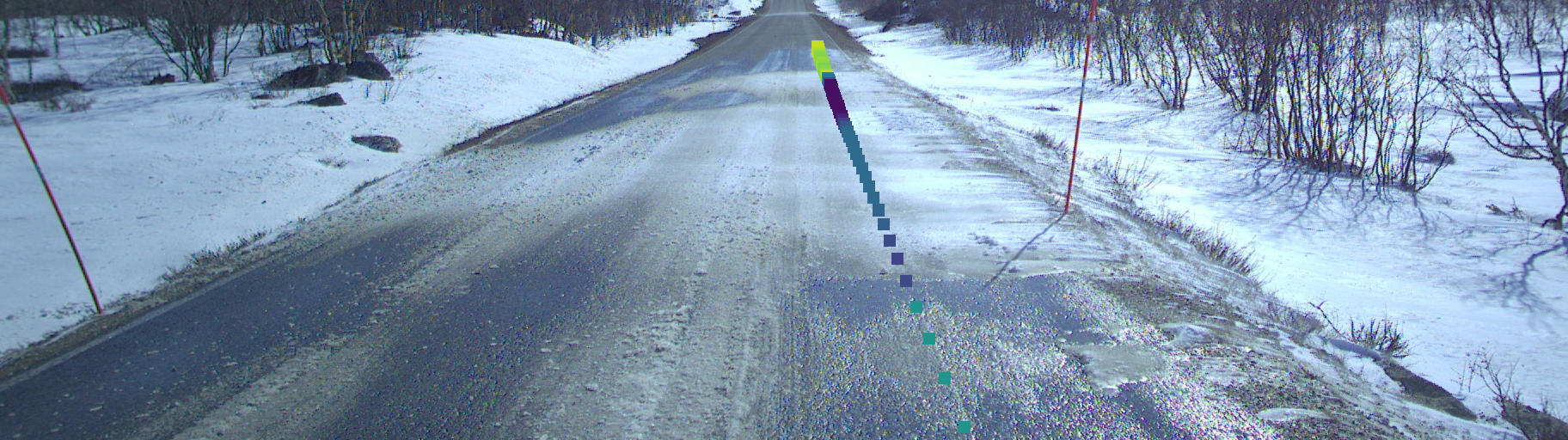}
&
\includegraphics[width=0.24\linewidth]{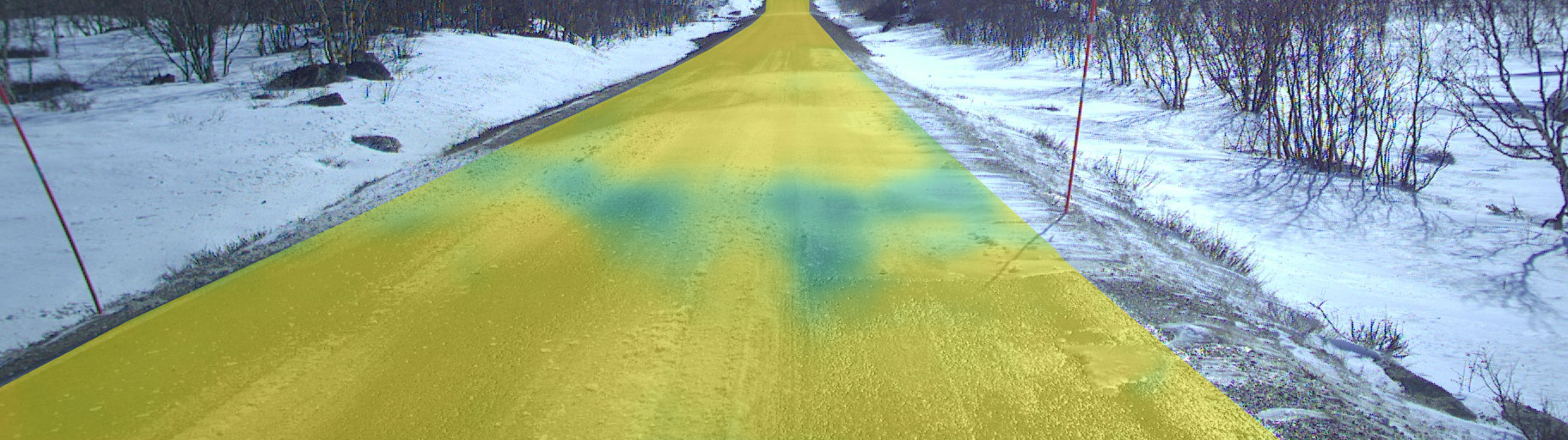}
&
\includegraphics[width=0.24\linewidth]{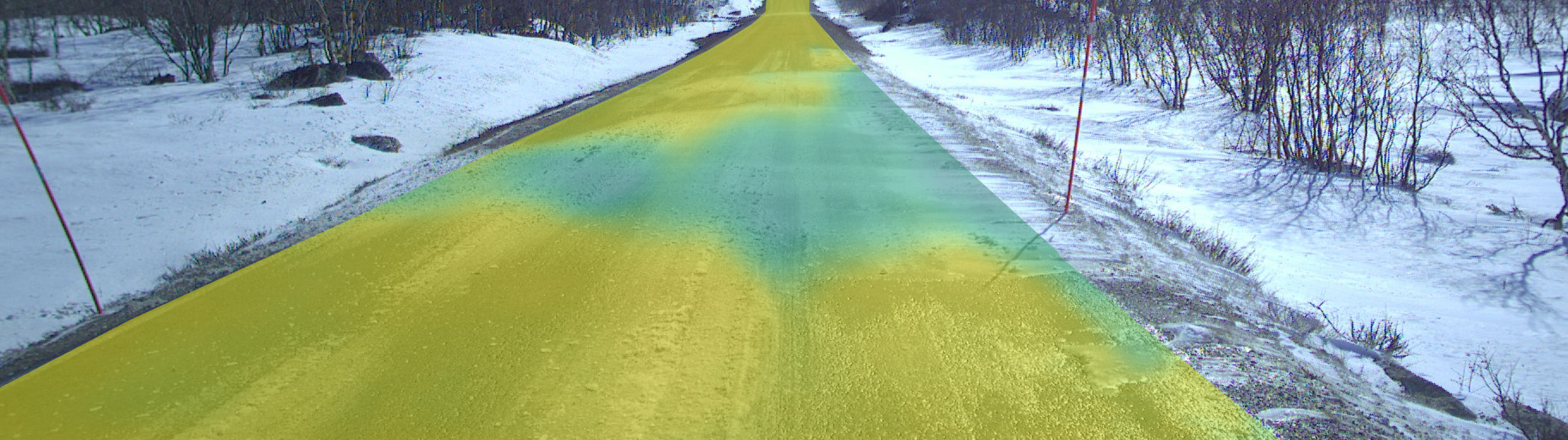}
&
\includegraphics[width=0.24\linewidth]{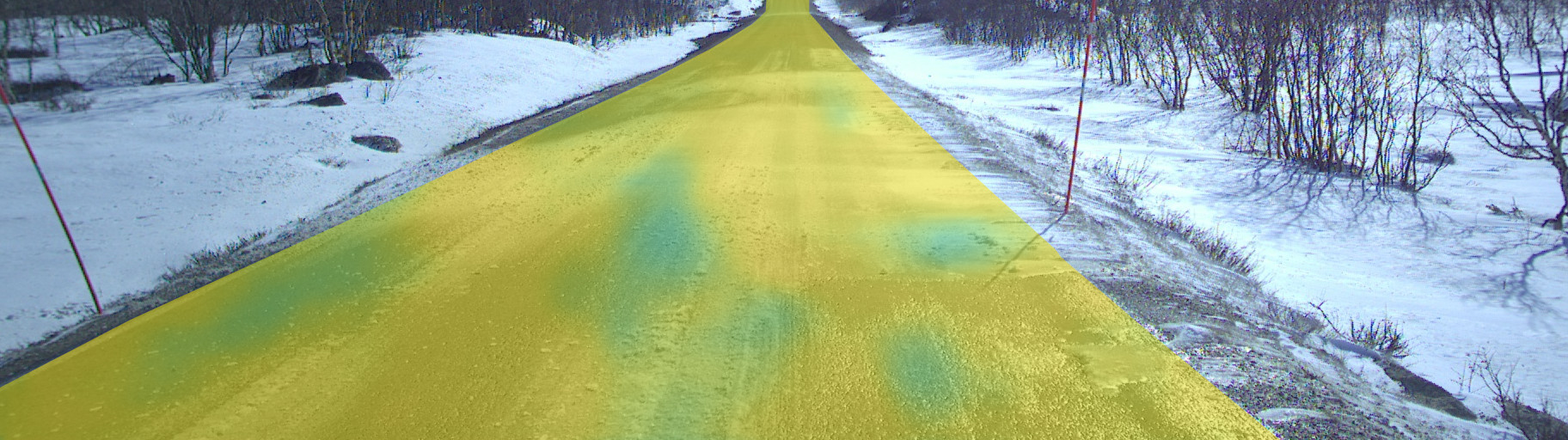}
\\

\includegraphics[width=0.24\linewidth]{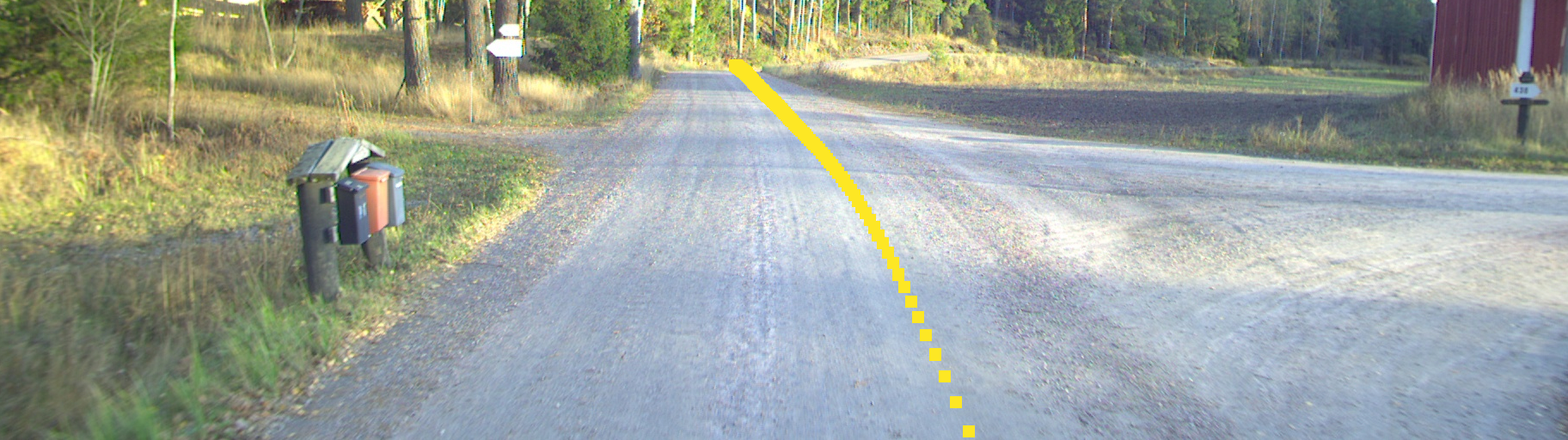}
&
\includegraphics[width=0.24\linewidth]{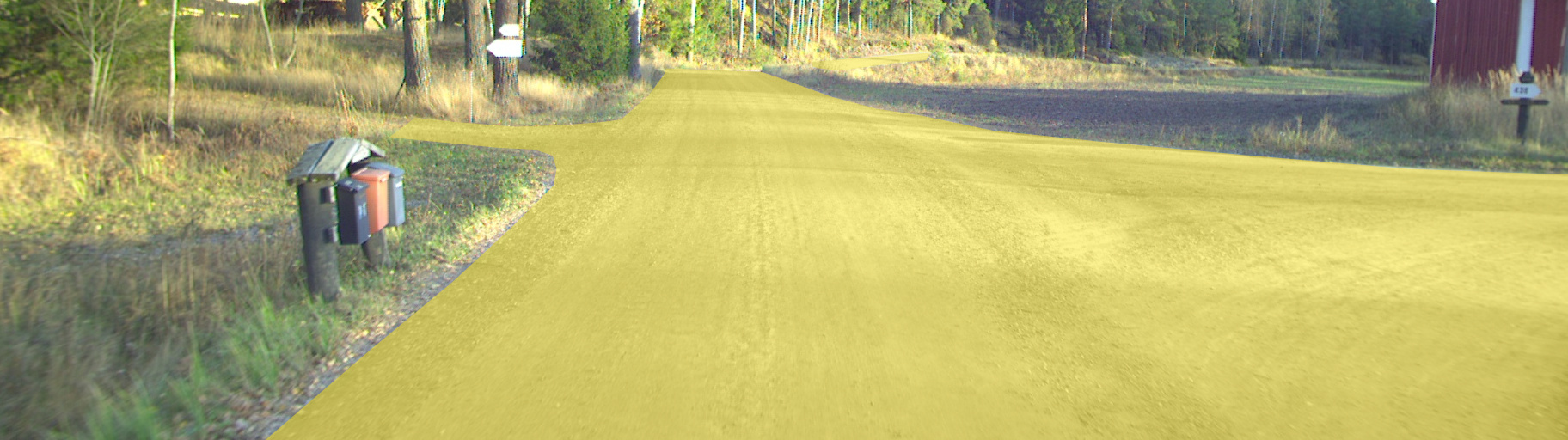}
&
\includegraphics[width=0.24\linewidth]{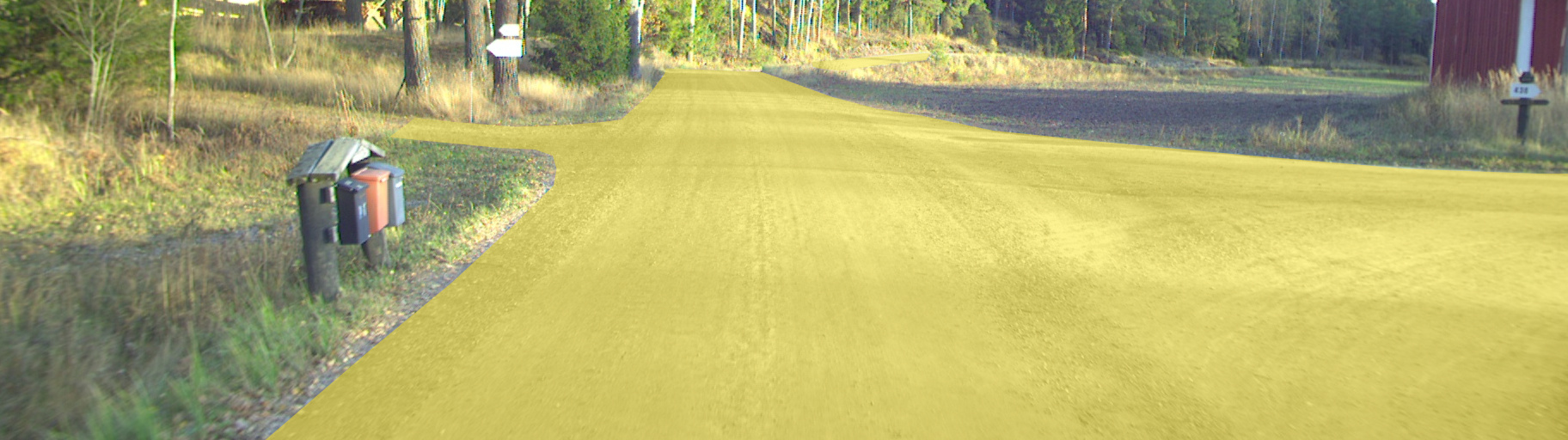}
&
\includegraphics[width=0.24\linewidth]{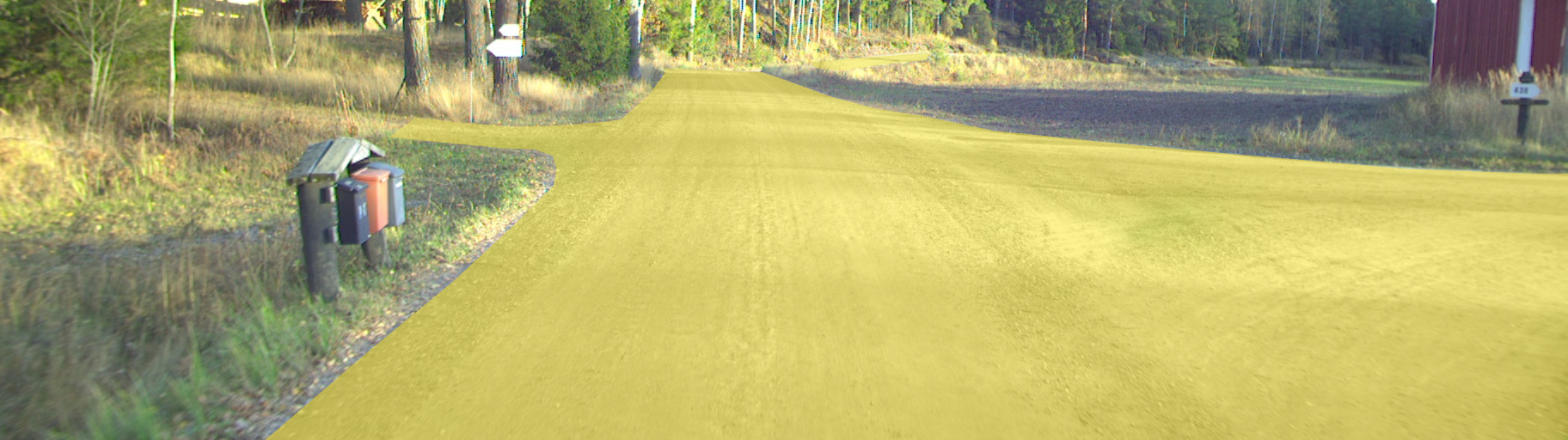}
\\

\includegraphics[width=0.24\linewidth]{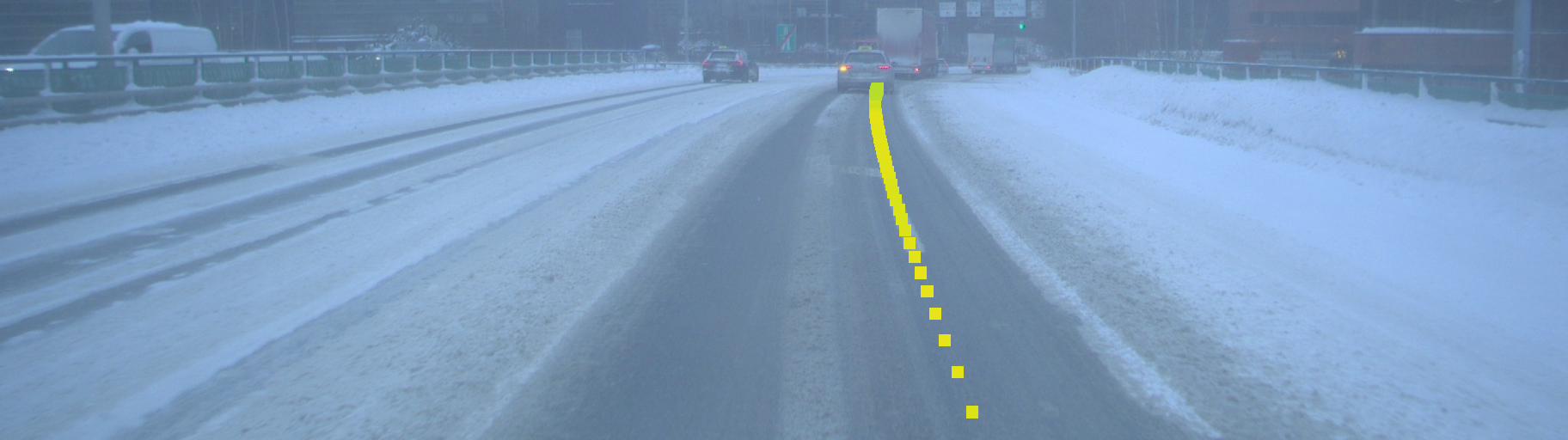}
&
\includegraphics[width=0.24\linewidth]{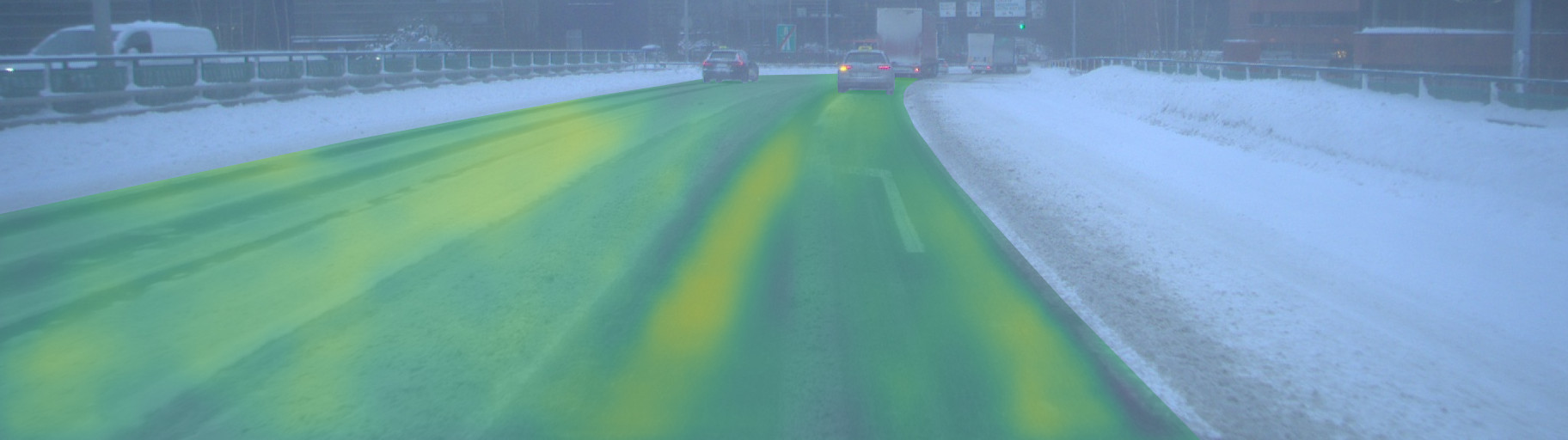}
&
\includegraphics[width=0.24\linewidth]{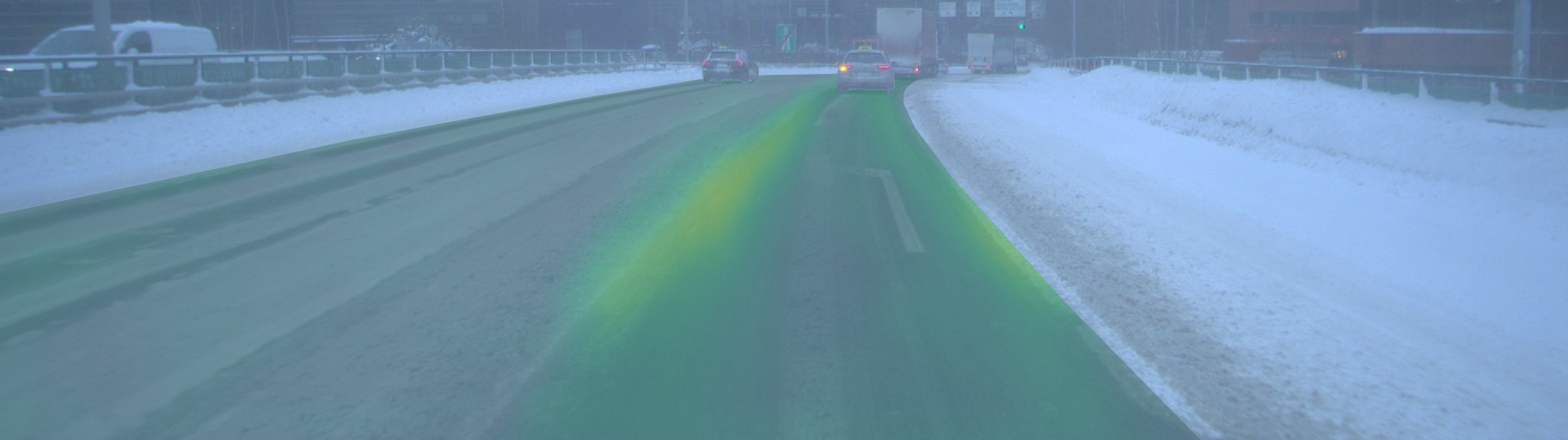}
&
\includegraphics[width=0.24\linewidth]{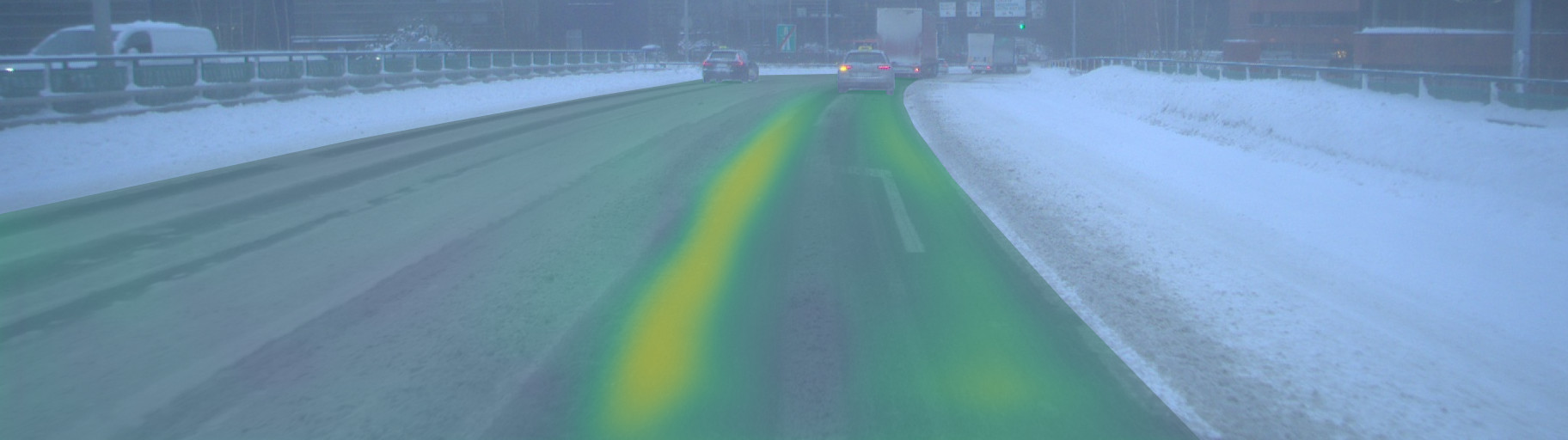}
\\

\includegraphics[width=0.24\linewidth]{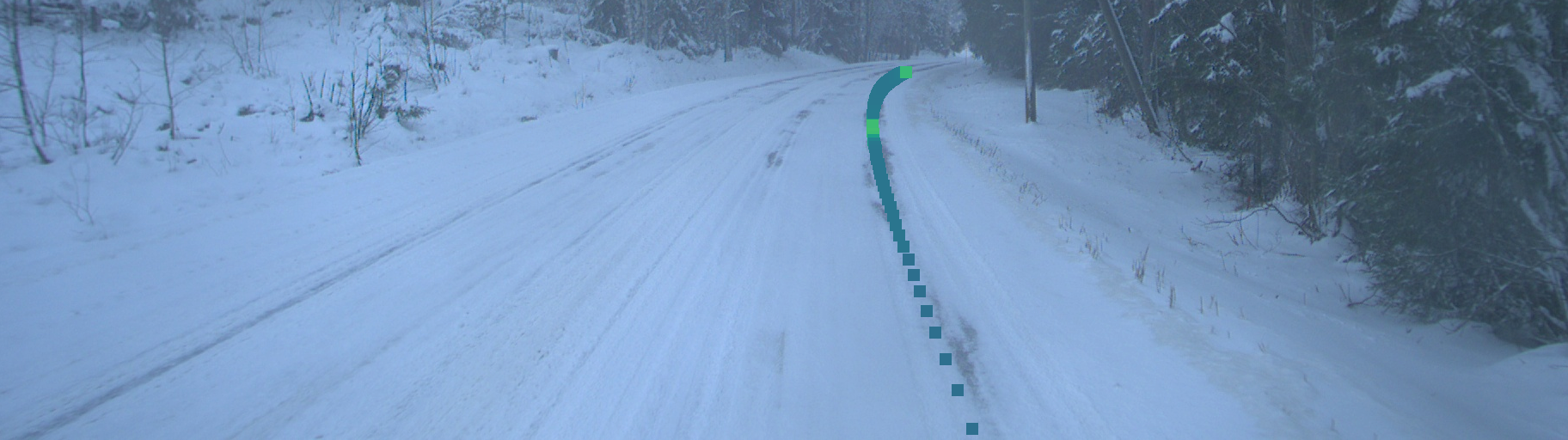}
&
\includegraphics[width=0.24\linewidth]{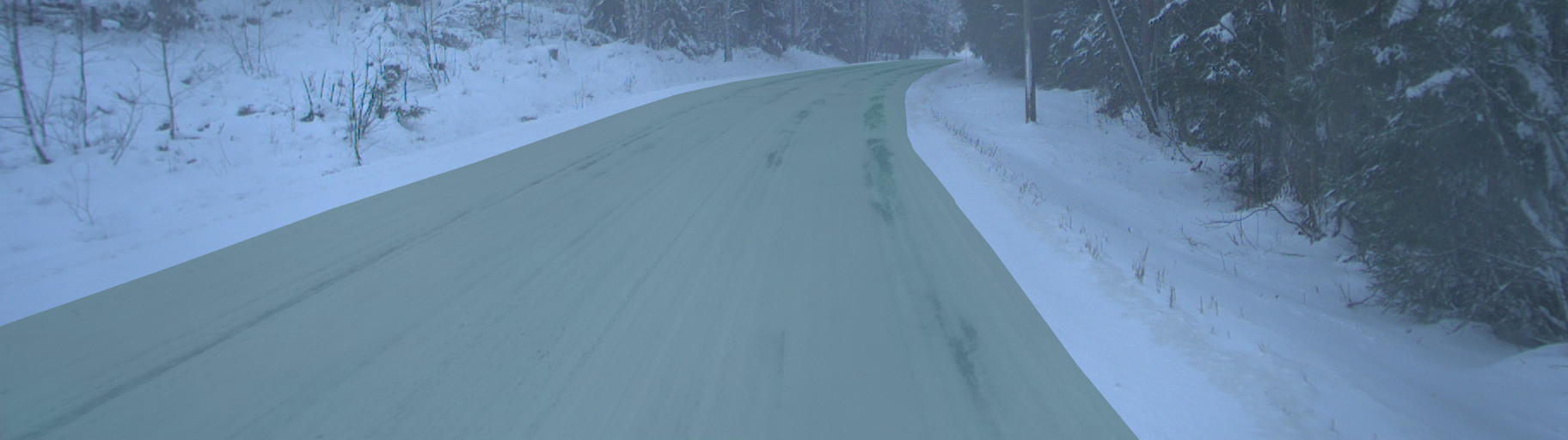}
&
\includegraphics[width=0.24\linewidth]{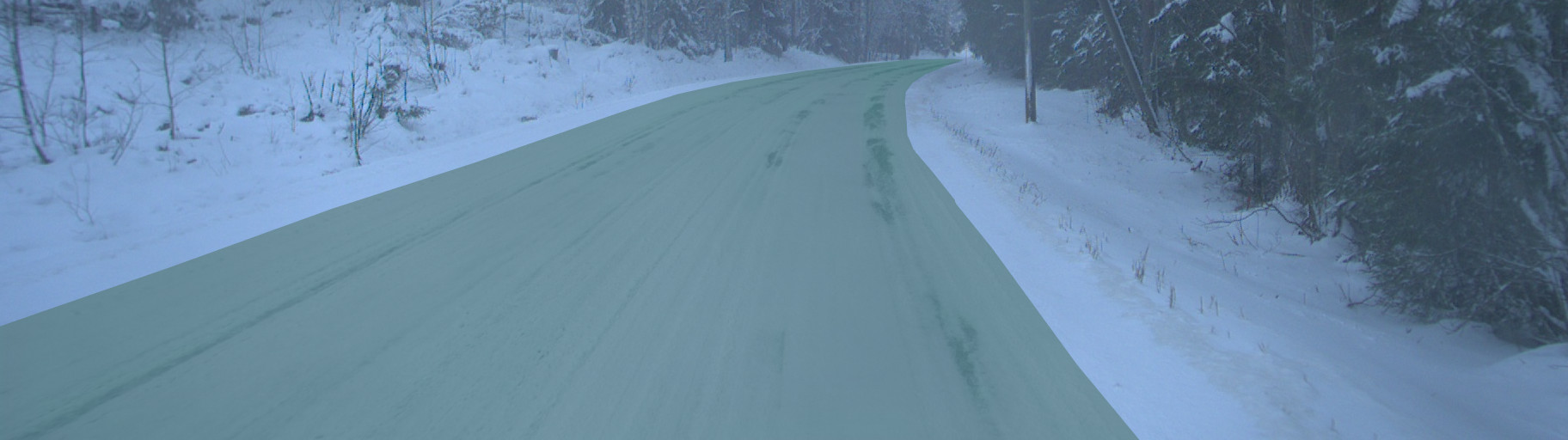}
&
\includegraphics[width=0.24\linewidth]{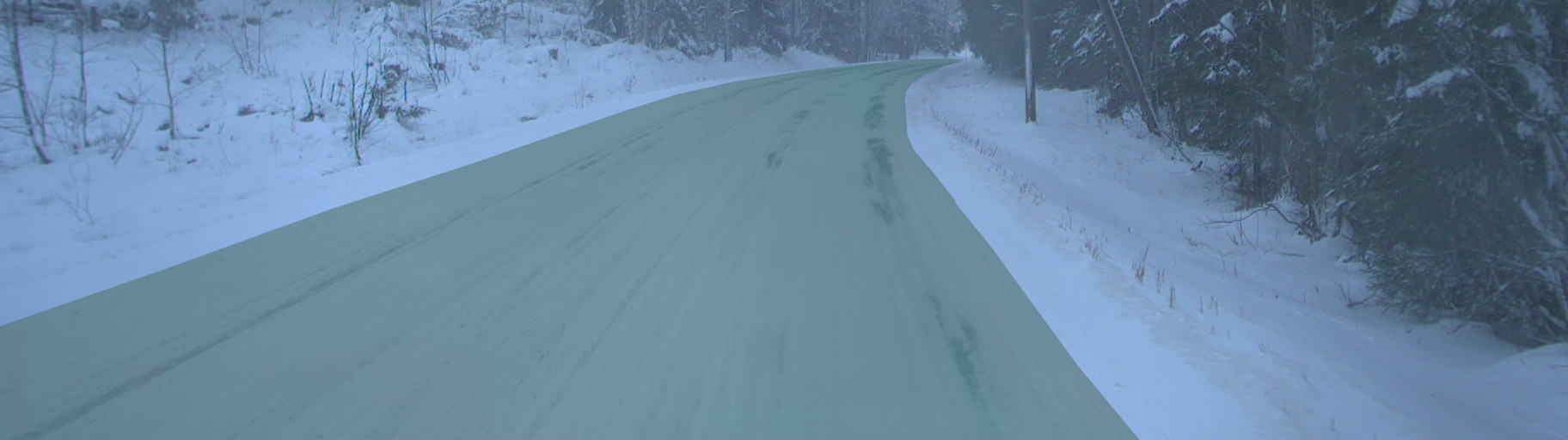}
\\

\includegraphics[width=0.24\linewidth]{imgs/worst_cases/gt/1677165393790740955.png}
&
\includegraphics[width=0.24\linewidth]{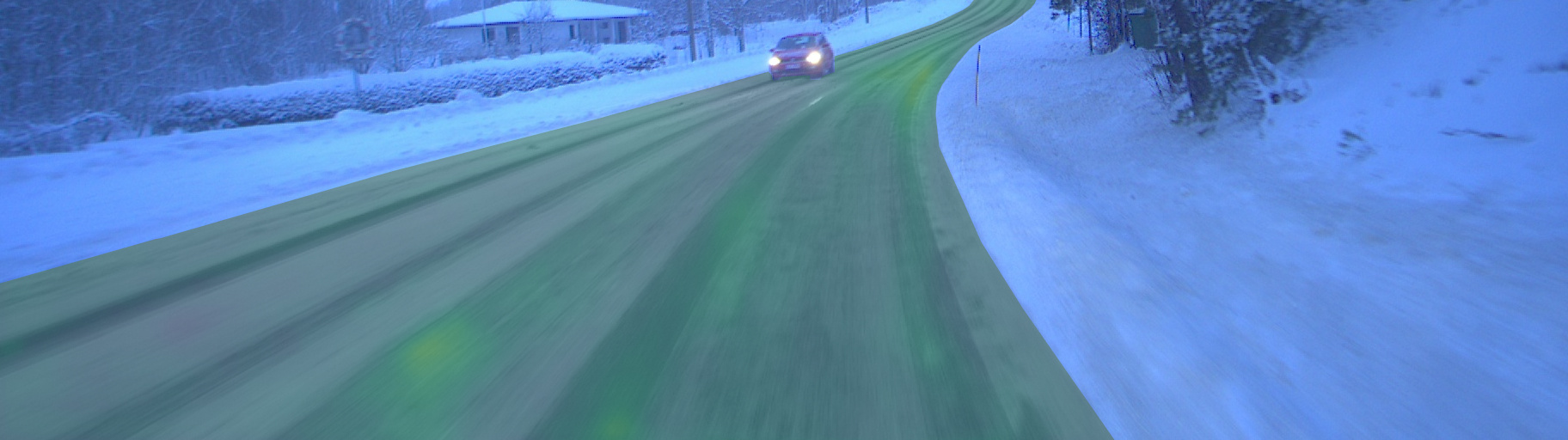}
&
\includegraphics[width=0.24\linewidth]{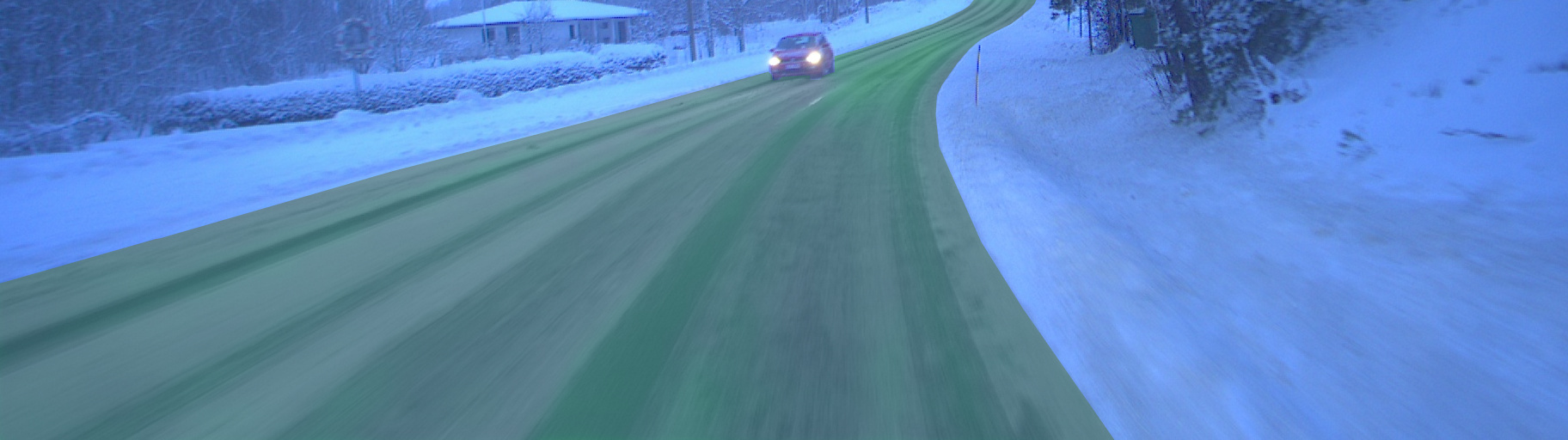}
&
\includegraphics[width=0.24\linewidth]{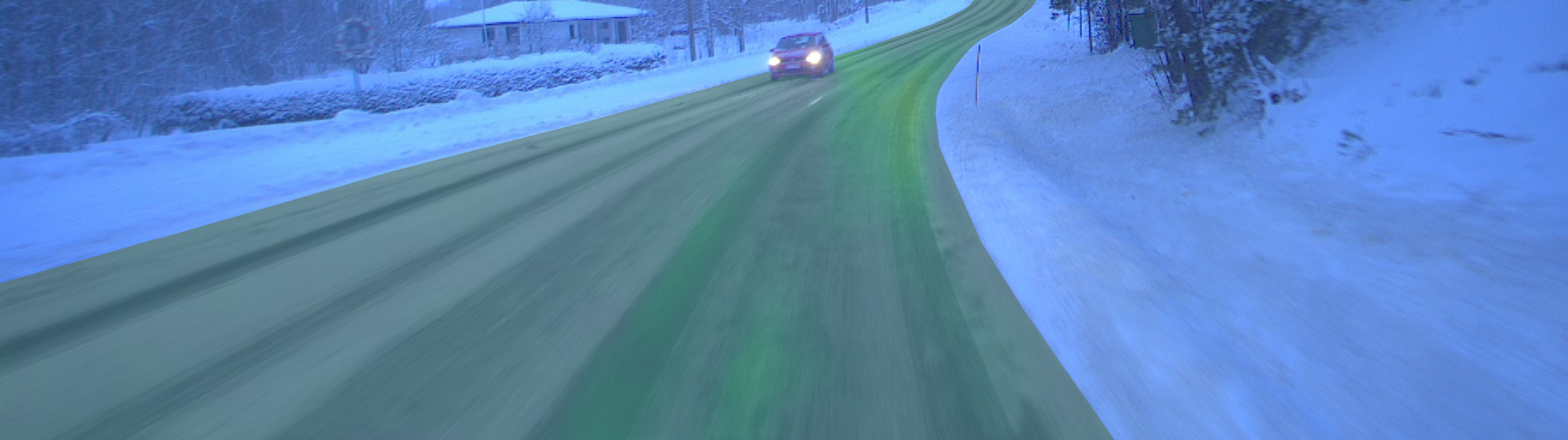}
\\

\end{tabular}

\includegraphics[width=0.4\textwidth]{imgs/cbar.pdf}

\captionof{figure}{Output visualizations of the additional sensor fusion models.}

\end{smashminipage}

\end{landscape}